\documentclass[10pt,twocolumn,letterpaper]{article}

\usepackage{cvpr}              

\usepackage{graphicx}
\usepackage{amsmath}
\usepackage{amssymb}
\usepackage{booktabs}
\usepackage{color}
\usepackage{epsfig}
\usepackage{graphicx}
\usepackage{amsmath}
\usepackage{verbatim}
\usepackage{amssymb}
\usepackage{bm}
\usepackage{multirow}
\usepackage{booktabs}
\usepackage{tabularx}
\usepackage{url}
\usepackage{graphicx}
\usepackage{mathrsfs}
\usepackage{amsmath}
\usepackage{multirow}
\usepackage{rotfloat}

\usepackage{graphicx}
\usepackage[table,xcdraw]{xcolor}
\usepackage[normalem]{ulem}
\useunder{\uline}{\ul}{}

\usepackage{bbding}
\usepackage{pifont}
\usepackage{wasysym}
\usepackage[table,xcdraw]{xcolor}
\usepackage[colorlinks,linkcolor=red,anchorcolor=green,citecolor=blue]{hyperref}

\usepackage[capitalize]{cleveref}
\crefname{section}{Sec.}{Secs.}
\Crefname{section}{Section}{Sections}
\Crefname{table}{Table}{Tables}
\crefname{table}{Tab.}{Tabs.}

\def\cvprPaperID{7236} 
\def\confName{CVPR}
\def\confYear{2022}

\begin{document}

\title{Target-aware Dual Adversarial Learning and a Multi-scenario Multi-Modality Benchmark to Fuse Infrared and Visible for Object Detection}
\author{Jinyuan Liu$^{\dag}$, Xin Fan$^{\ddag}$\thanks{Corresponding author.}, Zhanbo Huang$^\ddag$, Guanyao Wu$^\ddag$, Risheng Liu$^{\ddag,\S}$, Wei Zhong$^{\ddag}$, Zhongxuan Luo$^{\dag}$\\	
\normalsize$^\dag$School of Software Technology, Dalian University of Technology\\
\normalsize $^\ddag$DUT-RU International School of Information Science \& Engineering, Dalian University of Technology\\
\normalsize $^\S$Peng Cheng Laboratory\\
{\tt \small \{atlantis918\}@hotmail.com, \{zbhuang,rollingplain\}@mail.dlut.edu.cn, \{xin.fan,rsliu\}@dlut.edu.cn}
}

\twocolumn[{%
	\renewcommand\twocolumn[1][]{#1}%
	\maketitle
		\vspace{-1.2cm}
	\begin{center}
		\centering
		\captionsetup{type=figure}
		\includegraphics[width=.9\textwidth,height=3.95cm]{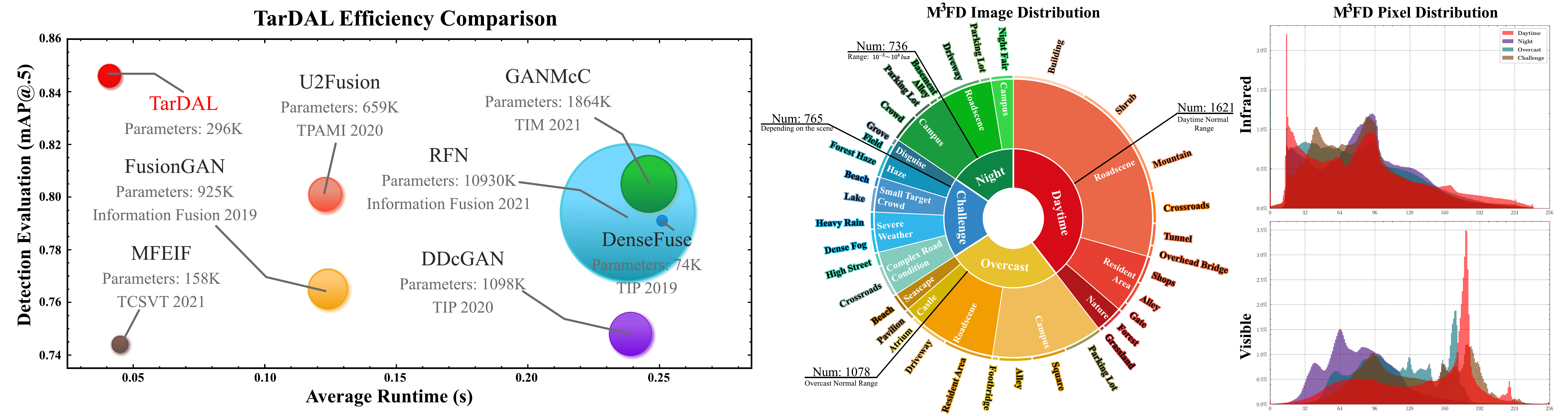}
		\captionof{figure}{From the left to right: Detection accuracy and computational comparisons with the state-of-the-art, the scenario and pixel distributions of our benchmark M$^3$FD. Ours outperforms all counterparts with higher detection rates, lower average runtime, and fewer training parameters. M$^3$FD covers comprehensive scenarios with a wide range of pixel variations especially on both modalities.}
		\label{fig:data}
	\end{center}%
}]
\maketitle

\begin{abstract}
	\vspace{-0.3cm} 
This study addresses the issue of fusing infrared and visible images that appear differently for object detection. Aiming at generating an image of high visual quality, previous approaches discover commons underlying the two modalities and fuse upon the common space either by iterative optimization or deep networks. These approaches neglect that modality differences implying the complementary information are extremely important for both fusion and subsequent detection task. This paper proposes a bilevel optimization formulation for the joint problem of fusion and detection, and then unrolls to a target-aware Dual Adversarial Learning (TarDAL) network for fusion and a commonly used detection network. The fusion network with one generator and dual discriminators seeks commons while learning from differences, which preserves structural information of targets from the infrared and textural details from the visible. Furthermore, we build a synchronized imaging system with calibrated infrared and optical sensors, and collect currently the most comprehensive benchmark covering a wide range of scenarios. Extensive experiments on several public datasets and our benchmark demonstrate that our method outputs not only visually appealing fusion but also higher detection mAP than the state-of-the-art approaches. The source code and benchmark are available at \url{https://github.com/dlut-dimt/TarDAL}. 
\end{abstract}

\vspace{-0.6cm} 
\section{Introduction}

\label{sec:intro}
Multi-modality imaging has attracted significant attention in a wide range of applications,~\emph{e.g.}, surveillance~\cite{survilence18} and autonomous driving~\cite{automaus2018}, with the rapid development of sensing hardware. Especially, the combination of infrared and visible sensors has remarkable advantages for subsequent intelligent processing~\cite{zhao2018multi,zhang2020vifb,li2021multiple}. Visible imaging provides rich details with high spatial resolution under well-defined lighting conditions while infrared sensors, capturing ambient temperature variations emitted from objects, highlight structures of thermal targets insensitive to lighting changes. Unfortunately, infrared images are often accompanied by blurred details with lower spatial resolution. Owing to their evident appearance discrepancy, it is challenging to fuse visually appealing images and/or to support higher-level vision tasks such as segmentation~\cite{pu2018graphnet,fu2019dual}, tracking~\cite{tracking2020,SiamRPN++}, and detection~\cite{takumi2017multispectral}, by making full use of the complementary information from the infrared and visible.

Numerous infrared and visible image fusion (IVIF) approaches that aim at improving visual quality have been developed in the past decades. Traditional multi-scale transform\cite{ma2017infrared,li2013image}, optimization model~\cite{ma2016infrared,zhao2020bayesian,bilevel}, spare representation~\cite{zhang2018sparse,zhu2018novel}, and subspace~ methods attempt to discover intrinsic common features of the two modalities and to design appropriate weighting rules for fusion. These approaches typically have to invoke a time consuming iterative optimization process. Recently, researchers introduce deep networks into IVIF by learning powerful feature representation and/or weighting strategies when redundant well-prepared image pairs are available for training~\cite{ma2019fusiongan,U2Fusion2020,ddcgan,li2018densefuse,MFEIF2021,GANMcC}. The fusion turns out to be an efficient inference process yielding fruitful quality improvements.

Nevertheless, either traditional or deep IVIF approaches strive for quality improving but leave alone the follow-up detection, which is the key to many practical computer vision applications. The fusion emphasizes more on `seeking commons' but neglects the differences of these two modalities on presenting structural information of targets and textural details of ambient background. These differences play a critical role on differentiating distinct features of targets for object detection meanwhile generating clear appearance of high contrast favorable for human inspection.     

Moreover,~learning from these differences (actually complementary information) demands a comprehensive collections of imaging data from the two modalities. The images capturing in scenarios varying with lighting and weather exhibit significantly different characteristics from both modalities. Unfortunately, existing data collections only cover limited conditions, placing an obstacle to learn the complementary information and validate the effectiveness. 

This paper proposes a bilevel optimization formulation for the joint problem of fusion and detection.~This formulation unrolls to a well-designed dual-adversarial fusion network, composed of one generator and two target-aware discriminators, and a commonly used detection network. One discriminator distinguishes foreground \emph{thermal targets} from the image domain of infrared imaging while the other one differentiates the background \emph{textural details} from gradient domain of the visible image. We also derive a cooperative training strategy to learn optimal parameters for the two networks. Figure~\ref{fig:data} demonstrates that our method accurately detects objects from target-distinct and visual-appealing fusion with less time and fewer parameters than the state-of-the-art (SOTA). Our contributions are four-fold:
\vspace{-0.6cm} 
\begin{itemize}
	\item We embrace both image fusion and object detection with a bilevel optimization formulation, producing high detection accuracy as well as the fused image with better visual effects. 
	\vspace{-0.3cm}
	\item We devise a target-aware dual adversarial learning network (TarDAL) with fewer parameters for detection-oriented fusion. This one-generator and dual-discriminator network `seeks commons while learning from differences' that preserves information of targets from the infrared and textural details from the visible.
	\vspace{-0.3cm}
	\item We derive a cooperative training scheme from the bi-level formulation yielding optimal network parameters for fast inference (fusion and detection).    
	\vspace{-0.3cm}
	\item We build a synchronized imaging system with well-calibrated infrared and optical sensors and collect a multi-scenario multi-modality dataset (M$^3$FD) with $4,177$ aligned infrared and visible image pairs and $23,635$ annotated objects. The dataset covers four major scenarios with various environments, illumination, season, and weather, having a wide range of pixel variations, as shown in Figure~\ref{fig:data}.
\end{itemize}
\section{Related Works}
The fusion module is critical to detect objects from multi-modality sensors. This section briefly reviews previous learning-based IVIF approaches closely related to ours and available benchmarks that are necessary for learningand empirical evaluation.
\begin{figure*}[!htb]
	\centering
	\setlength{\tabcolsep}{1pt} 
	
	\includegraphics[width=0.98\textwidth,height=0.24\textheight]{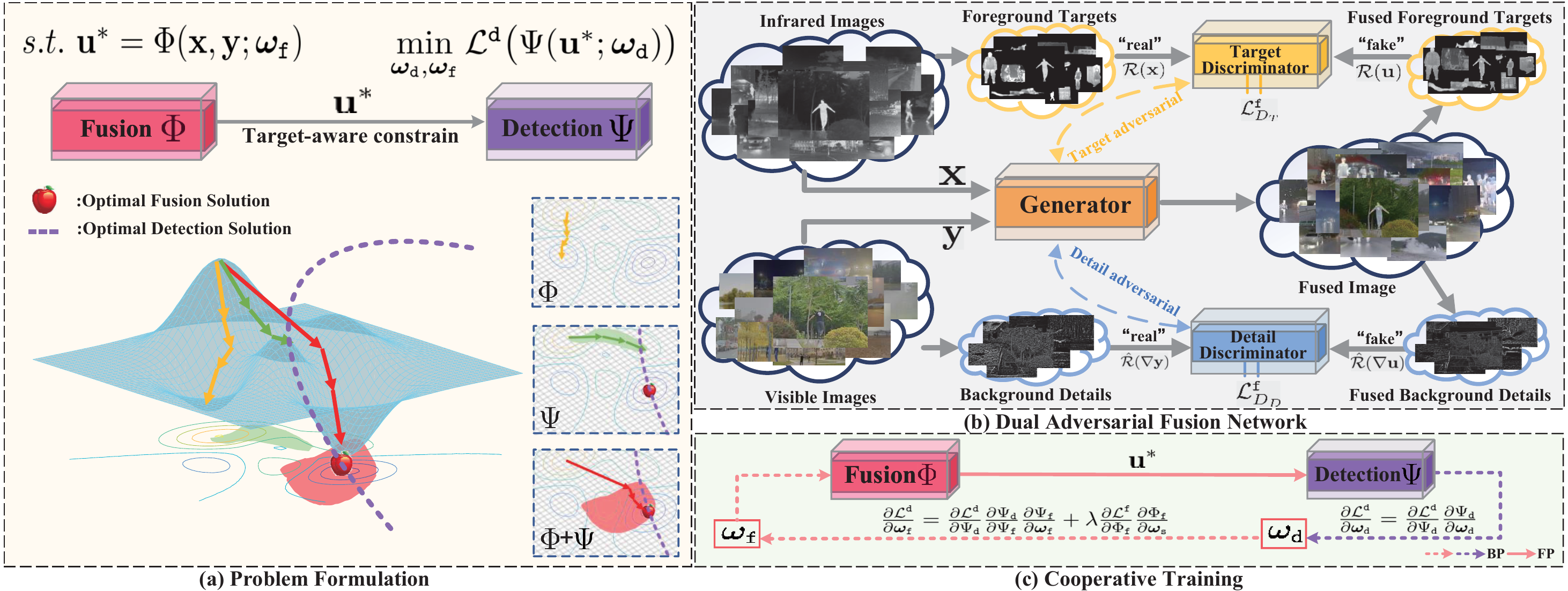}
	
	\caption{Methodology framework: (a) bilevel optimization formulation for fusion and detection, (b) target-aware adversarial dual learning network for fusion, and (c) cooperative training scheme.}
	\label{fig:workflow}
	\vspace{-0.5cm}  
\end{figure*}
\begin{table*}[!hbt]
	\begin{center}
		\centering
		\begin{tabular}{lccccccccc}
			\toprule
			\multicolumn{10}{l}{\textbf{Scene} : \ding{172}: Road \ \ \ \  \ding{173}: Campus \ \ \ \ \ding{174}: Street \ \ \ \ \ding{175}: Hash Weather \ \ \ \ \ding{176}: Disguise \ \ \ \ \ding{177}: Smoogy \ \ \ \ \ding{178}: Forest \ \ \  \ding{179}: Others}
			\\
			\hline
			\hline
			&Dataset&Img pairs&Resolution&Color&Camera angle&Nighttime&Objects&Scene&Annotation\\
			\hline
			&TNO&261&768$\times$576& \ding{55}&horizontal&65&few&\ding{172}\ding{176}\ding{177}\ding{178}\ding{179}&\ding{55}\\
			&INO&2100&328$\times$254&\ding{52}&surveillance&\ding{55}&few&\ding{172}\ding{175}\ding{179}&\ding{55}\\
			&OUS&285&320$\times$240&\ding{52}&surveillance&\ding{55}&few&\ding{172}&\ding{55}\\
			&Roadscene&221&768$\times$576&\ding{52}&driving&122&medium&\ding{172}\ding{174}&\ding{55}\\
			&Multispectral&2999&768$\times$576&\ding{52}&driving&1139&14146&\ding{172}\ding{174}&\ding{52}\\
			&M$^3$FD&\textbf{4200}&\textbf{1024}$\times$\textbf{768}&\ding{52}&\textbf{multiplication}&\textbf{1671}&\textbf{33603}&\ding{172}$\sim$\ding{179}&\ding{52}\\	
			\bottomrule
		\end{tabular}
	\end{center}
	\vspace{-0.5cm}  
	\caption{Illustration of M$^3$FD and existing aligned multi-modality datasets. Resolution refers to the average when it is different in a dataset.}\label{tab:m3data}
	\footnotesize
\end{table*}
\subsection{Learning-based approaches}
Deep learning has achieved promising progress in low-level vision tasks~\cite{malong,U2Fusion2020,malongcvpr,MFEIF2021,GANMcC,zhao2020didfuse,liu2021searching,liu2019knowledge} due to the powerful nonlinear fitting ability of multi-layer neural networks. Early works plugged deep networks into the IVIF process acting as a module of feature extraction or weights generation~\cite{li2018densefuse,MFEIF2021,smoa,bilevel}. Liu~\emph{et al.}~\cite{bilevel} cascaded two pre-trained CNNs, one for feature and the other for weights learning. Researchers also resort to an end-to-end architecture so that one-step network inference can generate a plausible fused image by one set of network parameters. Li~\emph{et al.}~\cite{rfn2021} introduced a residual fusion network to learn enhanced features in a common space, yielding structure consistent results favorable to human inspection.   

More recently, the IVIF approaches based on generative adversarial networks (GAN)~\cite{R2GAN,SPfusiongan,duanldiscriminator} produce appealing results by transporting different distributions to the desired one~\cite{ma2019fusiongan,ddcgan,GANMcC}. For the first time, Ma~\emph{et al.} introduced an adversarial game between the fused and visible in order to enhance texture details~\cite{ma2019fusiongan}. However, this signal adversarial mechanism may lose the vital information from the infrared. Ma~\emph{et al.} apply an identical adversarial strategy to both the visible and infrared, which partially compensates the infrared information~\cite{ddcgan}. Unfortunately, all these approaches fail capturing the different characteristics of these two imaging types. It is worth investigating these differences complementary to each other from which both fusion and detection can benefit.

\subsection{Benchmarks}
In recent years, we have witnessed the rapid evolution of IVIF benchmarks, including the TNO Image Fusion~\cite{TNO}, INO Videos Analytics\footnote{https://www.ino.ca/en/technologies/video-analytics-dataset/videos/}, OSU Color-Thermal\footnote{http://vcipl-okstate.org/
	pbvs/bench/}, and RoadScene~\cite{U2Fusion2020} and Multispectral datasets~\cite{takumi2017multispectral}. The TNO dataset~\cite{TNO} is the most commonly used public available dateset for IVIF, which contains 261 pairs of multispectral imagery at day and night time. The INO dataset is provided by the National Optics Institute of Canada and contains aligned infrared and visible pairs. It contributes to developing multiple sensor types for video analysis applications in challenging environments.The OSU Color-Thermal Database is established for fusion-based object detection containing 285 pairs of registered infrared and color visible images. The whole dataset is collected at a busy pathway on the Ohio State University Campus during the daytime. Xu~\emph{et al.}~released Roadscene, having 221 aligned infrared and visible pairs taken in the road scene containing rich objects, such as vehicles and pedestrians~\cite{U2Fusion2020}. Takumi~\emph{et al.}~\cite{takumi2017multispectral}~proposed a novel Multispectral dataset for autonomous driving that consists of RGB, NIR, MIR, and FIR images and annotated object categories. 

Table~\ref{tab:m3data} lists the profiles of these datasets such as scale, resolution, lighting, and scenario categories. The low image resolution, limited number of object and scenario types, and few labels stumble wide applications of existing datasets to the higher-level detection task on multiple modalities.

\section{The Proposed Method}
This section details our method, staring from the bilevel optimization formulation of fusion and detection. Then, we elaborate the target-aware dual adversarial learning network for fusion. Finally, we give a cooperative training scheme to learn optimal parameters for both fusion and detection.
\subsection{Problem formulation}
Unlike previous approaches catering for high visual quality, we state that IVIF has to generate an image that benefits~\emph{both visual inspection and computer perception}, namely detection-oriented fusion. Suppose that the infrared, visible and fused are all gray-scale with the size of $m\times n$, denoted as column vectors $\mathbf{x}$, $\mathbf{y}$, and $\mathbf{u}\in\mathbb{R}^{{mn}\times{1}}$,~respectively. 
Following the truism Stackelberg's theory~\cite{Ochs2015Bilevel,BILEVEL1,BILEVEL2}, we formulate the detection-oriented fusion as a bilevel optimization model: 
\vspace{-0.1cm} 
\begin{eqnarray}
&\quad\min\limits_{\bm{\omega}_{\mathtt{d}}} \mathcal{L}^{\mathtt{d}}\big(\Psi(\mathbf{u}^*;\bm{\omega}_{\mathtt{d}})\big),\\
&s.t. \ \mathbf{u}^*\in\arg\min\limits_{\mathbf{u}} f(\mathbf{u;x,y}) + g_{\mathtt{T}}(\mathbf{u;x}) + g_{\mathtt{D}}(\mathbf{u;y}),
\end{eqnarray}
where~$\mathcal{L}^{\mathtt{d}}$ is the detection-specific training loss and $\Psi$ denotes a detection network with learnable parameters~$\bm{\omega}_{\mathtt{d}}$.~$f$($\cdot$)~is an energy-based fidelity term containing the fused image~$\mathbf{u}$, and source images~$\mathbf{x}$ and $\mathbf{y}$ while~$g_{\mathtt{T}}$~($\cdot$)~and~$ g_{\mathtt{D}}$~($\cdot$) are two feasibility constraints defined on the infrared and visible, respectively.

Figure~\ref{fig:workflow}(a) illustrates that this bilevel formulation makes it possible to find the solution mutually favoring fusion and detection. Nevertheless, it is nontrivial to solve Eq.~(2) by a traditional optimizing technique as the fusion task is not a simple equality/inequality constraint. Instead, we introduce a fusion network~$\Phi$ with learned parameters~$\bm{\omega}_{\mathtt{f}}$ and convert the bilevel optimization to single-level:         
\begin{eqnarray}
&\quad\min\limits_{\bm{\omega}_{\mathtt{d}}, \bm{\omega}_{\mathtt{f}}} \mathcal{L}^{\mathtt{d}}\big(\Psi(\mathbf{u}^*;\bm{\omega}_{\mathtt{d}})\big),
s.t. \ \mathbf{u}^*=\Phi(\mathbf{x,y}; {\bm{\omega}_{\mathtt{f}}}).
\end{eqnarray}
Hence, we unroll the optimization into two learning networks~$\Phi$ and $\Psi$. We adopt YOLOv5\footnote{https://github.com/ultralytics/YOLOv5} as our backbone for the detection network~$\Psi$, where~$\mathcal{L}^{\mathtt{d}}$ also follows its setting, and carefully design the fusion network~$\Phi$ as below.
\vspace{-0.2cm} 
\subsection{Target-aware dual adversarial network}
Typical deep fusion methods strive for learning common features underlying the two modalities that appear differently. Instead, our fusion network seeks commons while learning from differences that imply complementary characteristics of these two types of imaging. Typically, the infrared highlights distinct structures of targets while the visible provides textural details of background.

We introduce an adversarial game that consists one generator and two discriminators in order to combine common with distinct features from the two modalities, as shown in Figure~\ref{fig:workflow}(b). The generator~$G$ is encouraged to provide a realistic fused image to simultaneously fool both discriminators. The target discriminator~$D_T$ evaluates the \emph{intensity} consistence between the targets from the \emph{infrared} and those masked out from the fused given by~$G$ (the top row of Figure~\ref{fig:workflow}(b)); the detail discriminator~$D_D$ discriminates the \emph{gradient} distribution of the \emph{visible} from that of the fused (the bottom row of Figure~\ref{fig:workflow}(b)). These two discriminators work in different domains as targets exhibit consistent intensity distribution while gradients characterize textures.

\textbf{Generator}: The generator contributes to generate a fused image that preserves overall structures and  maintains a similar intensity distribution as source images. The commonly used structural similarity index~(SSIM)~\cite{wang2004image} acts as the loss function: 
\vspace{-0.2cm} 
\begin{equation}
\mathcal{L}_{\mathtt{SSIM}} = (1-{\mathtt{SSIM}}_{\mathbf{u},\mathbf{x}})/2 + (1 - {\mathtt{SSIM}}_{\mathbf{u},\mathbf{y}})/2, 
\end{equation}
where $\mathcal{L}_{\mathtt{SSIM}}$ denotes structure similarity loss. To balance the pixel intensity distribution of source images, we introduce a pixel loss based on the saliency degree weight~(SDW). Supposing that the saliency value of~$\mathbf{x}$ at the $k$th pixel can be obtained by~$\bm{S}_{\mathbf{x}(k)}=\sum\limits_{i=0}^{255}\bm{H}_{\mathbf{x}}(i)\lvert \mathbf{x}(k)-i\rvert $, where~$\mathbf{x}(k)$ is the value of the $k$th pixel and~$\bm{H}_\mathbf{x}$ is the histogram  of pixel value~$i$, we define the pixel loss $\mathcal{L}_{\rm pixel}$ as:
\begin{equation}
\mathcal{L}_{\mathtt{pixel}} = \lVert \mathbf{u}- \bm{\omega_1}\mathbf{x}\rVert_1 + \lVert \mathbf{u}- \bm{\omega_2}\mathbf{y}\rVert_1,
\end{equation}
where~~$\bm{\omega_1}= \bm{S}_{\mathbf{x}}(k)/[\bm{S}_{\mathbf{x}}(k) - \bm{S}_{\mathbf{y}}(k)], \bm{\omega_2}  = 1 - \bm{\omega_1}$.

We employ a 5-layer dense block\cite{huang2019convolutional} as $G$ to extract common features, and then use a merge block with three convolutional layers for feature aggregation. Each convolutional layer consists of one convolutional operation, batch normalization and ReLU activation function. The generated fused images $\mathbf{u}$ has the same size with the sources. 

\textbf{Target and detail discriminators}: The target discriminator~$D_T$ is used to distinguish the foreground thermal targets of fused result to the infrared while the detail discriminator~$D_D$ contributes to distinguish the background details of fused result to the visible. We employ a pre-trained saliency detection network~\cite{deng2018r3net} to calculate the target mask~$\mathbf{m}$ from infrared images so that the two discriminators can perform on their respective regions (target and background). Thus, we define the adversarial loss $\mathcal{{L}}^{\rm adv}_{\mathtt{f}}$ as:
\vspace{-0.1cm} 
\begin{equation}
\mathcal{{L}}_{D_T}^{\mathtt{f}}=\mathbb{E}_{x\sim \tilde{p}({\mathcal{R}(\mathbf{x})})}[D(x)] - \mathbb{E}_{\tilde{x}\sim \tilde{p}({\mathcal{R}(\mathbf{u})})}[D(\tilde{x})],
\end{equation}
\begin{equation}
\mathcal{{L}}_{D_D}^{\mathtt{f}}=\mathbb{E}_{x\sim \tilde{p}({\mathcal{\hat{R}}(\nabla\mathbf{y})})}[D(x)] - \mathbb{E}_{\tilde{x}\sim \tilde{p}({\mathcal{\hat{R}}(\nabla\mathbf{u})})}[D(\tilde{x})],
\end{equation}
\begin{equation}
\mathcal{{L}}^{\rm adv}_{\mathbf{f}} =\mathcal{{L}}_{D_T}^\mathbf{f}+\mathcal{{L}}_{D_D}^\mathbf{f},
\end{equation}
where~$\mathcal{R} = \mathbf{x} \odot {\mathbf{m}}$~and~$\mathcal{\hat{R}} = 1 - \mathcal{R}$, differentiating targets from background, and $\odot$~denotes
the point-wise multiplication. $\nabla(\cdot)$ denotes a gradient operation, \emph{e.g.,} Sobel.

The adversarial loss functions of these discriminators calculate the Wasserstein divergence to mutually identify whether the foreground thermal targets and background texture details are realistic, defined as:
\vspace{-0.3cm} 
\begin{equation}
\mathcal{{L}}_{D_T}=\mathcal{{L}}_{D_T}^{\mathtt{f}}+ k\mathbb{E}_{\tilde{x}\sim \tilde{r}({\mathcal{R}(\mathbf{x})})}[(\lVert \nabla D_T(\tilde{x})\rVert)^{p}], \label{LDT}
\end{equation}
\begin{equation}
\mathcal{{L}}_{D_D}=\mathcal{{L}}_{D_D}^{\mathtt{f}} + k\mathbb{E}_{\tilde{x}\sim \tilde{r}({\mathcal{\hat{R}}(\nabla\mathbf{x})})}[(\lVert \nabla D_D(\tilde{x})\rVert)^{p}], \label{LDE}
\end{equation}
where~$\tilde{r}(x)$~denotes sample space that is similar to~$\tilde{p}(x)$. Commonly,~$k$~and~$p$~sets to 2 and 6, respectively.

\begin{figure}[!htb]
	\centering
	\setlength{\tabcolsep}{1pt}
	\begin{tabular}{c}
		
		\includegraphics[width=0.46\textwidth,height=0.06\textheight]{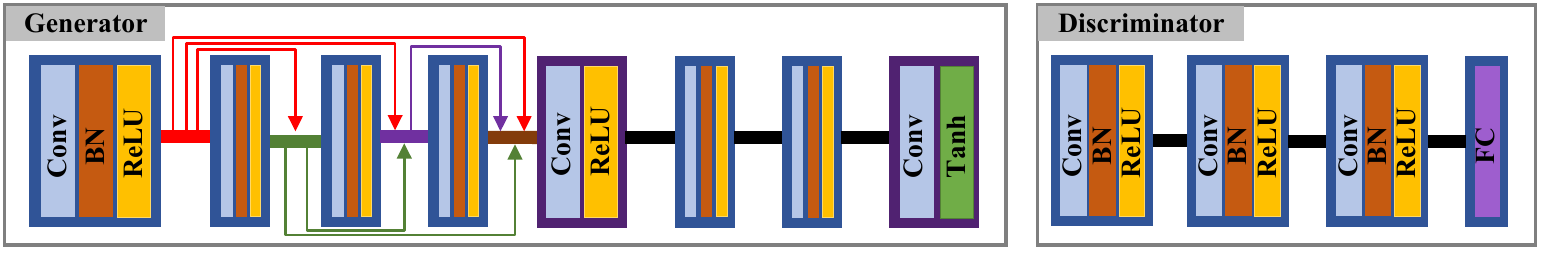}
		\\ 				
	\end{tabular}
	\vspace{-0.3cm}
	\caption{The architectures of our generator and discriminator.}
	\label{fig:arch}
\end{figure}

The two discriminators $D_T$ and $D_D$ share the same network structure, having four convolutional layers and one fully connection layer. Figure~\ref{fig:arch} demonstrates the detailed architectures of the generator and dual discriminators.

Totally, $\mathcal{L}^{\mathtt{f}}$ is combination of the aforementioned three main parts: 
\vspace{-0.5cm} 
\begin{equation}
\mathcal{L}^{\mathtt{f}} = \mathcal{L}_{\mathtt{SSIM}} + \alpha \mathcal{L}_{\mathtt{pixel}} + \beta \mathcal{{L}}^{\rm adv}_{\mathtt{f}},\label{LG} 
\end{equation}
where~$\alpha$~and~$\beta$~are the trade-off parameters.
\vspace{-0.2cm} 
\vspace{-0.6cm} 
\subsection{Cooperative training strategy}
\begin{figure*}[!htb]
	\centering
	\setlength{\tabcolsep}{1pt} 
	
	\includegraphics[width=0.98\textwidth,height=0.12\textheight]{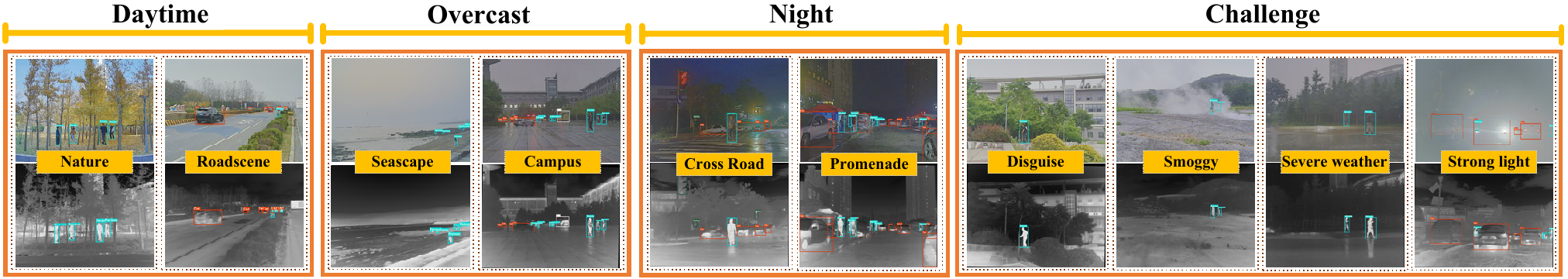}
	
	\caption{Visualization of infrared-visible images in our M$^3$FD dataset. The dataset covers extensive scenarios with various environments, illumination, season, and weather.}
	\label{fig:sample}
	\vspace{-0.5cm}  
\end{figure*}
The bilevel optimization naturally derive a cooperative training strategy to obtain  optimal network parameters~$\bm{\omega}=(\bm{\omega_{\mathtt{d}}}, \bm{\omega_{\mathtt{f}}})$. We introduce a a fusion regularizer~$\mathcal{L}^{\mathtt{f}}$ and convert Eq.~(3) optimizing detection subject to the fusion constraint to to a mutual optimization: 
\vspace{-0.2cm} 
\begin{eqnarray}
&\quad\min\limits_{\bm{\omega}_{\mathtt{d}}, \bm{\omega}_{\mathtt{f}}} \mathcal{L}^{\mathtt{d}}\big(\Psi(\mathbf{u}^*;\bm{\omega}_{\mathtt{d}})\big) + \lambda \mathcal{L}^{\mathtt{f}}\big(\Phi(\mathbf{x,y};\bm{\omega}_{\mathtt{f}})\big),\\
&s.t. \ \mathbf{u}^*=\Phi(\mathbf{x,y}; {\bm{\omega}_{\mathtt{f}}}),
\end{eqnarray}
where~$\lambda$ is the trade-off parameter. Rather than designing a weighting rule, this regularizer can well balance fusion and detection.

Figure~\ref{fig:workflow}(c) illustrates the flow of gradient propagation to cooperatively train the fusion and detection networks. The loss gradients with respect to~$\bm{\omega}_\mathtt{d}$~and~$\bm{\omega}_\mathtt{f}$ are calculated as:
\begin{equation}
\small
\frac{\partial\mathcal{L}^{\mathtt{d}}}{\partial\bm{\omega}_\mathtt{d}}=\frac{\partial\mathcal{L}^\mathtt{d}}{\partial\Psi_\mathtt{d}}\frac{\partial\Psi_\mathtt{d}}{\partial\bm{\omega}_\mathtt{d}}, \quad \frac{\partial\mathcal{L}^{\mathtt{d}}}{\partial\bm{\omega}_\mathtt{f}}=\frac{\partial\mathcal{L}^\mathtt{d}}{\partial\Psi_\mathtt{d}}\frac{\partial\Psi_\mathtt{d}}{\partial\Psi_\mathtt{f}}\frac{\partial\Psi_\mathtt{f}}{\partial\bm{\omega}_\mathtt{f}}+\lambda\frac{\partial\mathcal{L}^\mathtt{f}}{\partial\Psi_\mathtt{f}}\frac{\partial\Psi_\mathtt{f}}{\partial\bm{\omega}_\mathtt{f}}.
\small
\end{equation}
These equations reveal that the gradients of the detection loss w.r.t. the detection parameters along with those w.r.t. the fusion parameters are all back propagated and the latter also consists of the gradients of the fusion loss w.r.t. the fusion parameters.

Finally, this strategy cannot only generate a visually appealing image but also output accurate detection given the trained network parameters, enabling us to find the optimal solution to detection-oriented fusion and to converge more efficient than independent training schemes.

\section{Multi-scenario Multi-modality Benchmark}
Existing datasets with infrared and visible images can hardly be applied to learn and/or evaluate detection from multi-modality data. Our benchmark M$^3$FD contains infrared and visible images of high resolution covering diverse object types under various scenarios as given in the last row of Table~\ref{tab:m3data}.
\begin{figure}[!htb]
	\centering
	\setlength{\tabcolsep}{1pt}
	\begin{tabular}{c}
		
		\includegraphics[width=0.46\textwidth,height=0.1\textheight]{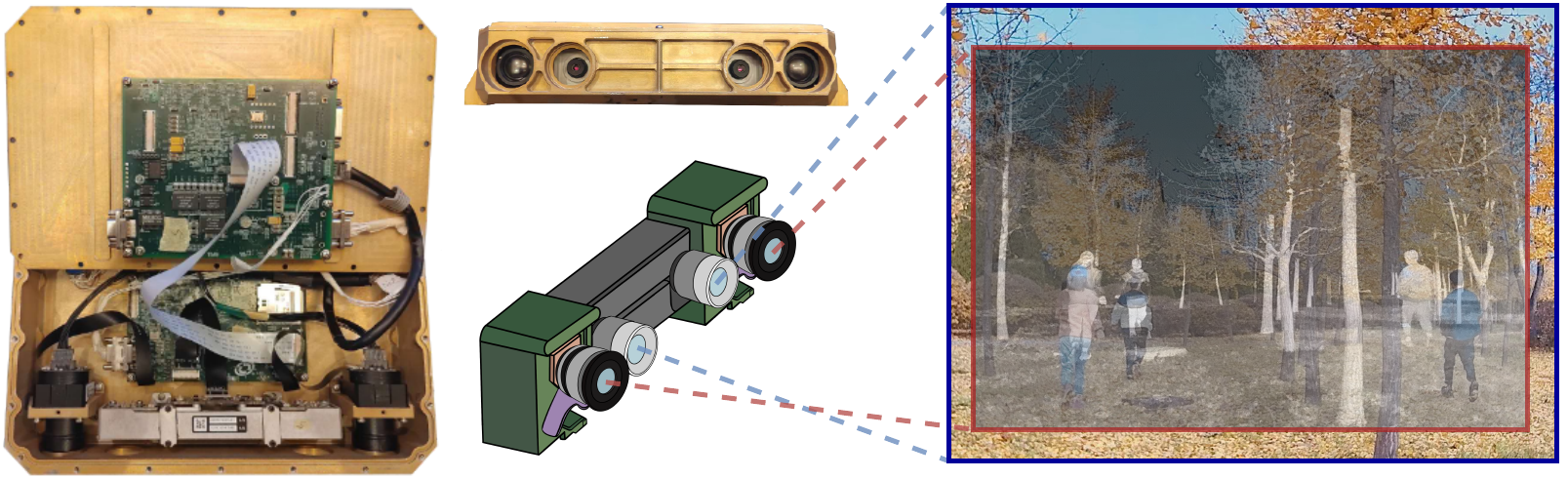}
		\\ 				
	\end{tabular}
	\vspace{-0.2cm}
	\caption{Illustration of our synchronize imaging system.}
	\label{fig:SYSTEM}
\end{figure}

We constructed a synchronized system containing one binocular optical camera and one binocular infrared sensor (shown in Figure~\ref{fig:SYSTEM}) in order to capture corresponding two-modality images of natural scenes. The baselines (distance between the focal centers of binocular lens) of the visible and infrared binocular cameras are $12cm$ and $20cm$, respectively. The optical center distance between the visible and infrared senors is $4cm$. Visible images have a high resolution of 1024$\times$768 and a wide imaging range while infrared images have a standard resolution of 640$\times$512 and the wavelength range is $8-14\mu m$.

We first calibrated all cameras to estimate their internal and external parameters, and then calculated a homography matrix that projects coordinates of infrared images to those of the visible. Eventually, we obtained well-aligned infrared/visible image pairs with the size of $1024\times768$ by warping all images to a common coordinate\footnote{This set includes pairs from one set of infrared and visible sensors, and the depth data from the binocular cameras will be published in the future.}.

We categorized all $4,200$ aligned pairs in M$^3$FD into four typical types,~\emph{i.e.}~Daytime, Overcast, Night, and Challenge, with ten sub-scenarios shown in Figure~\ref{fig:sample}. Meanwhile, we annotated $33,603$ objects of six classes,~\emph{i.e.,} People, Car, Bus, Motorcycle, Truck and Lamp, which commonly occur in surveillance and autonomous driving. The quantity and diversity of M$^3$FD pave the possibility to learn and evaluate object detection by fusing images.

\section{Experiments}
We performed~experimental~evaluations~on~four datasets~(three for IVIF,~\emph{i.e.,} TNO, Roadscene, and M$^3$FD, and two for object detection,~\emph{i.e.,} MS and M$^3$FD). 180/3,500 multi-modality images are selected and cropped to 24k/151k patches with 320$\times$320 pixels by random cropping and augmented for training the fusion and detection task, respectively. The tuning parameters $\alpha$ and $\beta$ are set to 20 and 0.1, respectively. The Adam optimizer updates the network parameters with the learning rate of $1.0\times 10^{-3}$ and exponential decay. The epoch is set to $300$ with batch size of $64$. Our approach is implemented on PyTorch with an NVIDIA Tesla V100 GPU.
\subsection{Results of infrared-visible image fusion}
We evaluate the fusion performance of TarDAL by making a comparison with 7 state-of-the-art methods, including DenseFuse~\cite{li2018densefuse}, FusionGAN~\cite{ma2019fusiongan}, RFN~\cite{rfn2021}, GANMcC~\cite{GANMcC}, DDcGAN~\cite{ddcgan}, MFEIF~\cite{MFEIF2021}, and U2Fusion~\cite{U2Fusion2020}.

\begin{figure*}[!htb]
	\centering
	\setlength{\tabcolsep}{1pt}
	\begin{tabular}{cccccccccc}
		
		\includegraphics[width=0.093\textwidth,height=0.075\textheight]{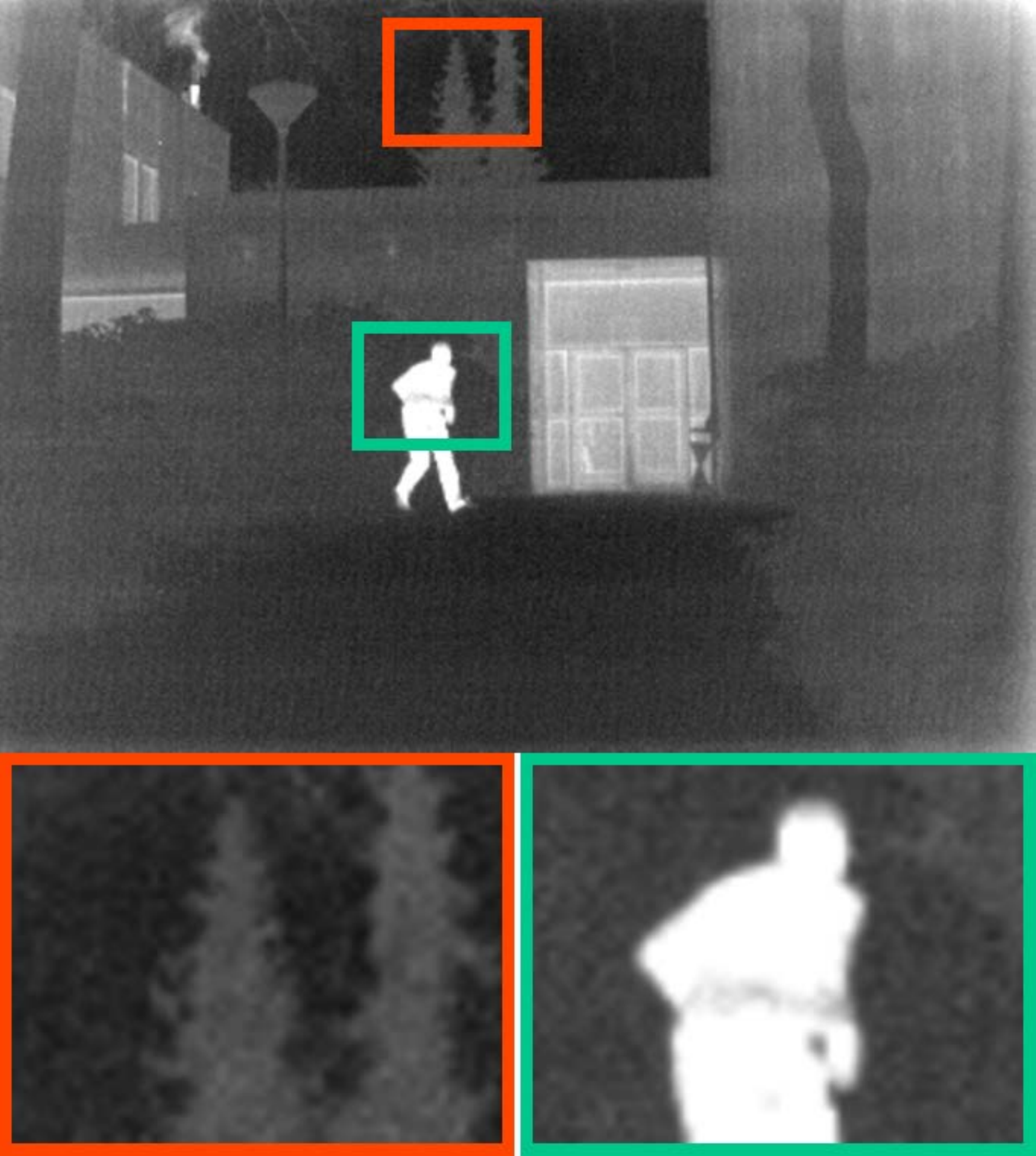}
		&\includegraphics[width=0.093\textwidth,height=0.075\textheight]{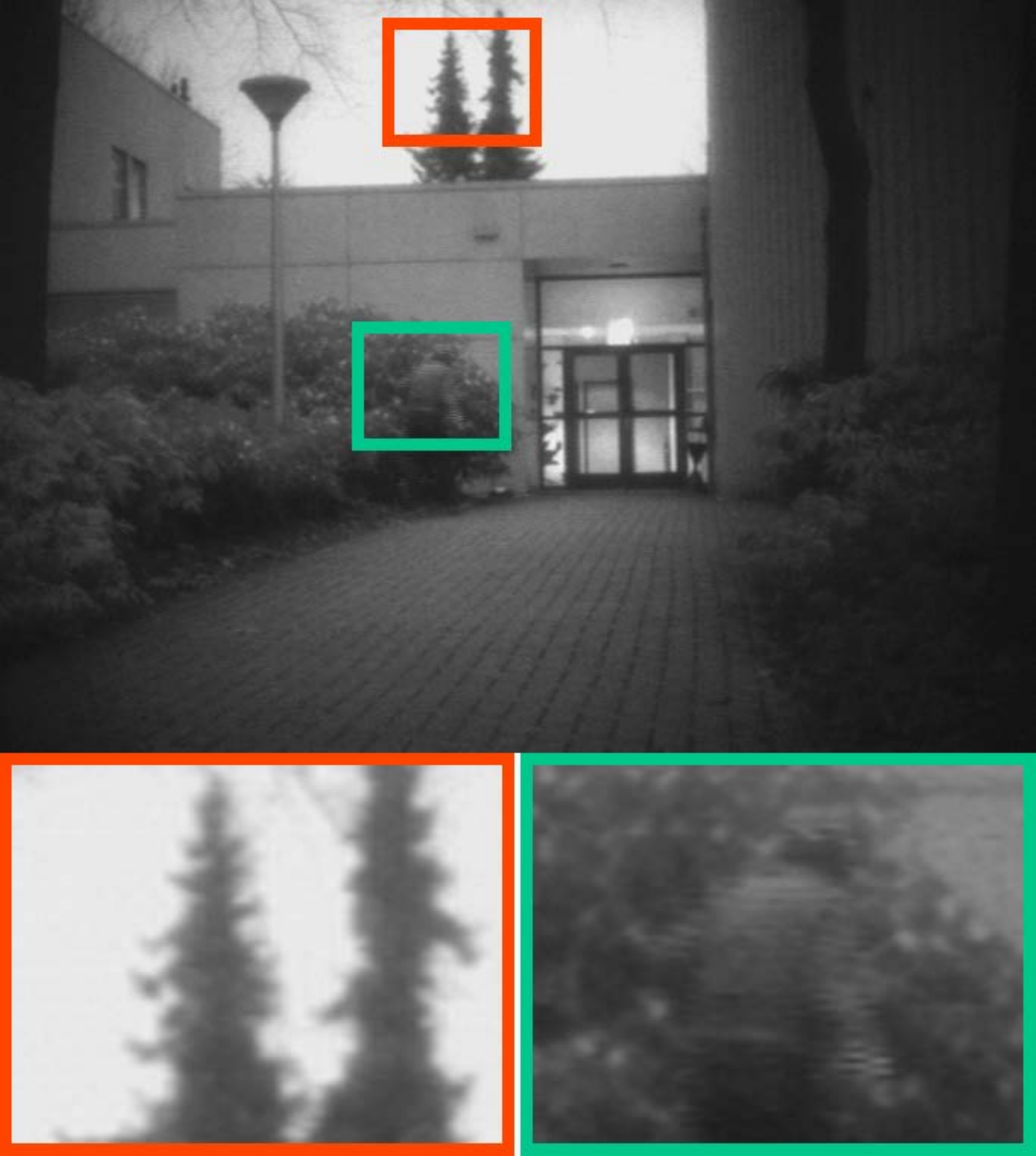}
		&\includegraphics[width=0.093\textwidth,height=0.075\textheight]{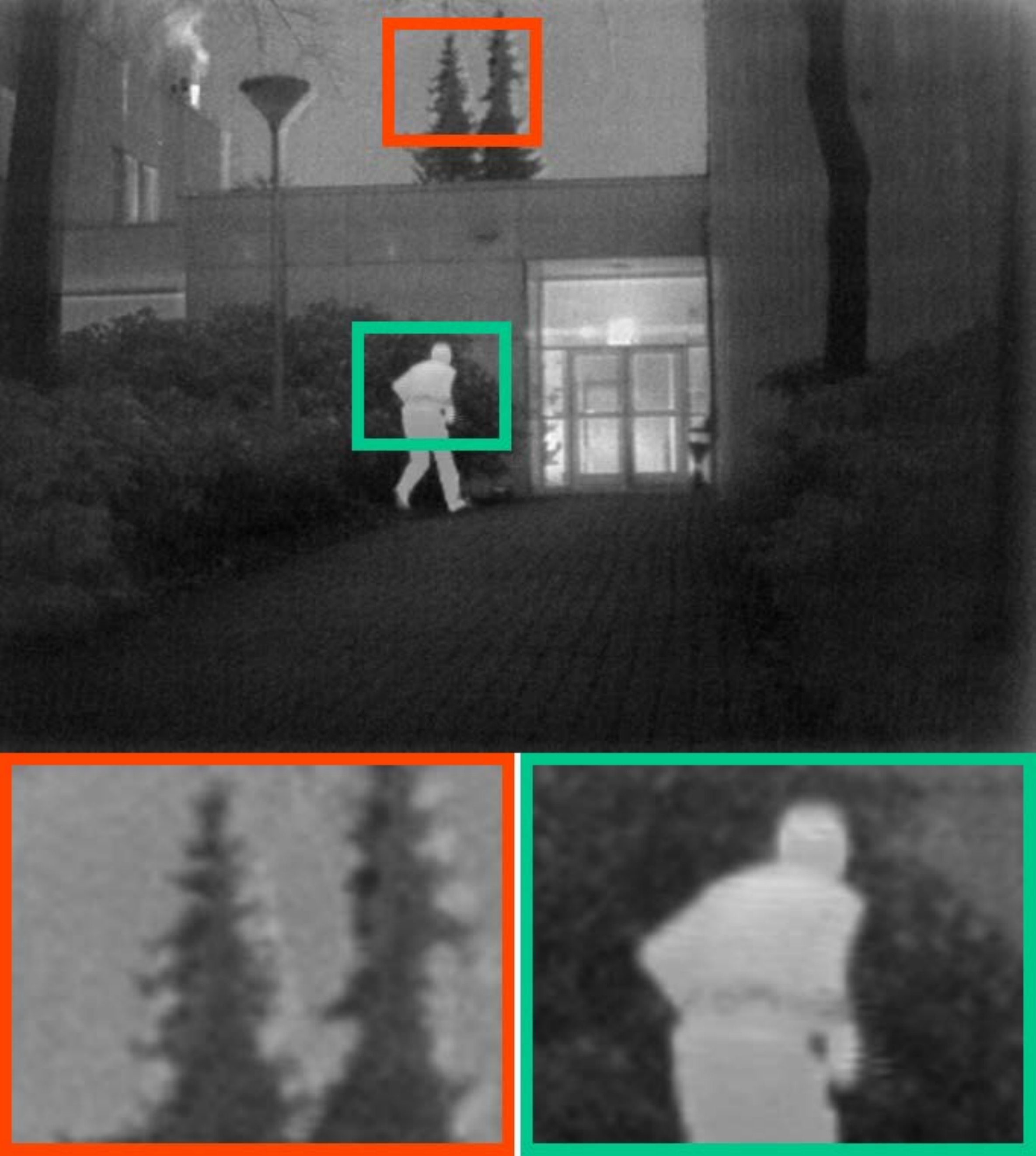}
		&\includegraphics[width=0.093\textwidth,height=0.075\textheight]{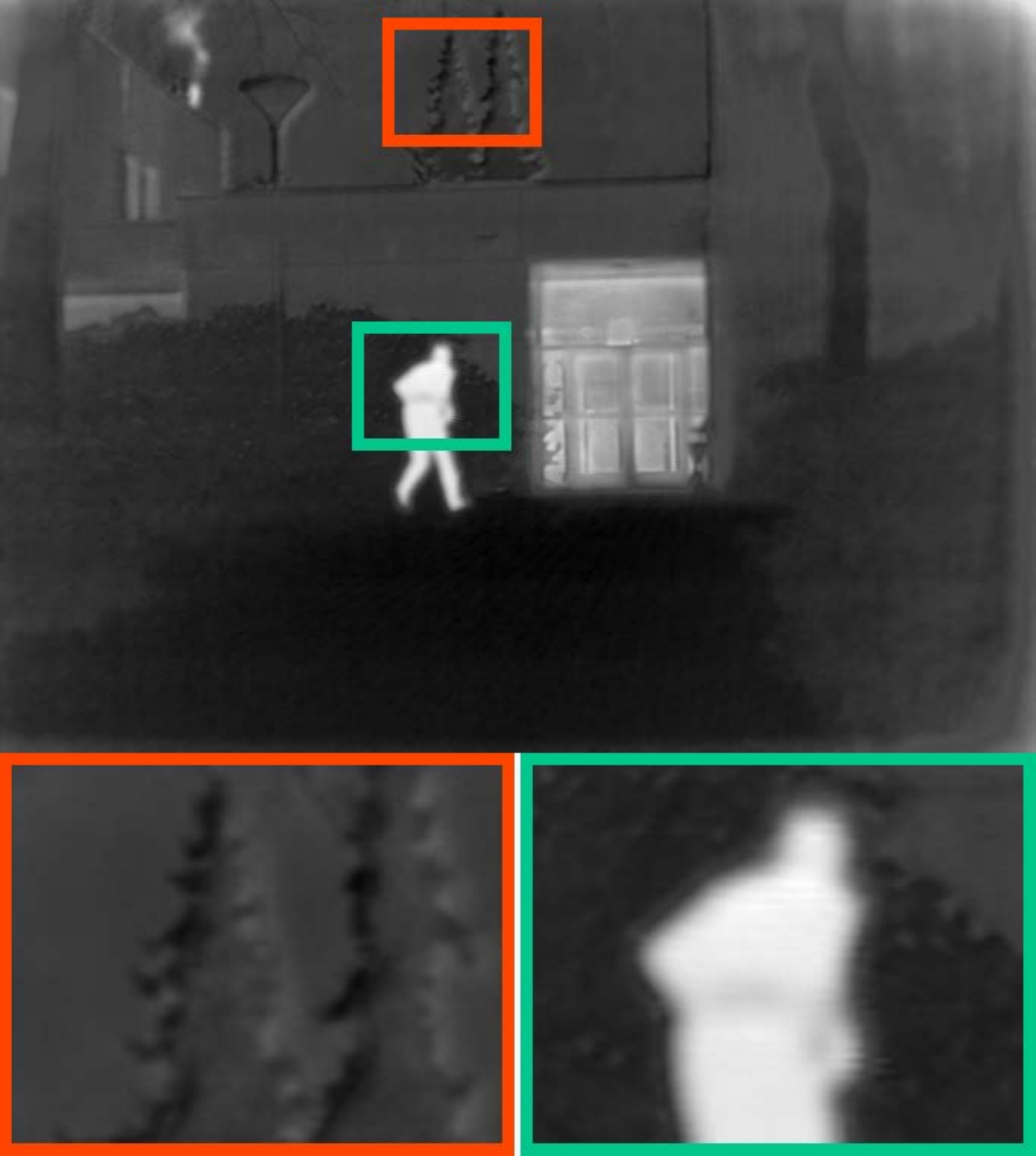}
		&\includegraphics[width=0.093\textwidth,height=0.075\textheight]{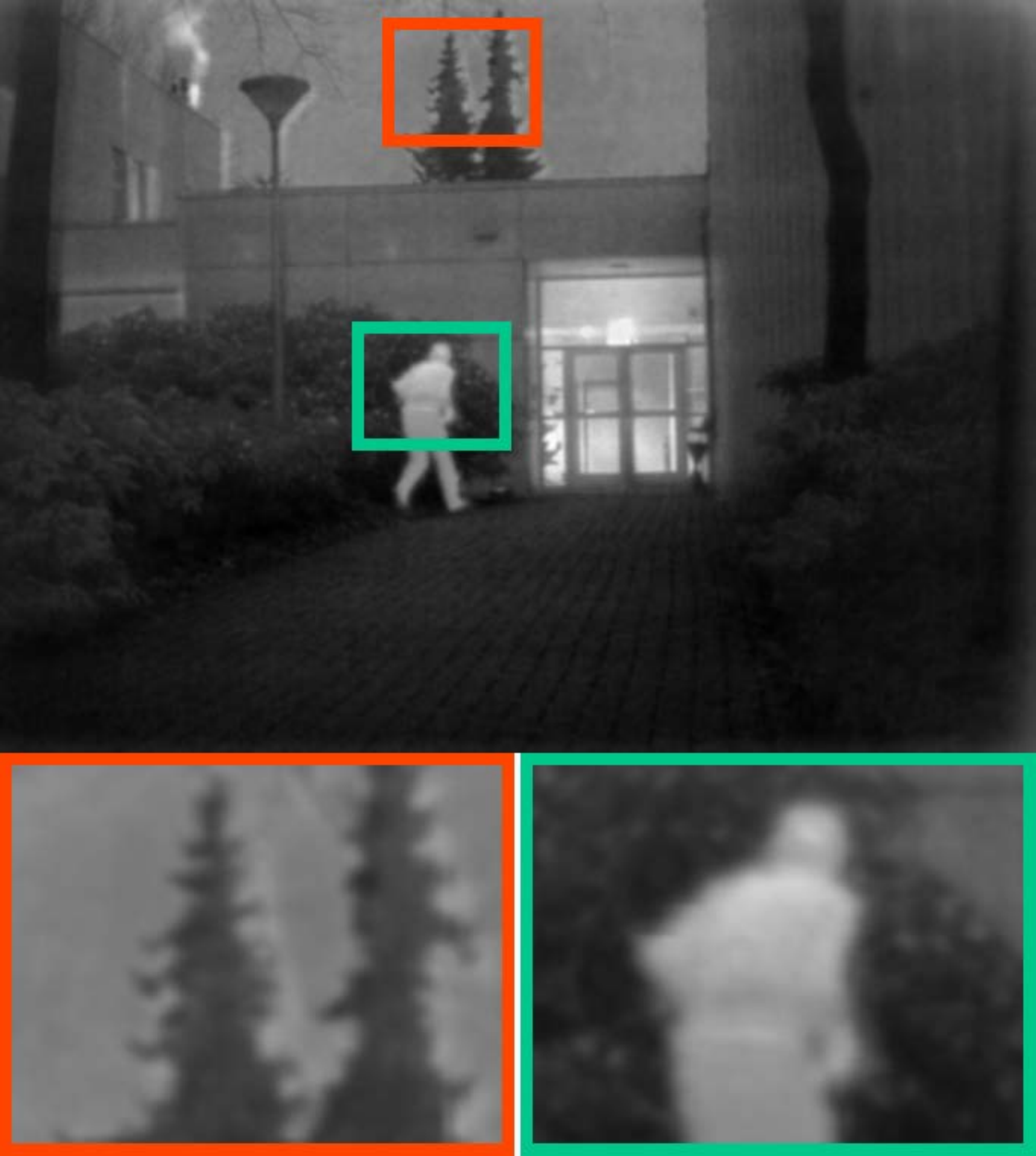}
		&\includegraphics[width=0.093\textwidth,height=0.075\textheight]{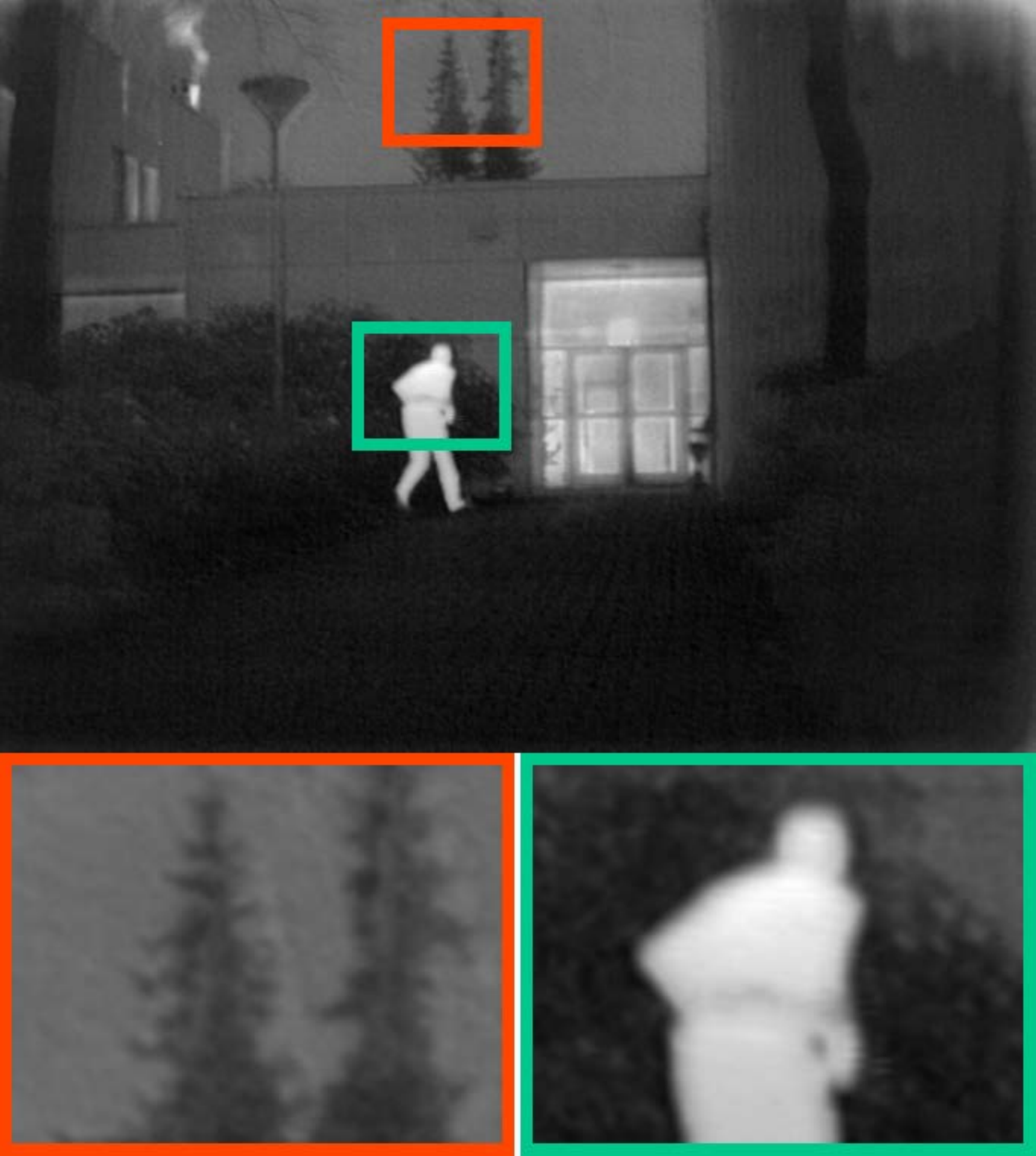}
		&\includegraphics[width=0.093\textwidth,height=0.075\textheight]{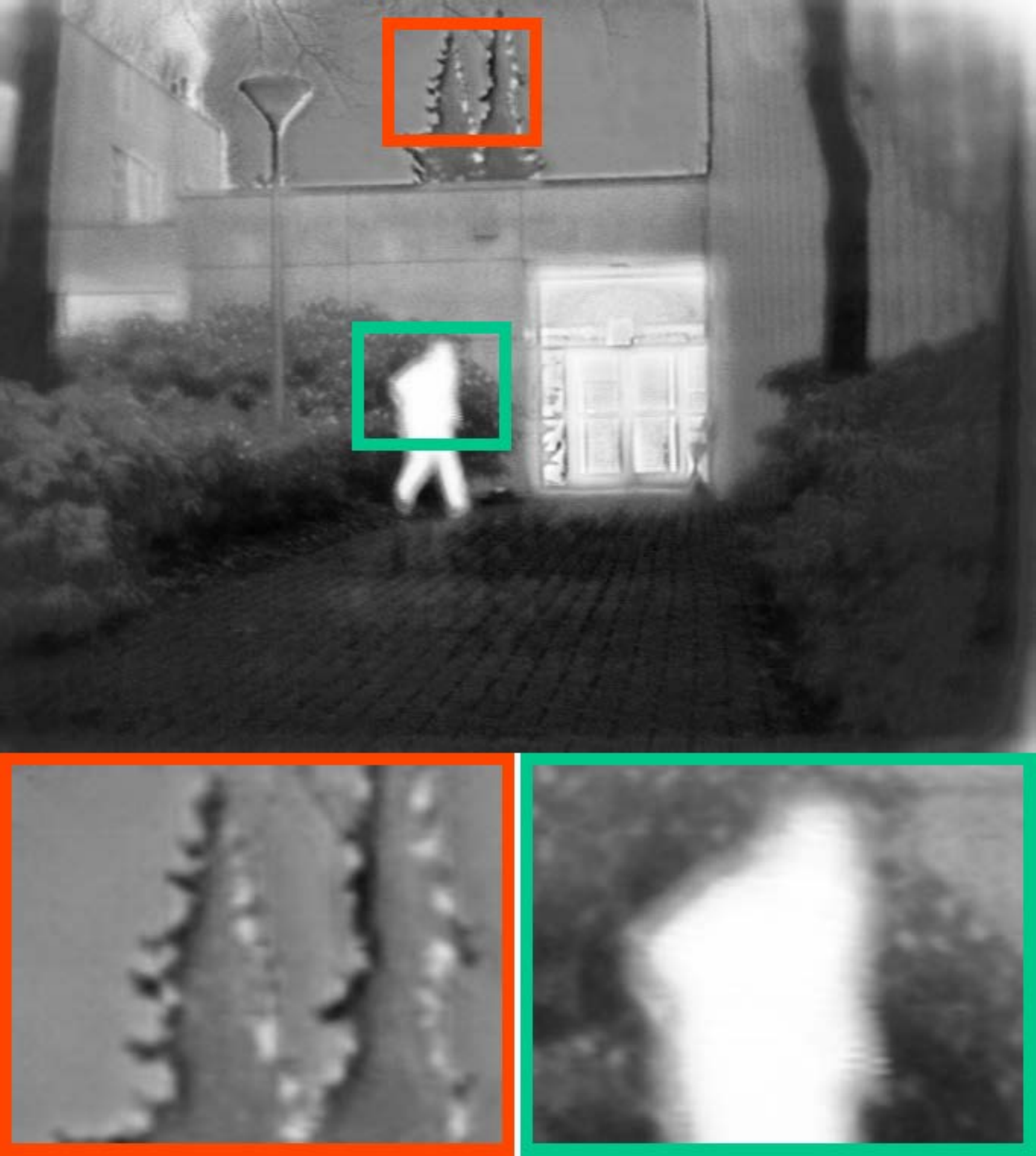}
		&\includegraphics[width=0.093\textwidth,height=0.075\textheight]{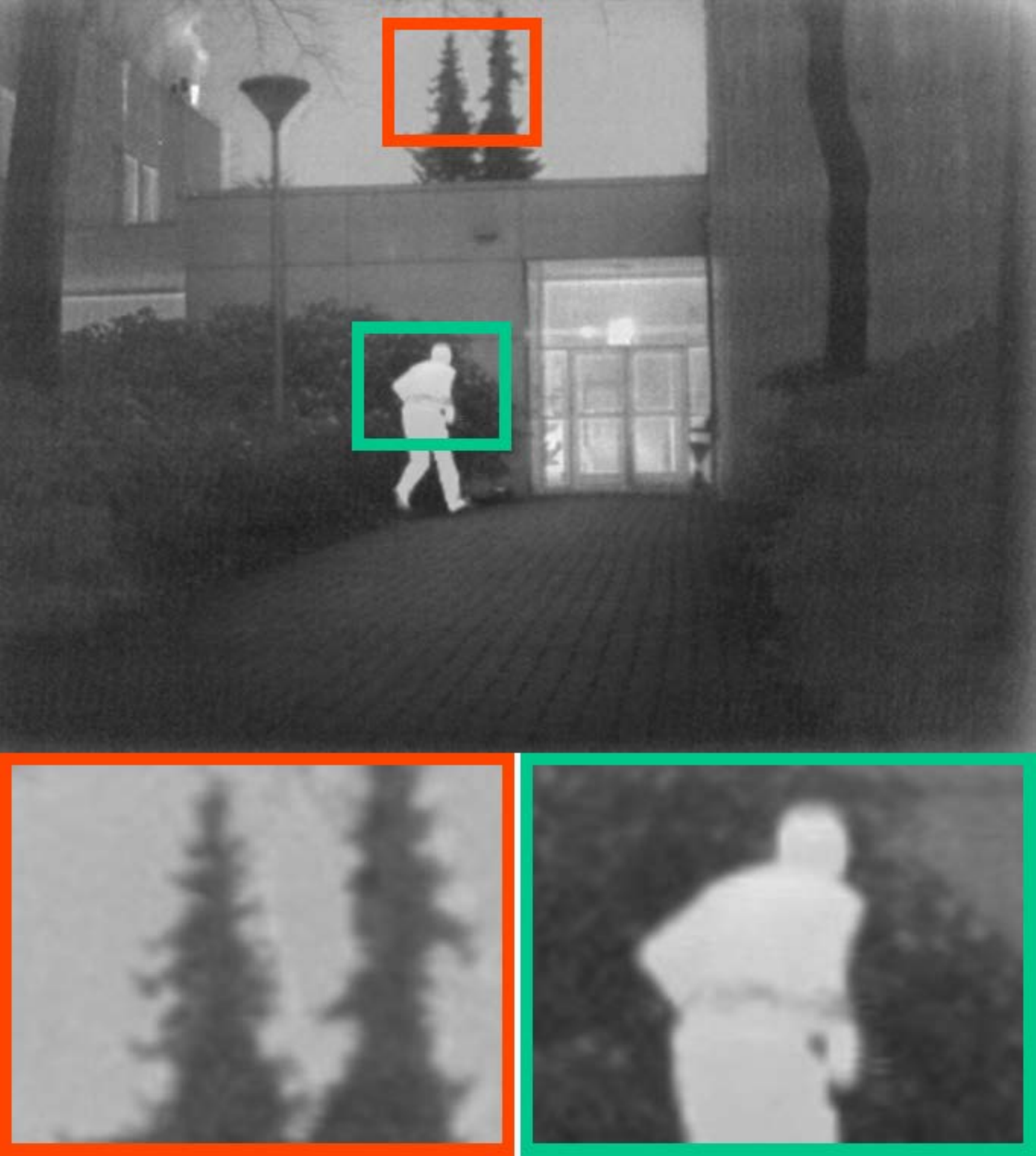}
		&\includegraphics[width=0.093\textwidth,height=0.075\textheight]{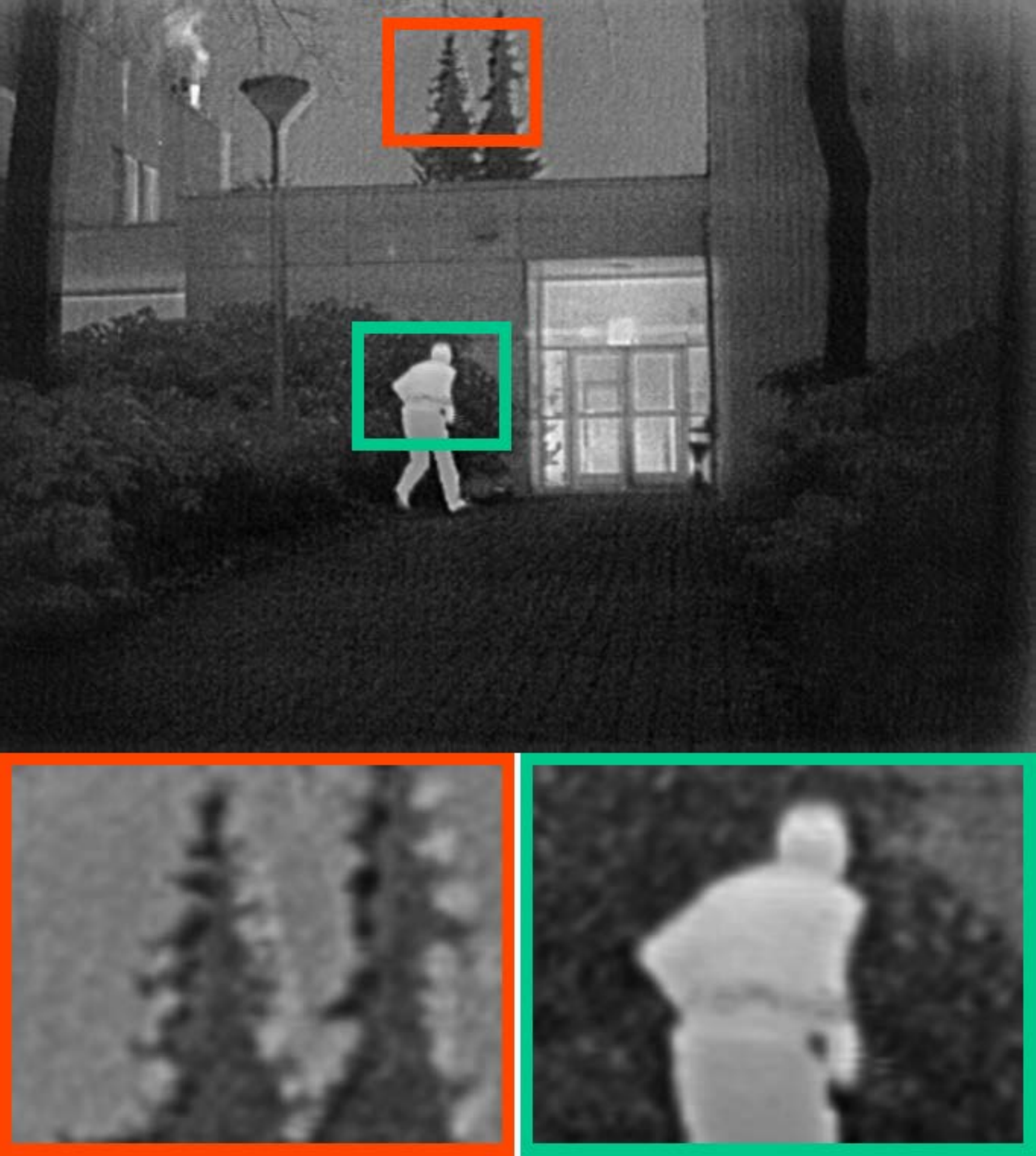}
		&\includegraphics[width=0.093\textwidth,height=0.075\textheight]{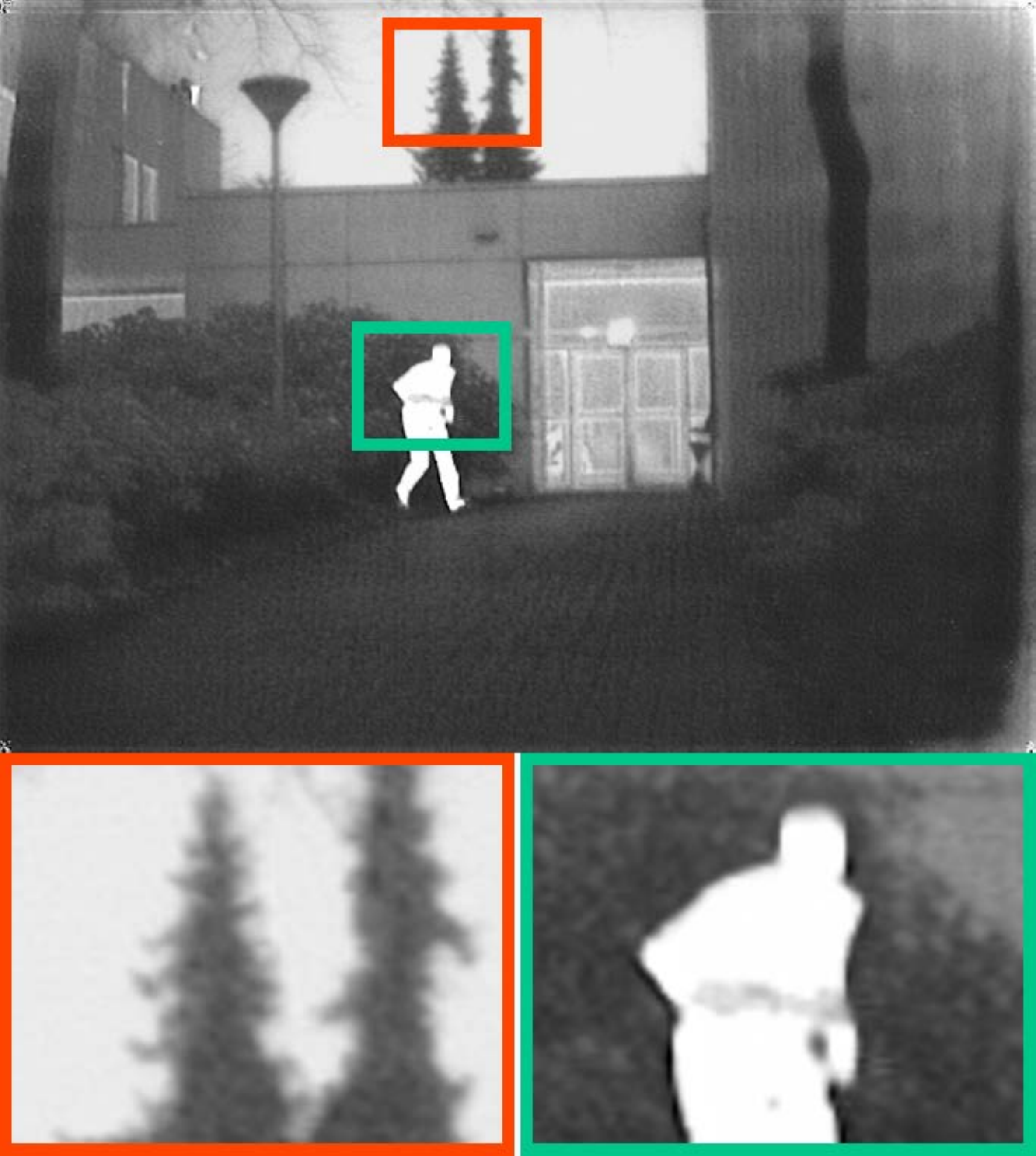}
		\\
		\includegraphics[width=0.093\textwidth,height=0.075\textheight]{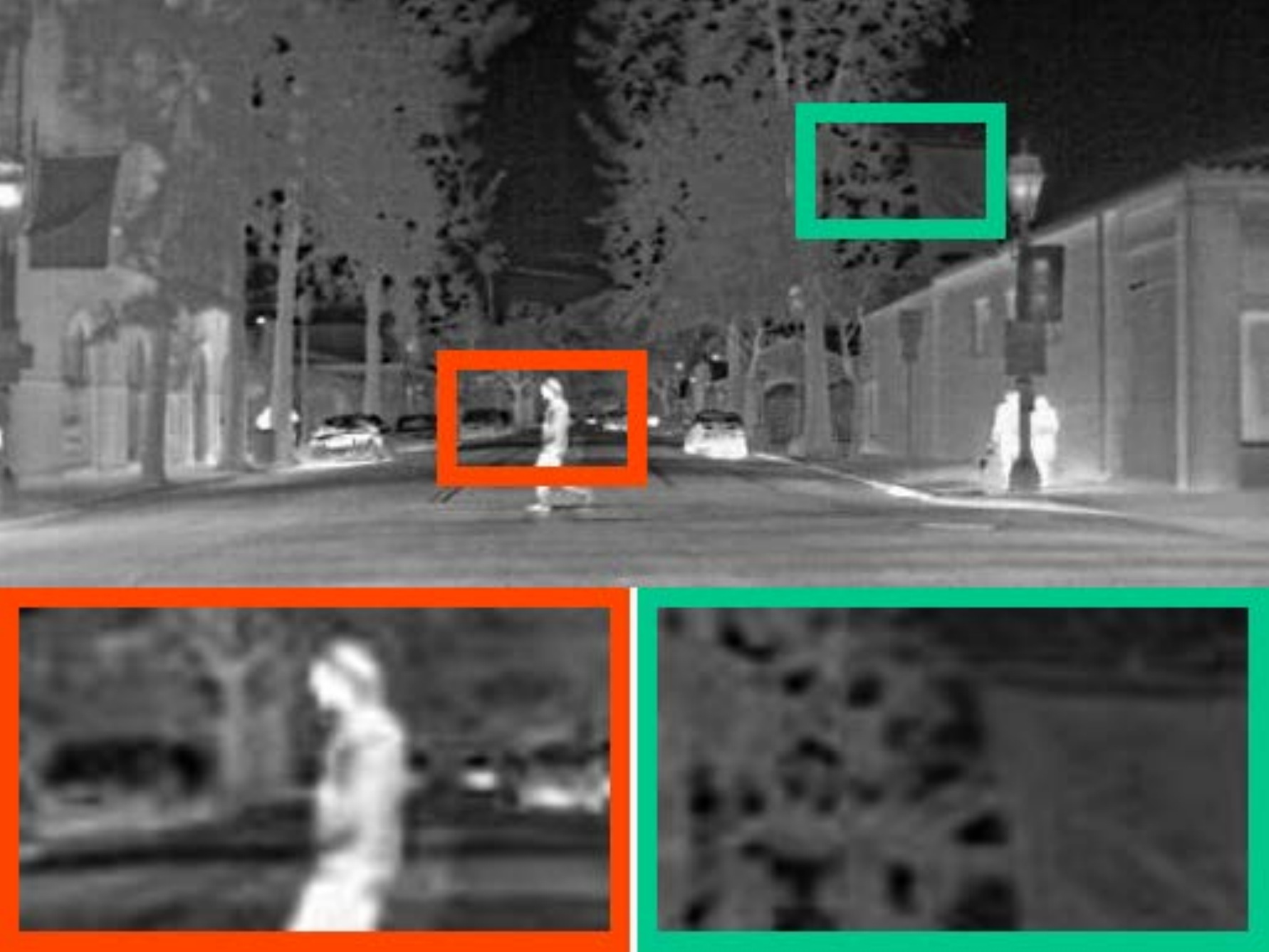}
		&\includegraphics[width=0.093\textwidth,height=0.075\textheight]{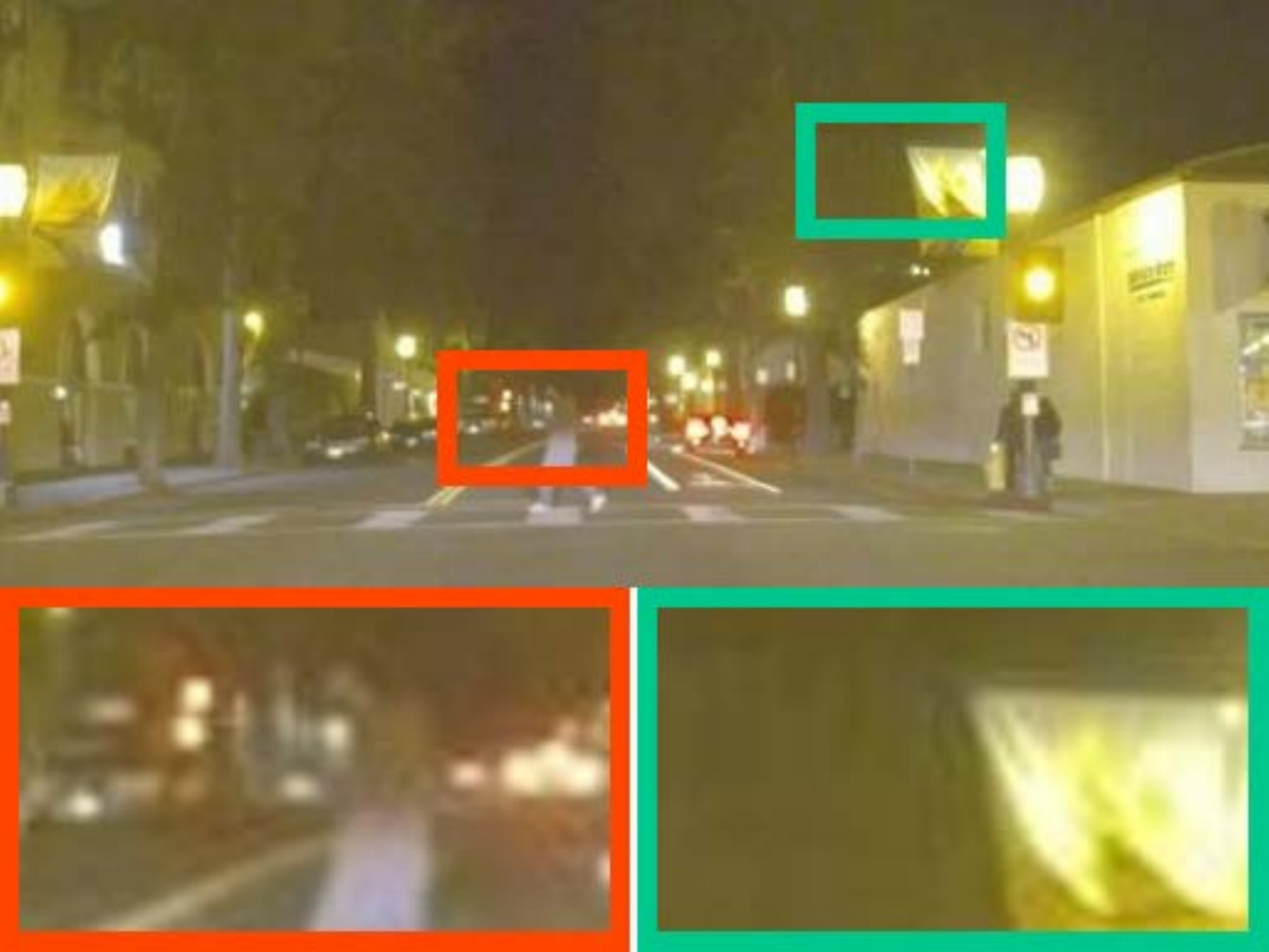}
		&\includegraphics[width=0.093\textwidth,height=0.075\textheight]{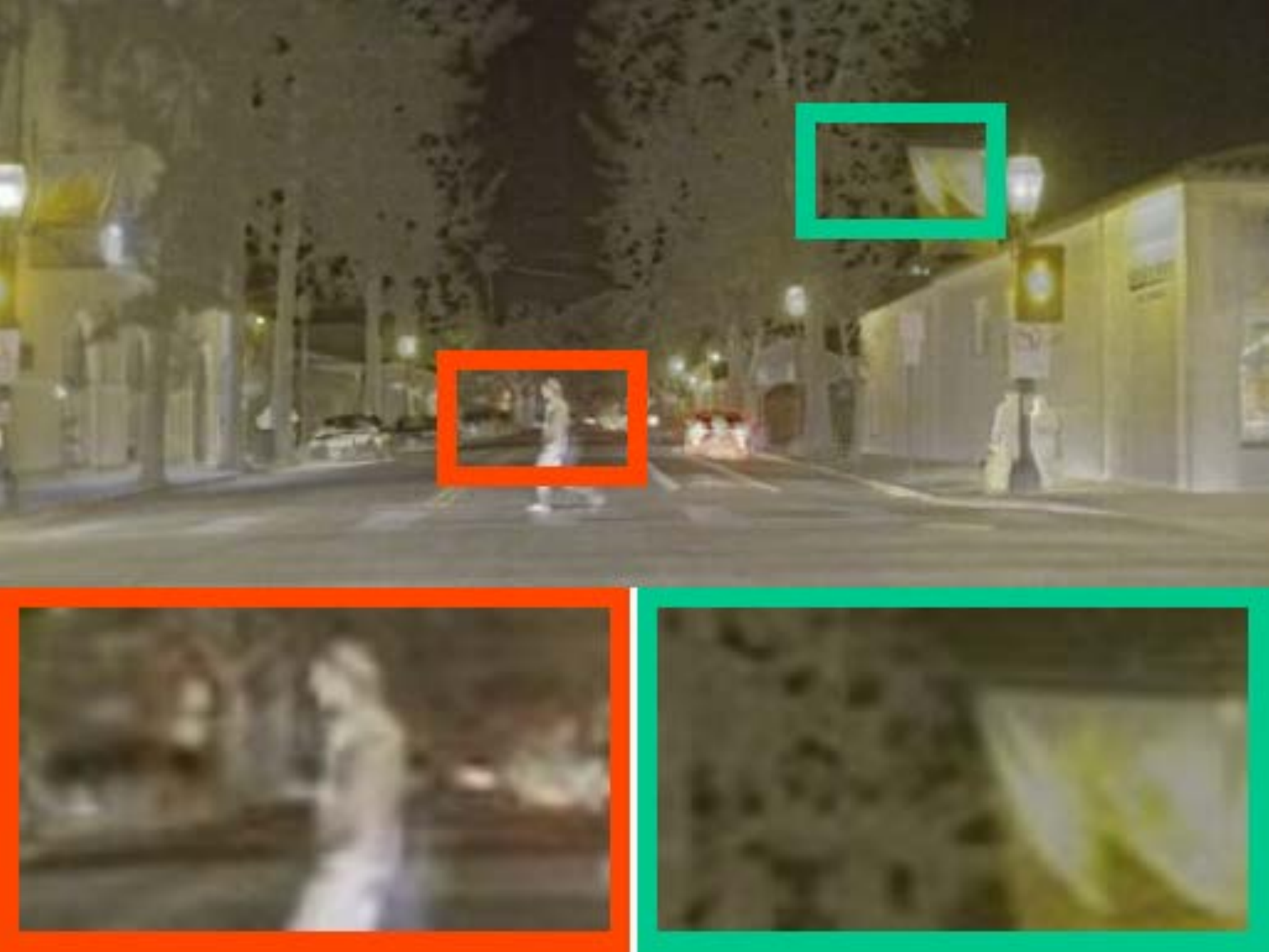}
		&\includegraphics[width=0.093\textwidth,height=0.075\textheight]{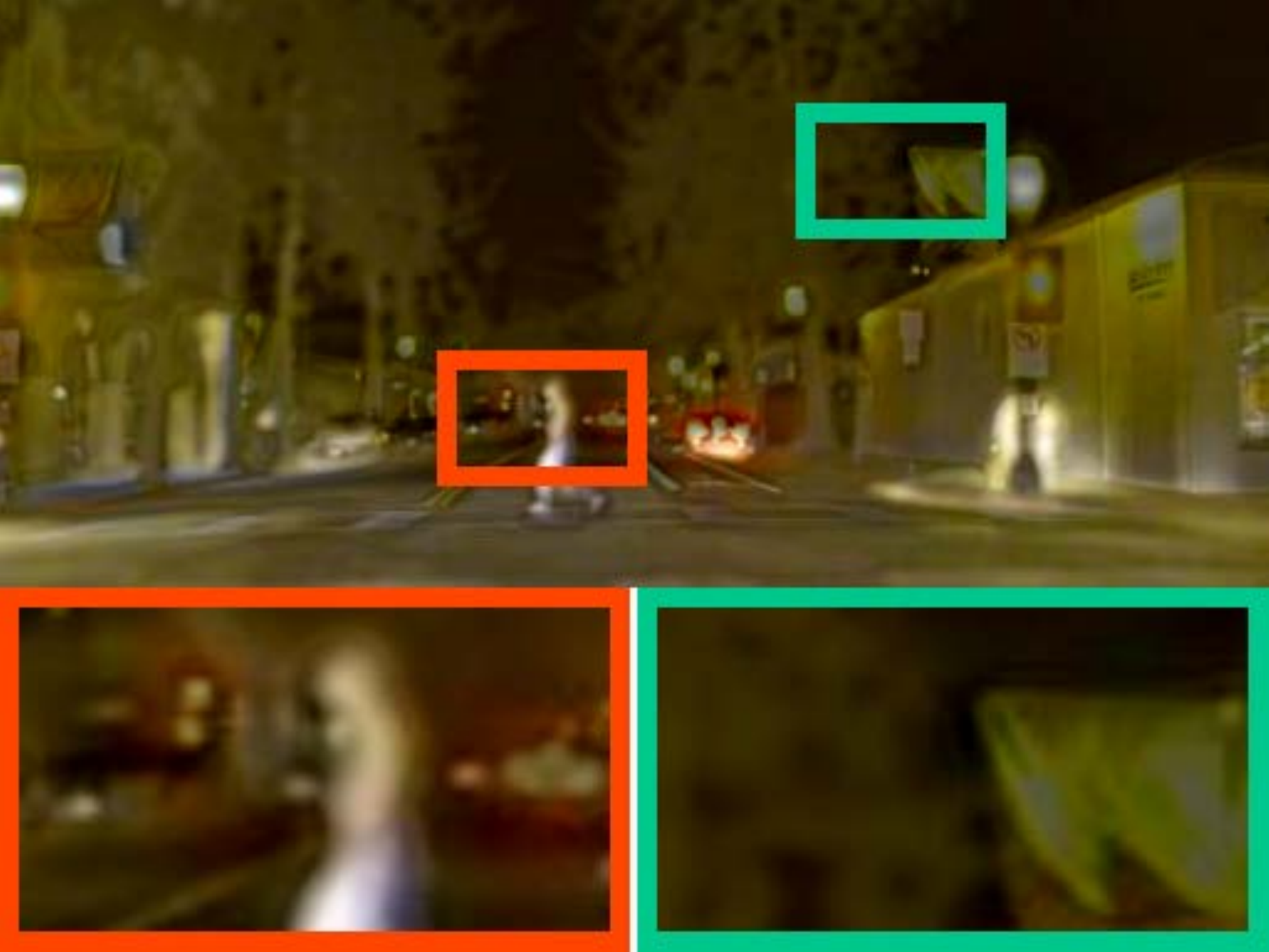}
		&\includegraphics[width=0.093\textwidth,height=0.075\textheight]{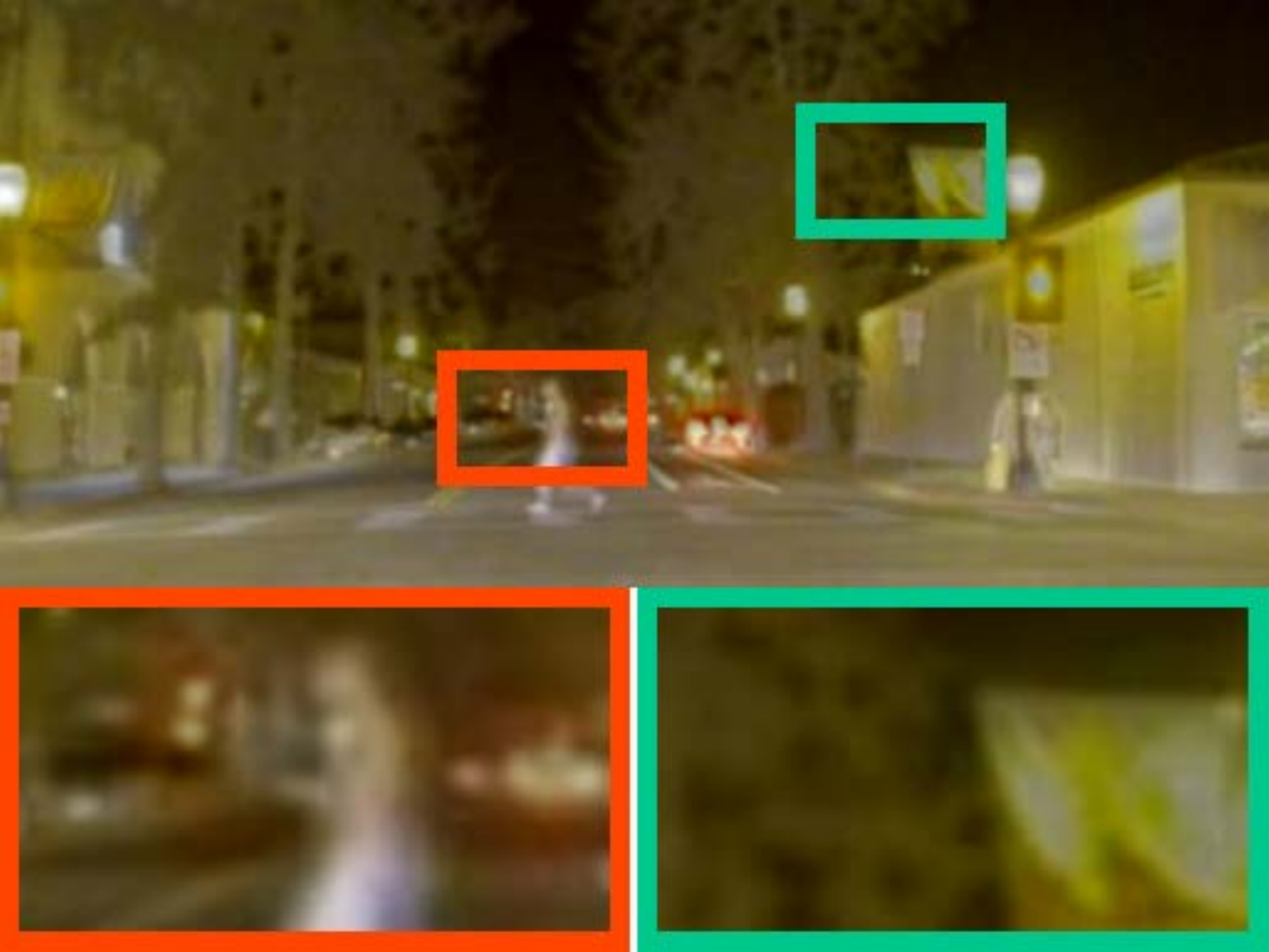}
		&\includegraphics[width=0.093\textwidth,height=0.075\textheight]{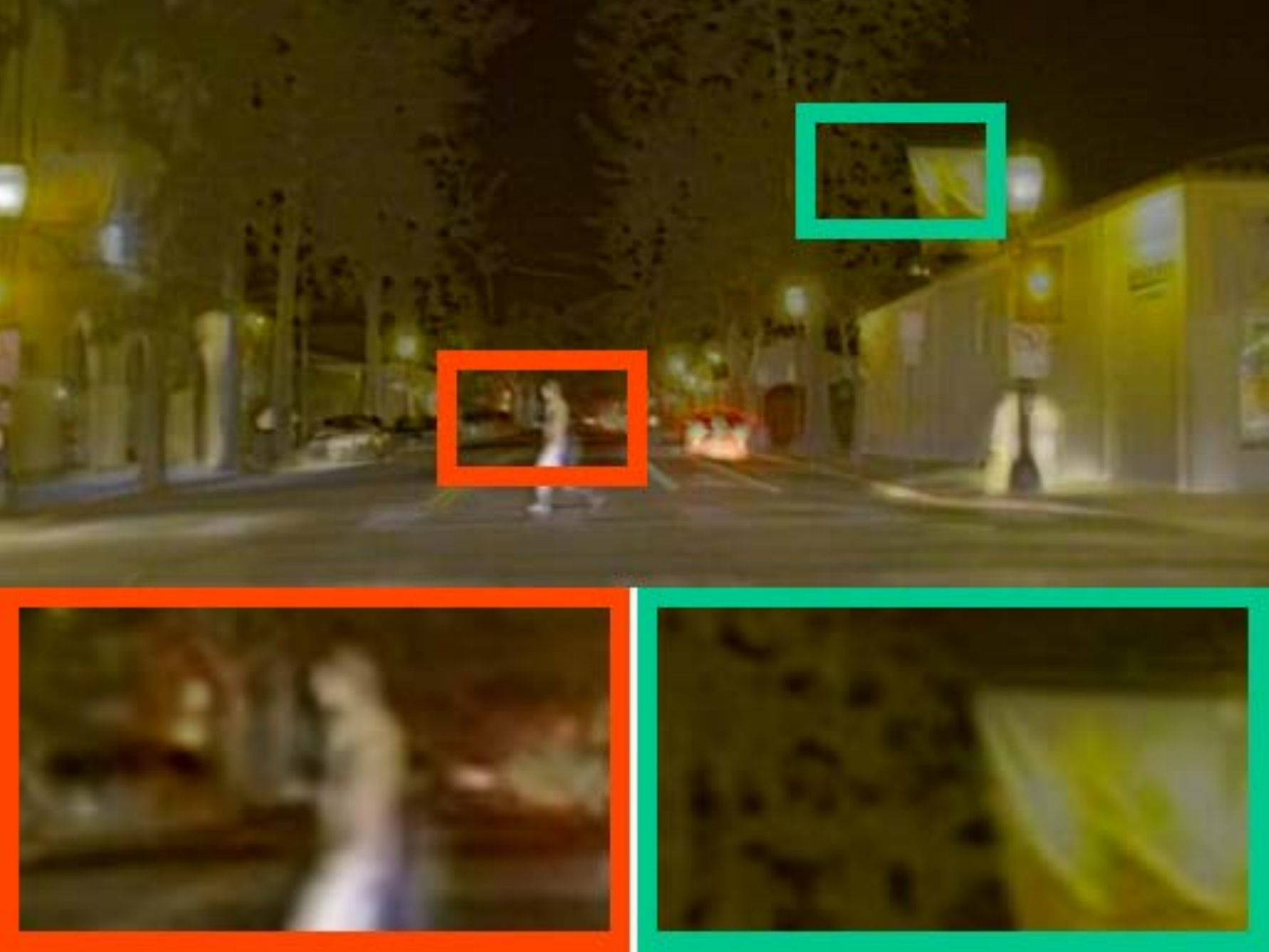}
		&\includegraphics[width=0.093\textwidth,height=0.075\textheight]{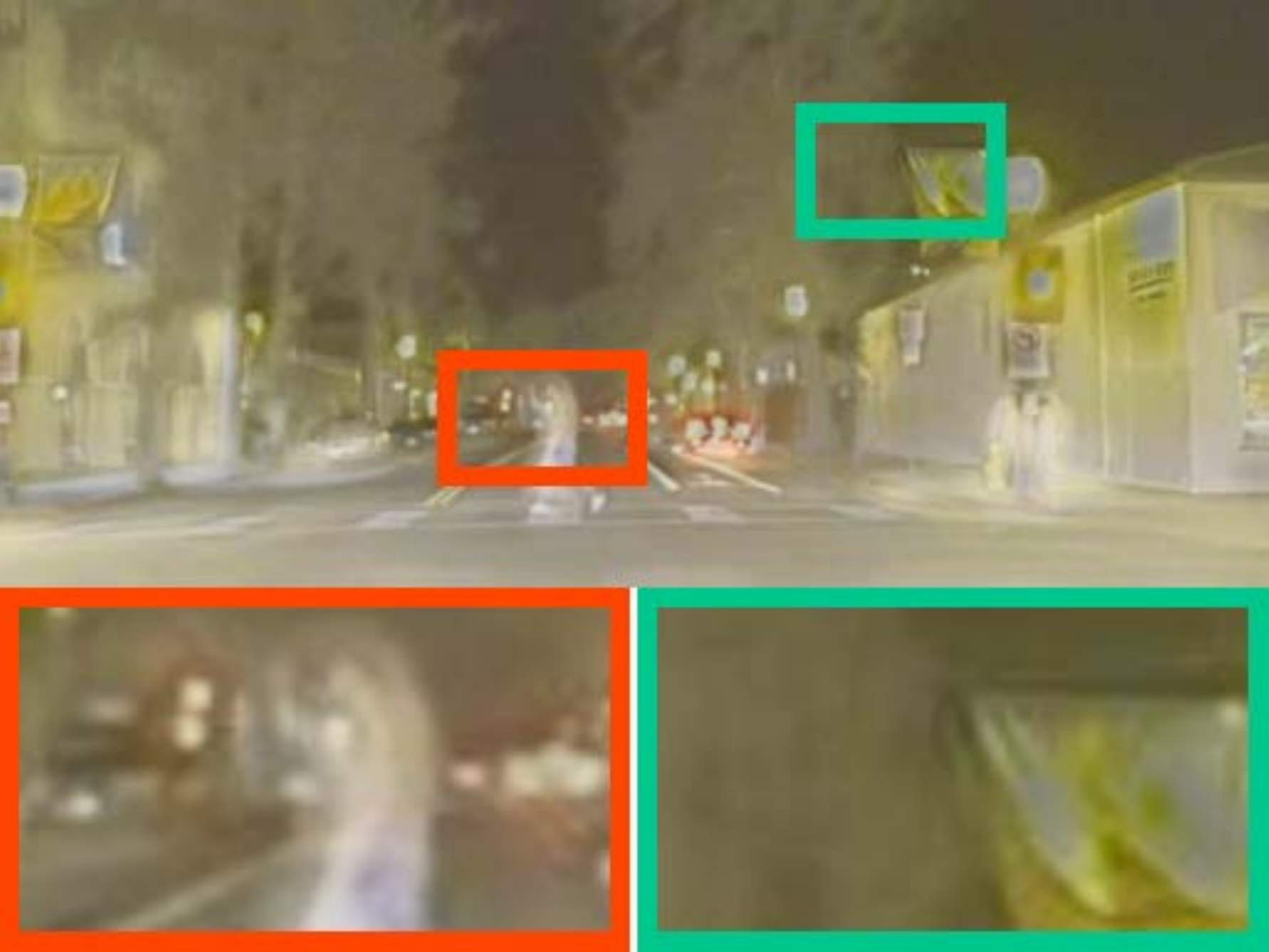}
		&\includegraphics[width=0.093\textwidth,height=0.075\textheight]{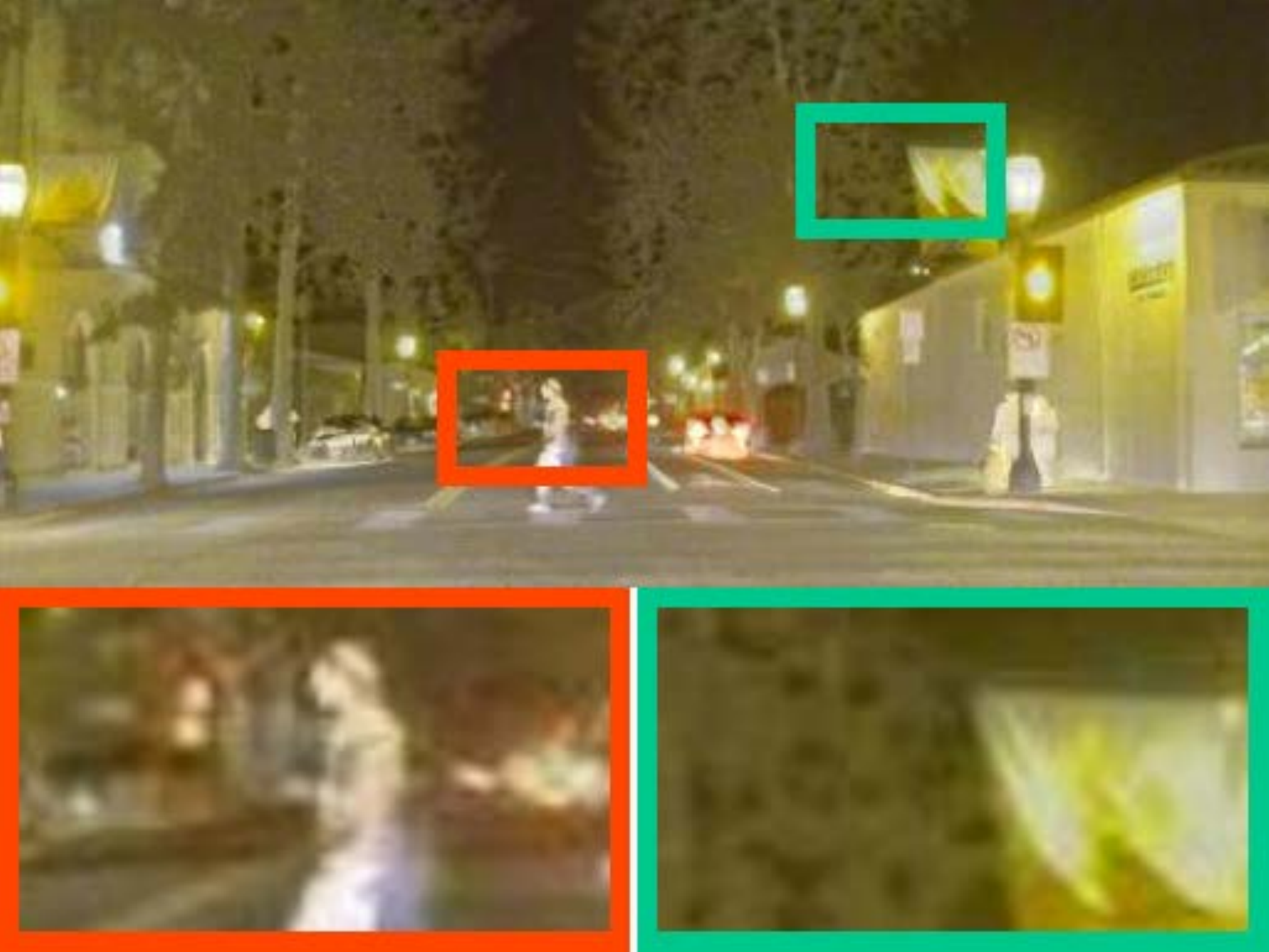}
		&\includegraphics[width=0.093\textwidth,height=0.075\textheight]{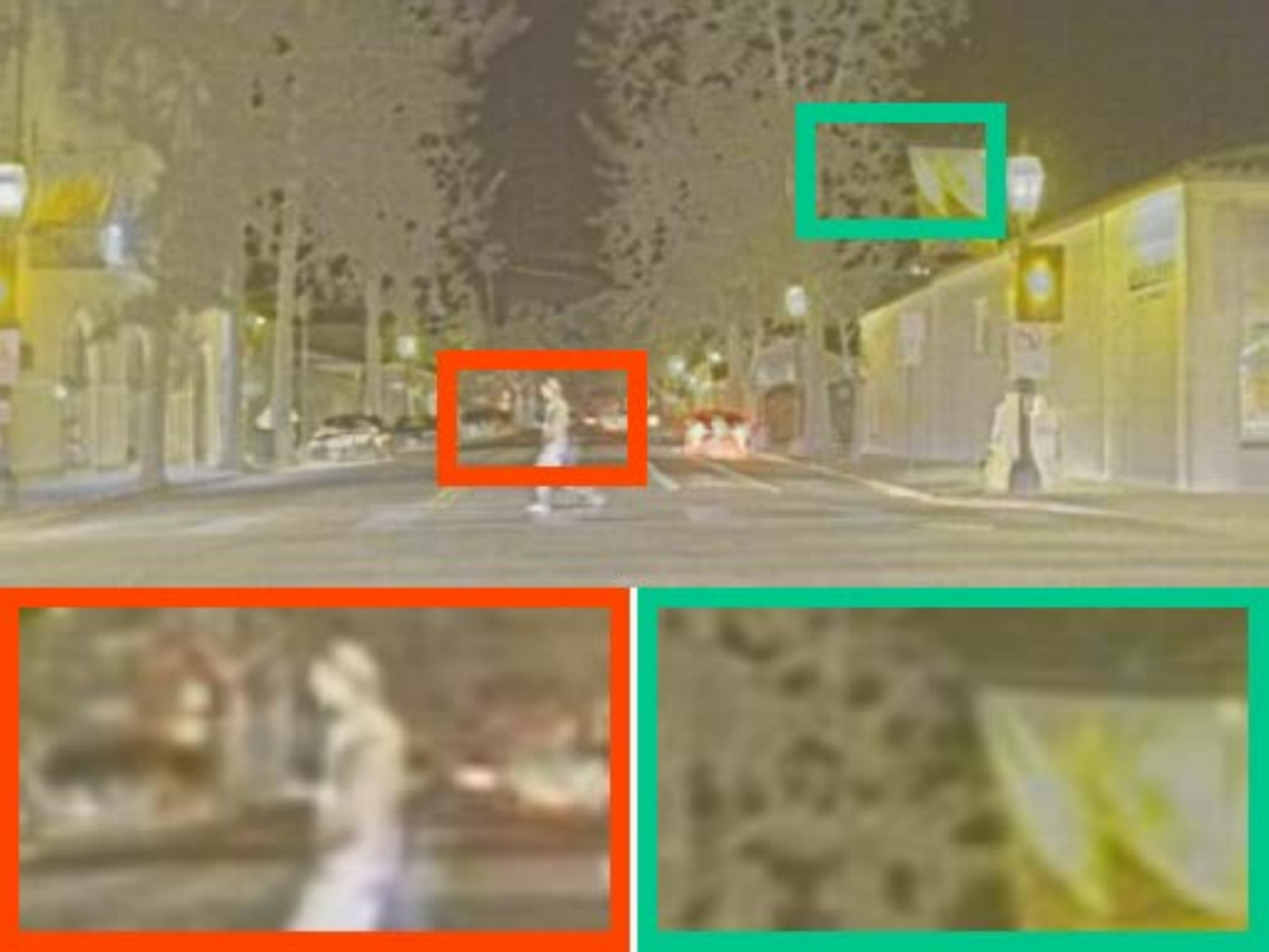}
		&\includegraphics[width=0.093\textwidth,height=0.075\textheight]{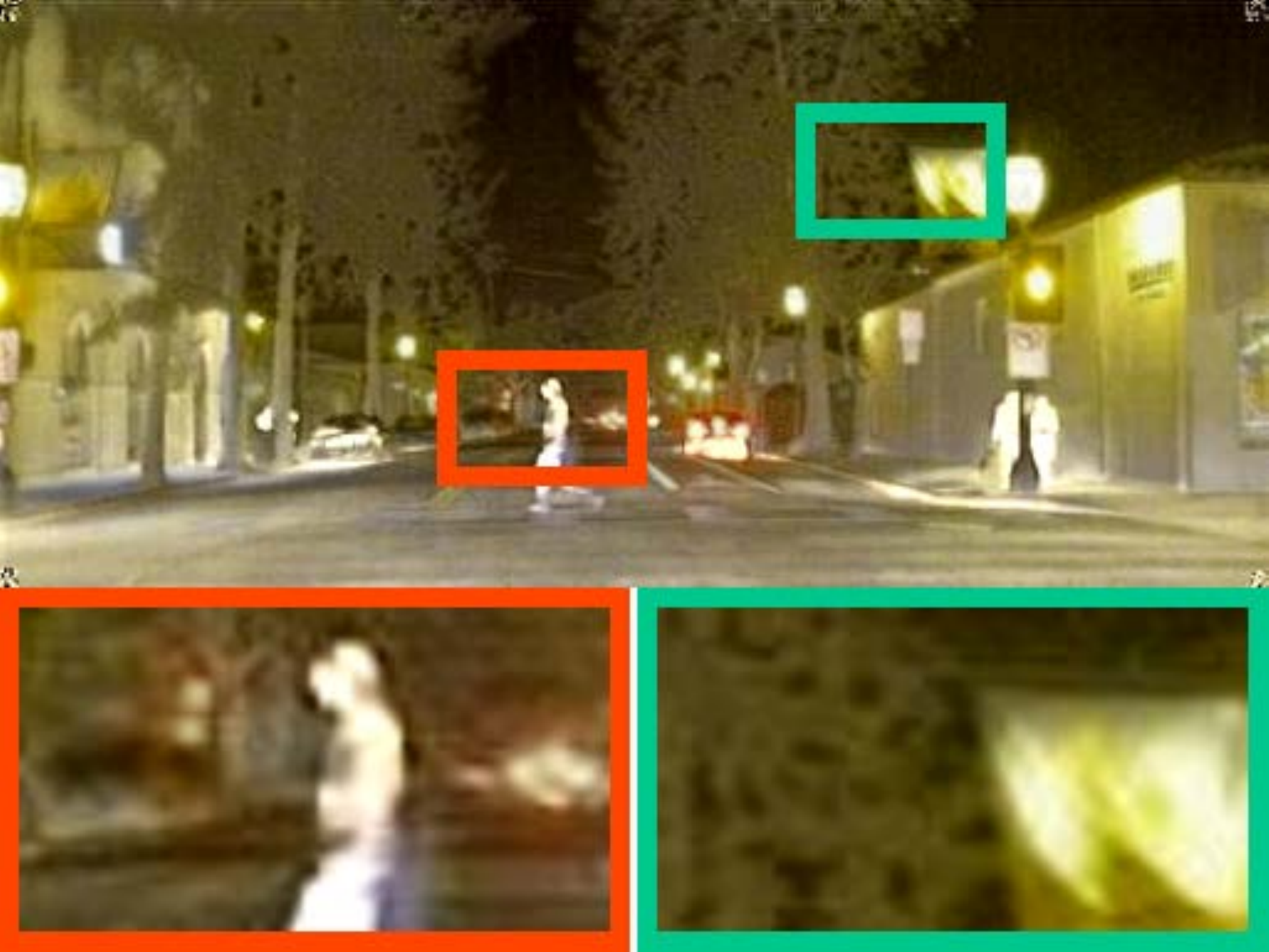}
		\\
		\includegraphics[width=0.093\textwidth,height=0.075\textheight]{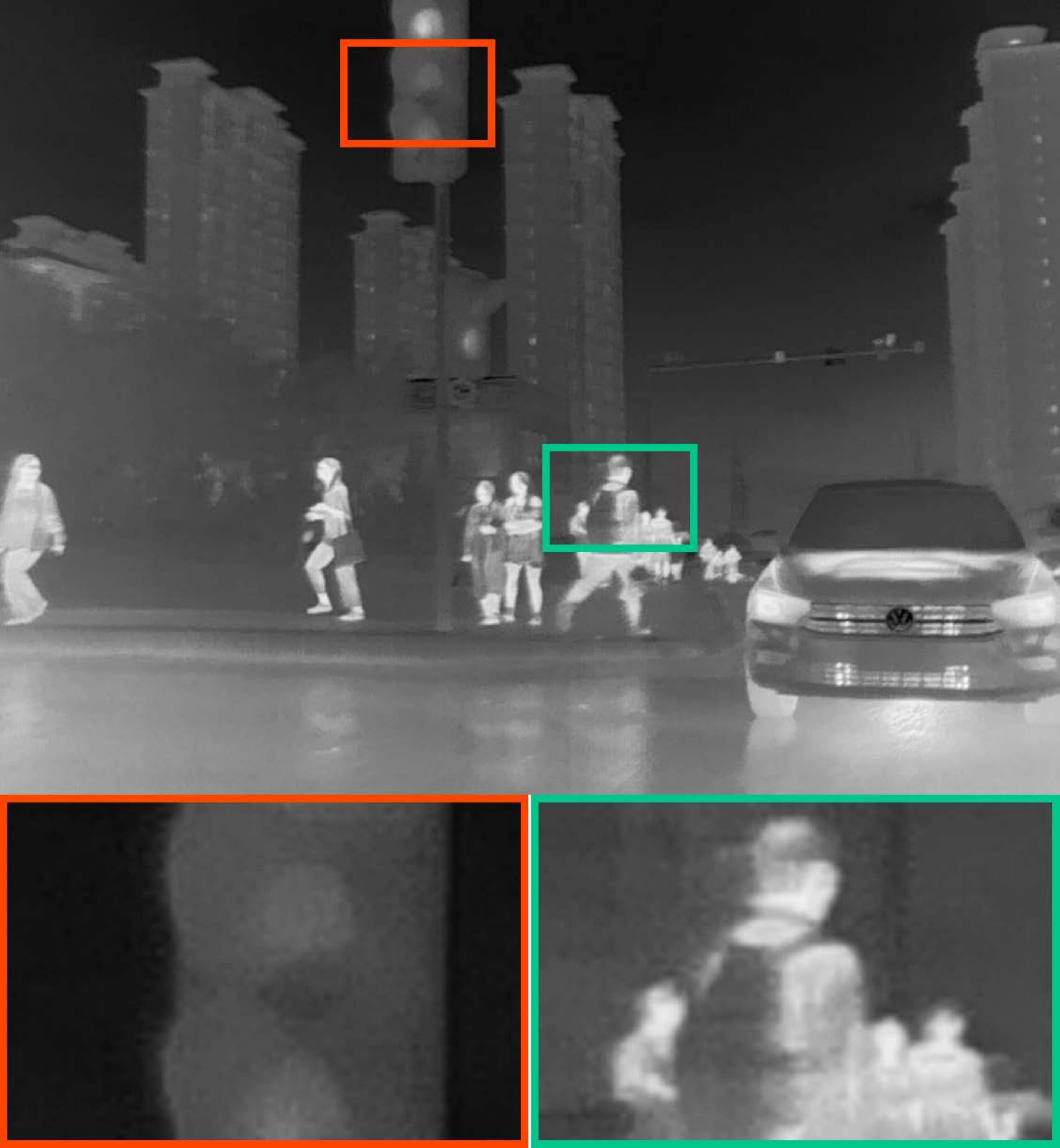}
		&\includegraphics[width=0.093\textwidth,height=0.075\textheight]{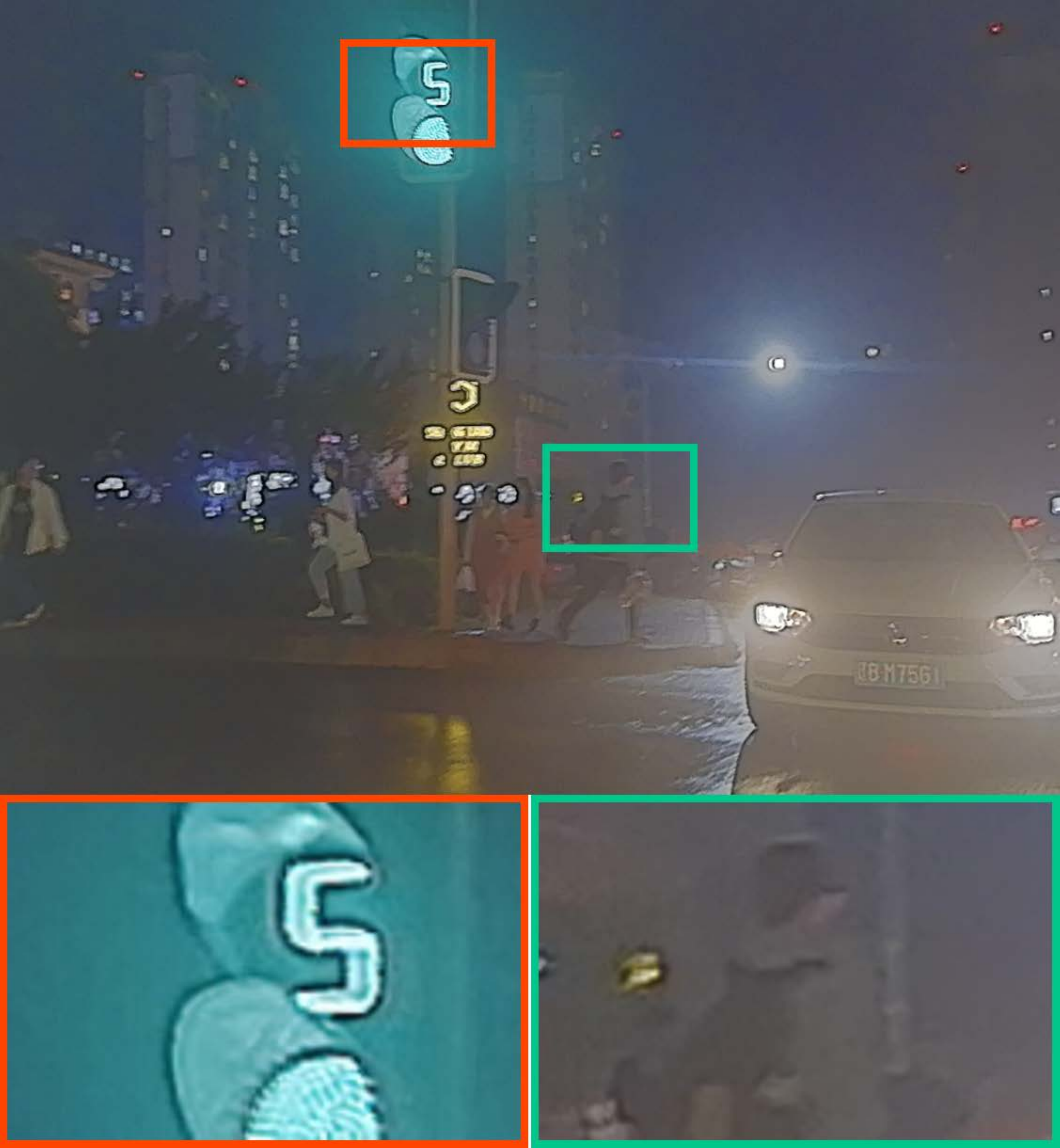}
		&\includegraphics[width=0.093\textwidth,height=0.075\textheight]{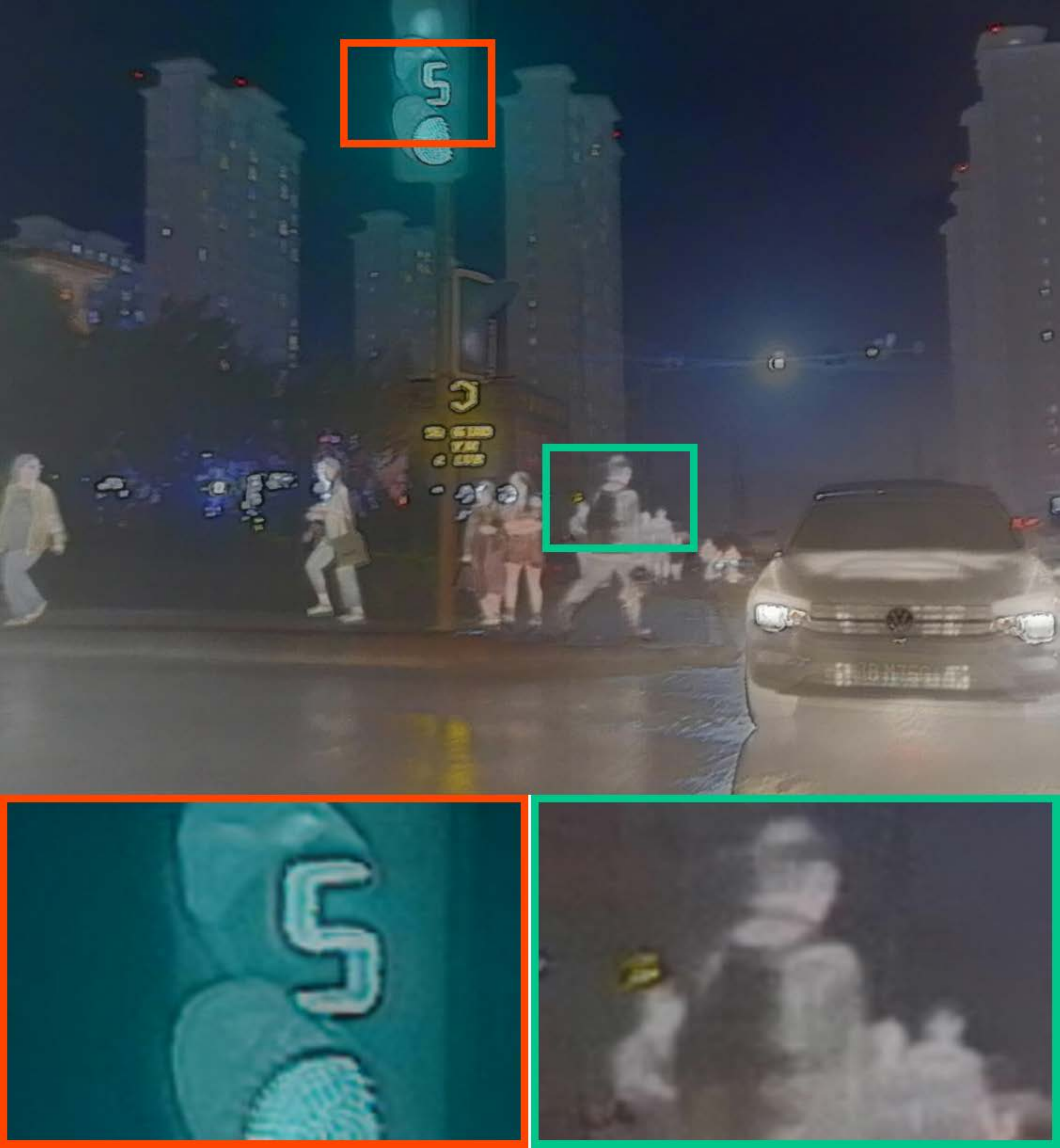}
		&\includegraphics[width=0.093\textwidth,height=0.075\textheight]{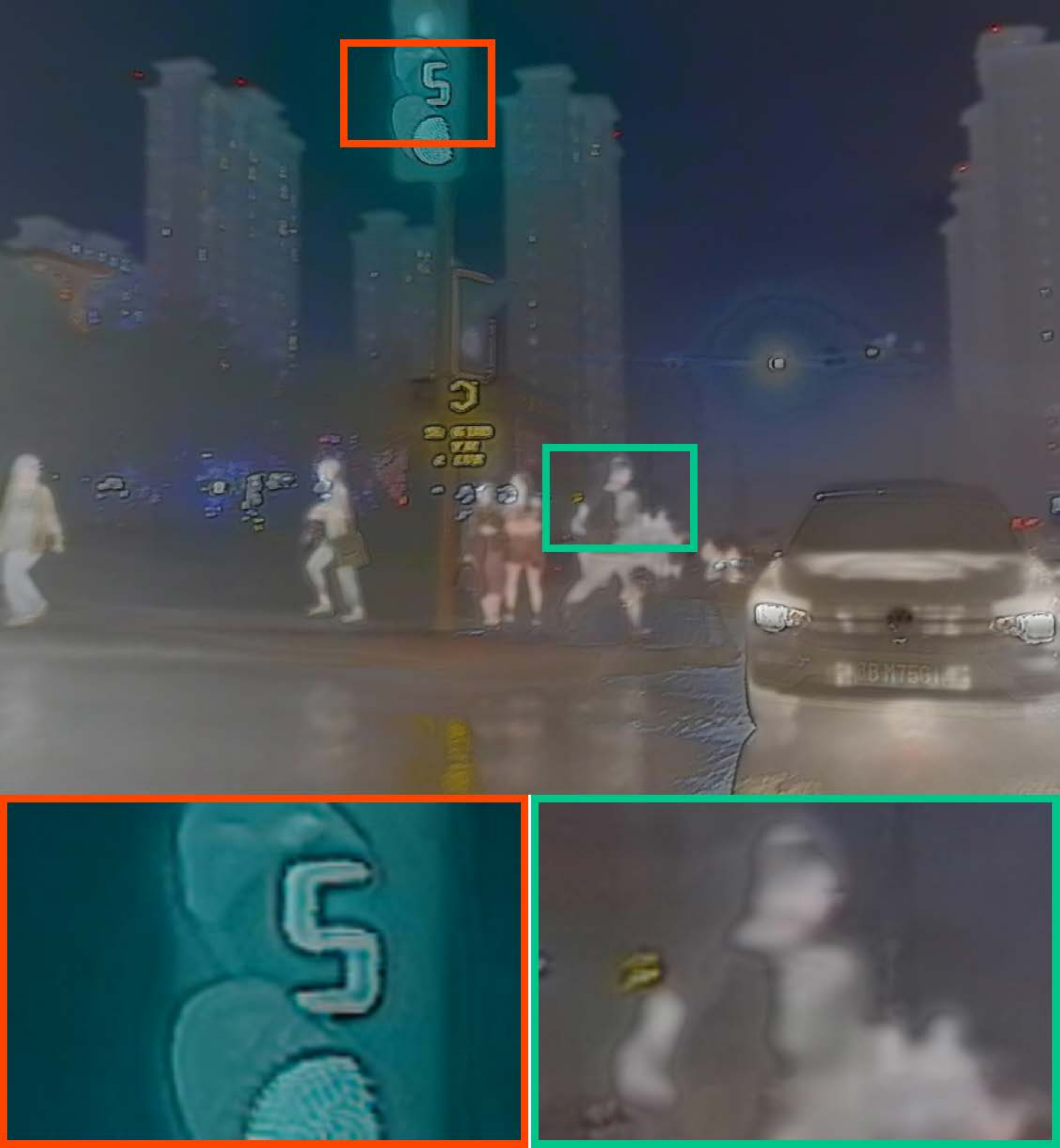}
		&\includegraphics[width=0.093\textwidth,height=0.075\textheight]{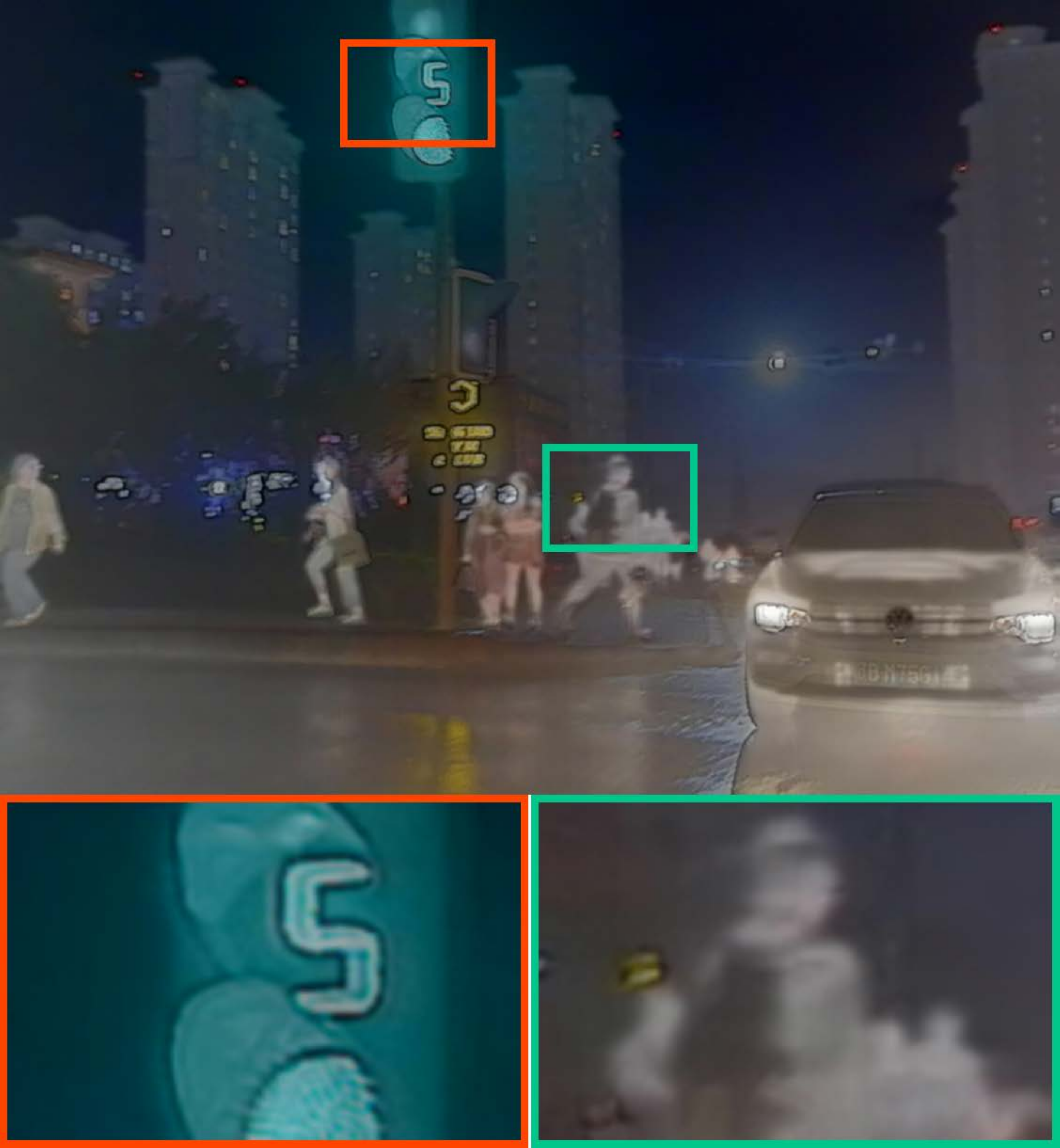}
		&\includegraphics[width=0.093\textwidth,height=0.075\textheight]{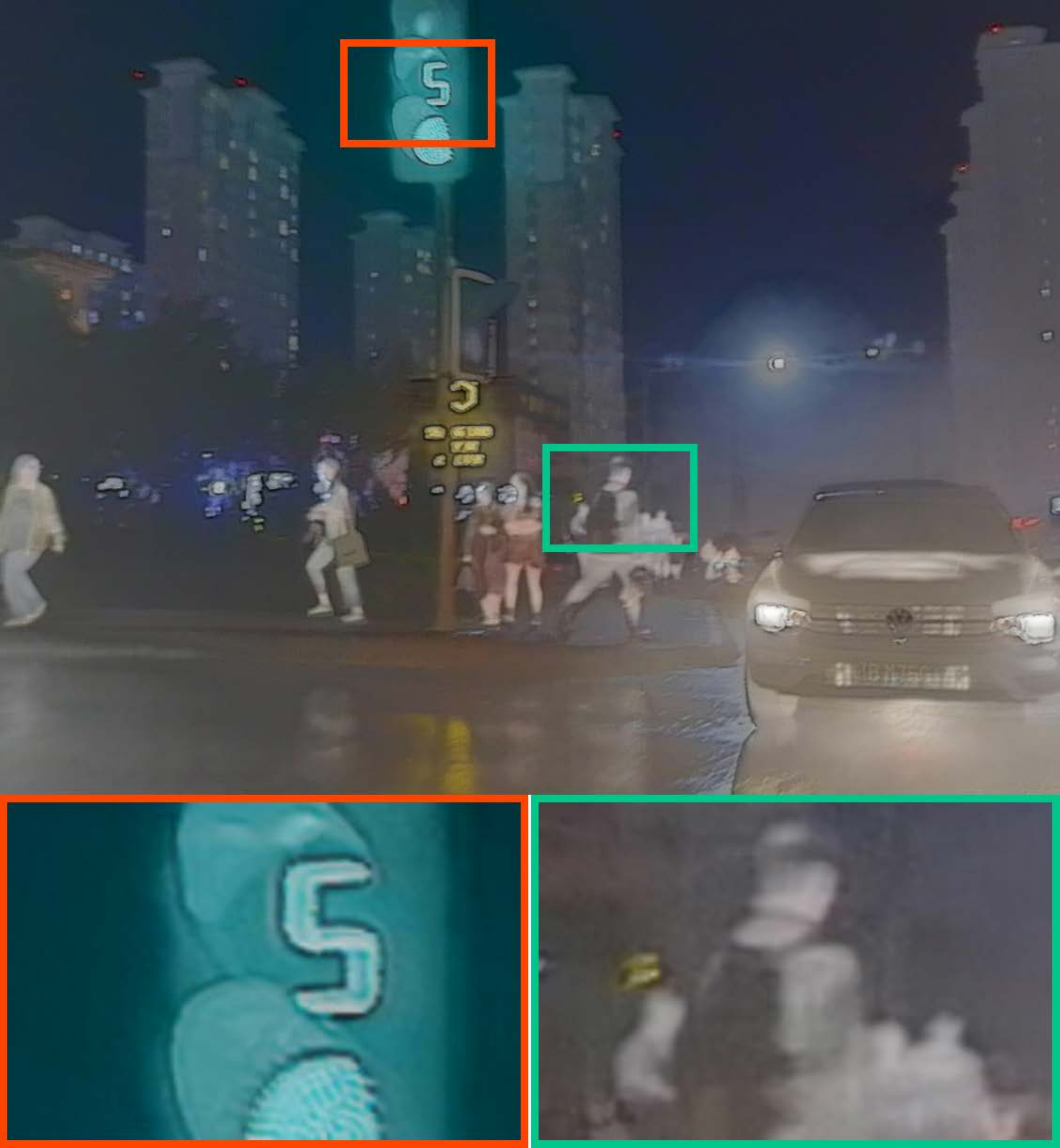}
		&\includegraphics[width=0.093\textwidth,height=0.075\textheight]{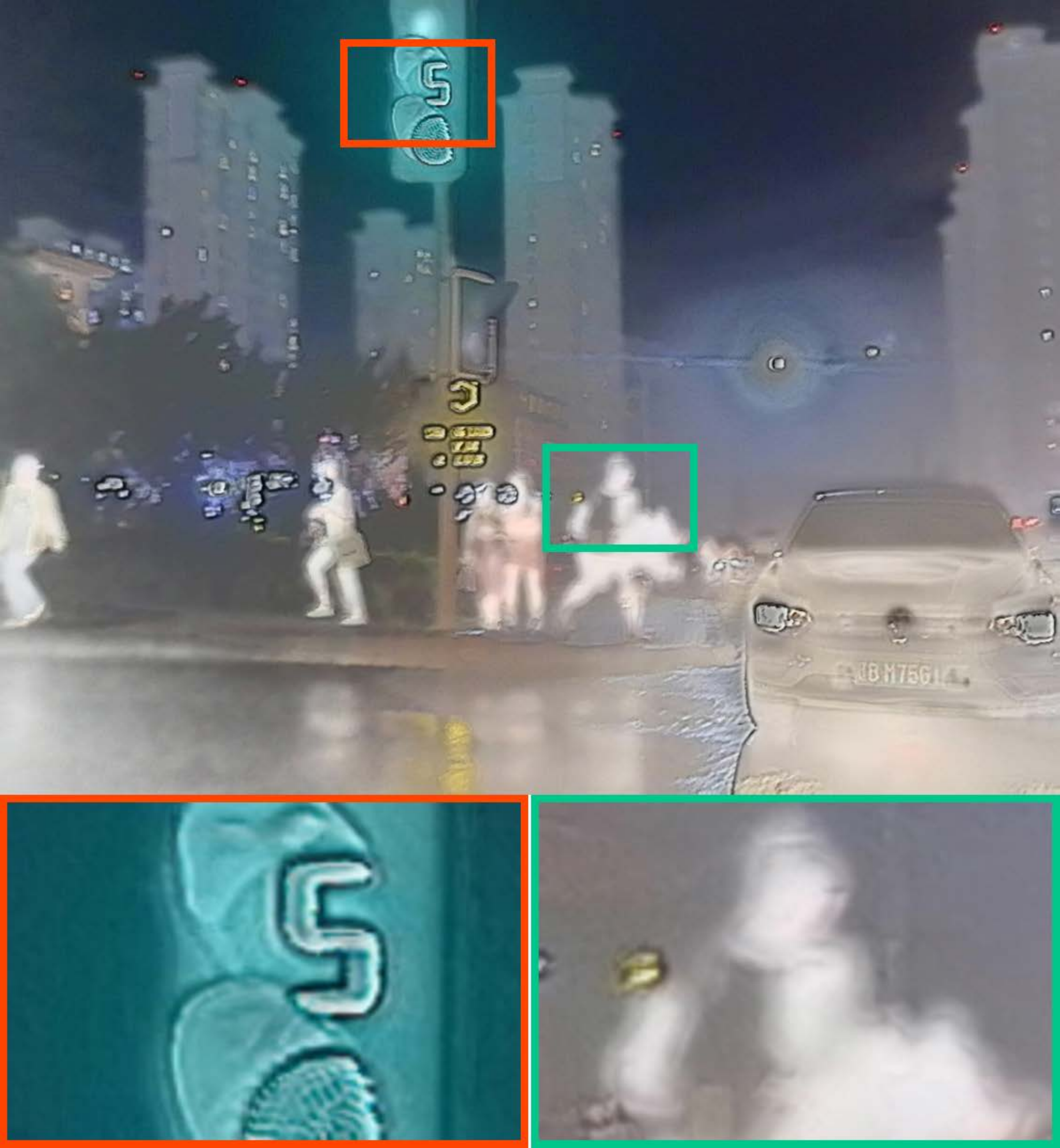}
		&\includegraphics[width=0.093\textwidth,height=0.075\textheight]{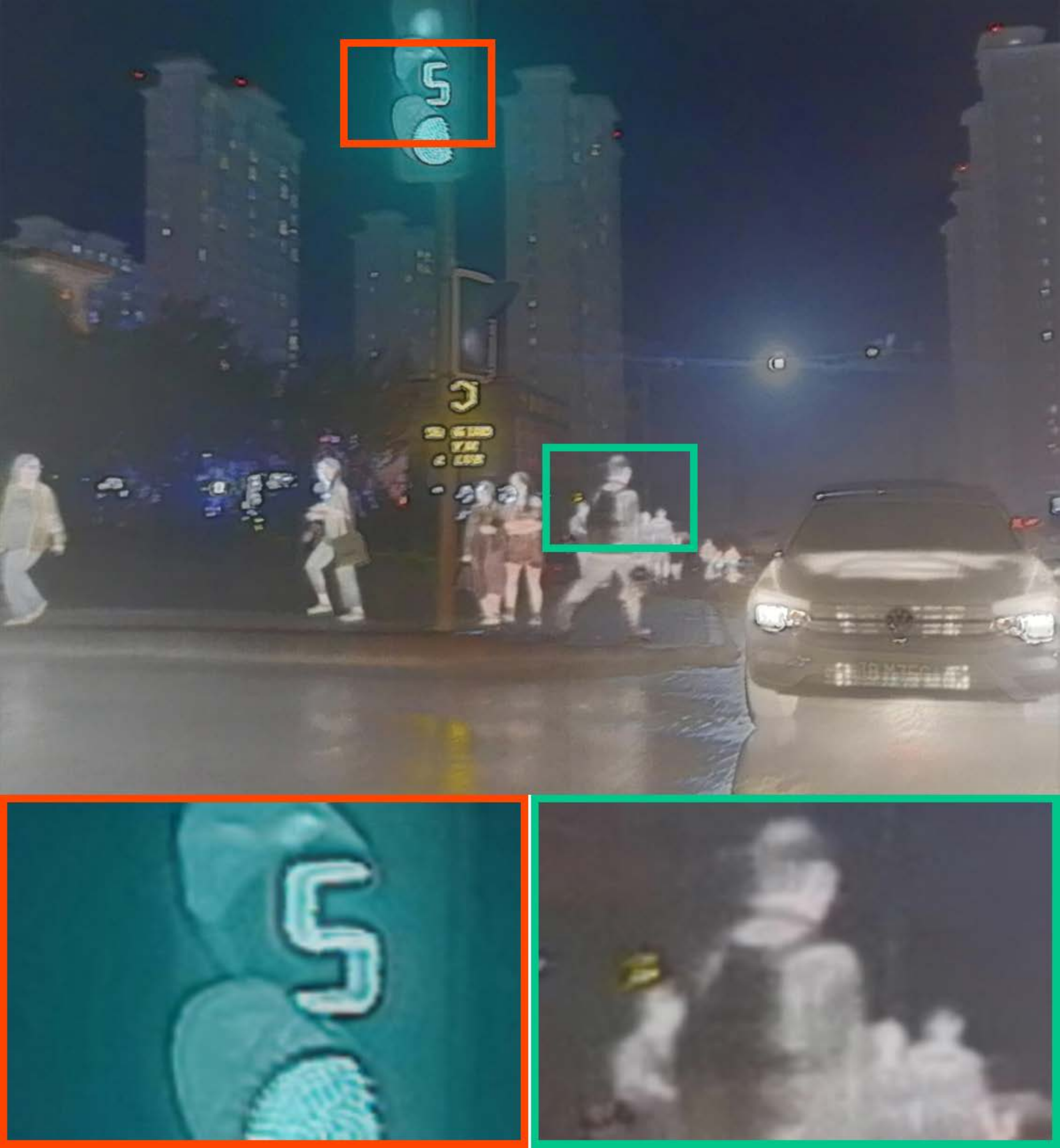}
		&\includegraphics[width=0.093\textwidth,height=0.075\textheight]{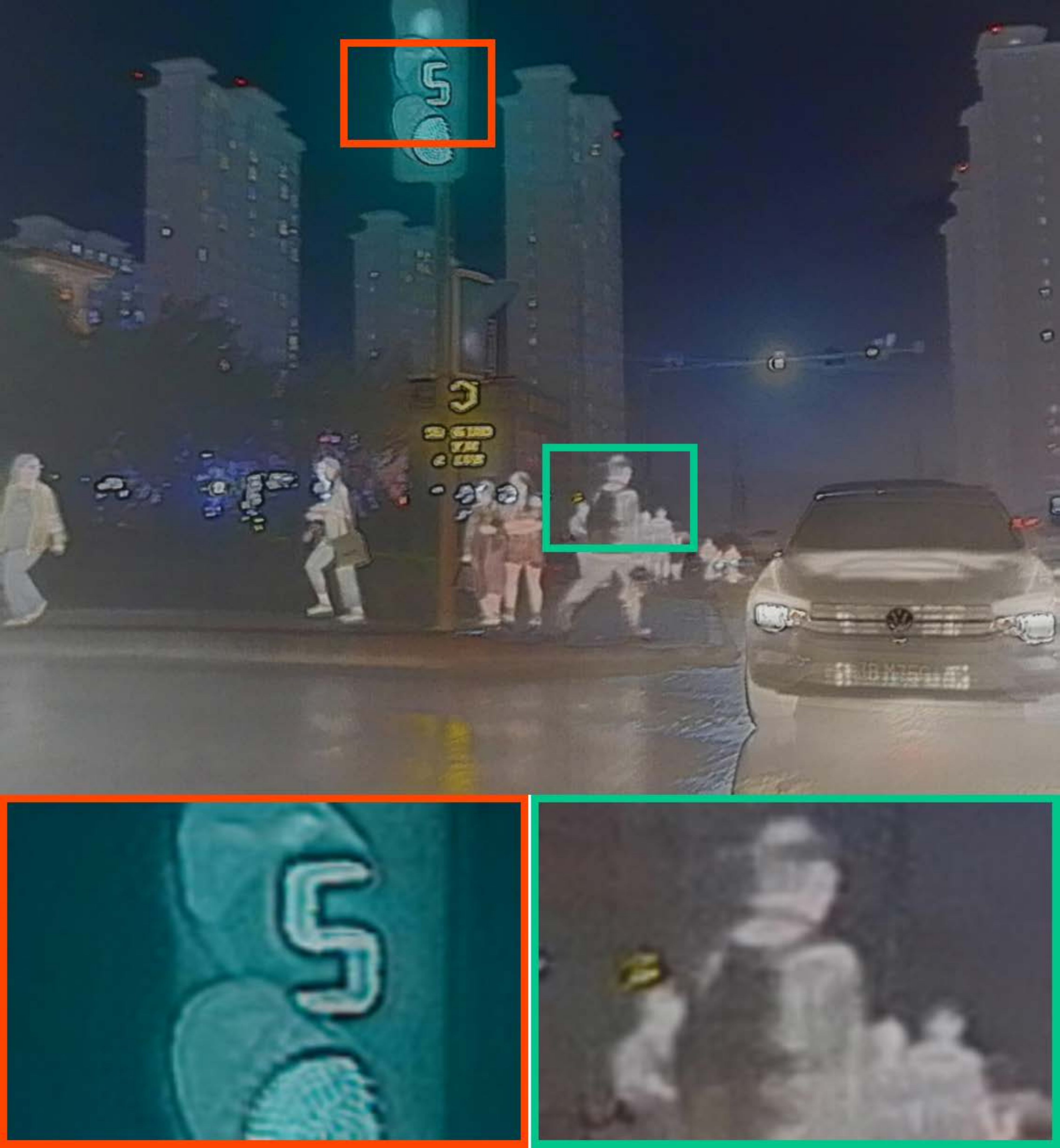}
		&\includegraphics[width=0.093\textwidth,height=0.075\textheight]{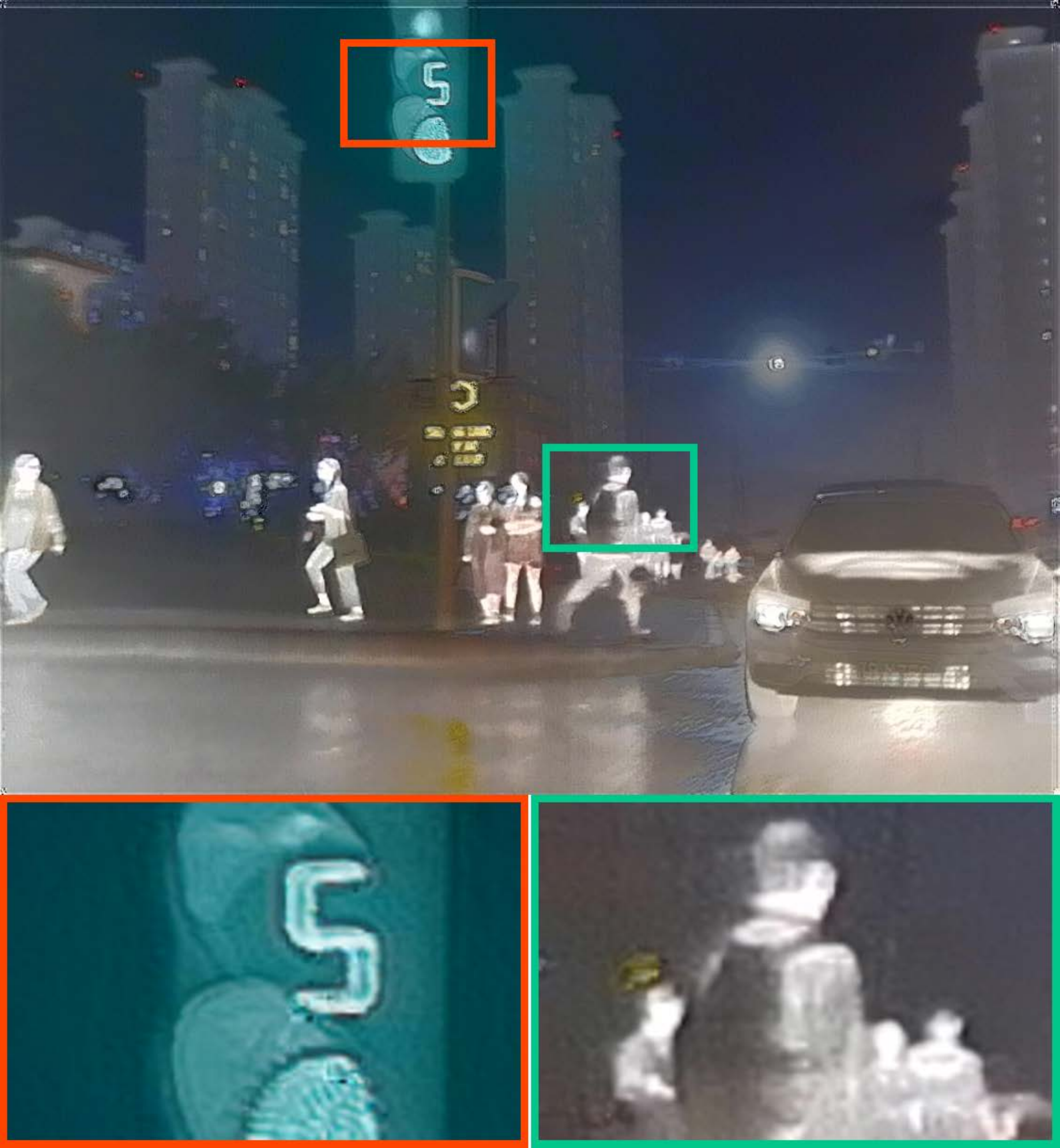}
		\\
		\footnotesize Ir image&\footnotesize Vis image&\footnotesize DenseFuse&\footnotesize FusionGAN&\footnotesize RFN&\footnotesize GANMcC&\footnotesize DDcGAN&\footnotesize MFEIF&\footnotesize U2Fusion&\footnotesize TarDAL	
		\\			
	\end{tabular}
	\caption{Visual comparisons of our TarDAL with state-of-the-art methods on typical image pairs in TNO, RoadScene and M$^3$FD datasets. }
	\label{fig:Visual}
\end{figure*}

\begin{figure*}[!htb]
	\centering
	\setlength{\tabcolsep}{1pt} 
	\begin{tabular}{c}		
		\includegraphics[width=0.98\textwidth,height=0.115\textheight]{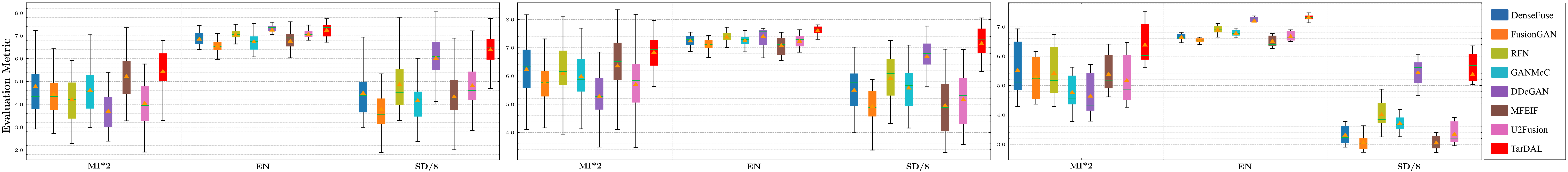}	
	\end{tabular}
	\vspace{-0.3cm}  
	\caption{Quantitative comparisons with seven IVIF methods on TNO, RoadScene and M$^3$FD datasets, respectively. The x-axis represents metrics and the y-axis are the values.~($*$). In the boxes, the orange lines and the green tangles denote medium and mean values. }
	\label{fig:numcpr}
\end{figure*} 
\noindent\textbf{Qualitative Comparisons} The intuitive qualitative results on three typical image pairs from three datasets are shown in Figure~\ref{fig:Visual}. Compared with other existing methods, our TarDAL has two significant advantages. First, the discriminative target from infrared images can be well preserved. As shown in Figure~\ref{fig:Visual}~(the green tangles of the second group), the people in our method exhibits high contrast and distinctive prominent contour, so that it is benefit to visual observation . Second, our results can preserve abundant textural details from visible images~(the green tangles of the first and third group), which are more in line with human visual system. In contrast, visual inspection shows that DenseFuse, and FusionGAN cannot highlight the discriminative targets well, while GANMcC and DDcGAN fail to obtain rich textural details. Note that our TarDAL is able to generate more visual-friendly fused results with clear targets, sharper edge contours, and preserve abundant textural details. 

\noindent\textbf{Quantitative Comparisons} Subsequently, we compare our TarDAL with the above-mentioned competitors quantitatively on 400 image pairs~(~20 image pairs from TNO, 40 image pairs from the RoadScene, and 340 image pairs from M$^3$FD). Besides, three evaluation metrics, \emph{i.e.}, mutual information~(MI)~\cite{Qu2002Information}, entropy~(EN)~\cite{Roberts2008Assessment} and standard deviation~(SD)~\cite{aslantas2015new} are introduced for evaluation. The quantitative results are reported in Figure~\ref{fig:numcpr}. As can be seen from the statistical results, our method continuously generates the largest or the second-largest mean value on three datasets among all evaluation metrics. Meanwhile, achieving a lower variance indicates that our method is more stable in dealing with various visual scenes. In specific, the largest average value on MI proves that our method transfers more considerable information from both two source images. Values of EN and SD reveal that our results contain abundant information and the highest contrast between targets and the background. In conclusion, our method stably reserves useful information to a certain degree, especially the most discriminative target, the richest texture details, and considerable structure similarity with the source images.

\subsection{Results of infrared-visible object detection}
To thoroughly discuss how does IVIF influences multi-modality object detection performance, two datasets,~\emph{i.e.,} Multispectral and M$^3$FD, are employed. In which, we utilized YOLOv5 as the baseline model for object detection. For fair comparison, we retain the detection model on the fused result of seven state-of-the-art methods, respectively.

\begin{figure*}[!htb]
	\centering
	\setlength{\tabcolsep}{1pt}
	\begin{tabular}{cccccccccc}
		
		\includegraphics[width=0.0945\textwidth,height=0.06\textheight]{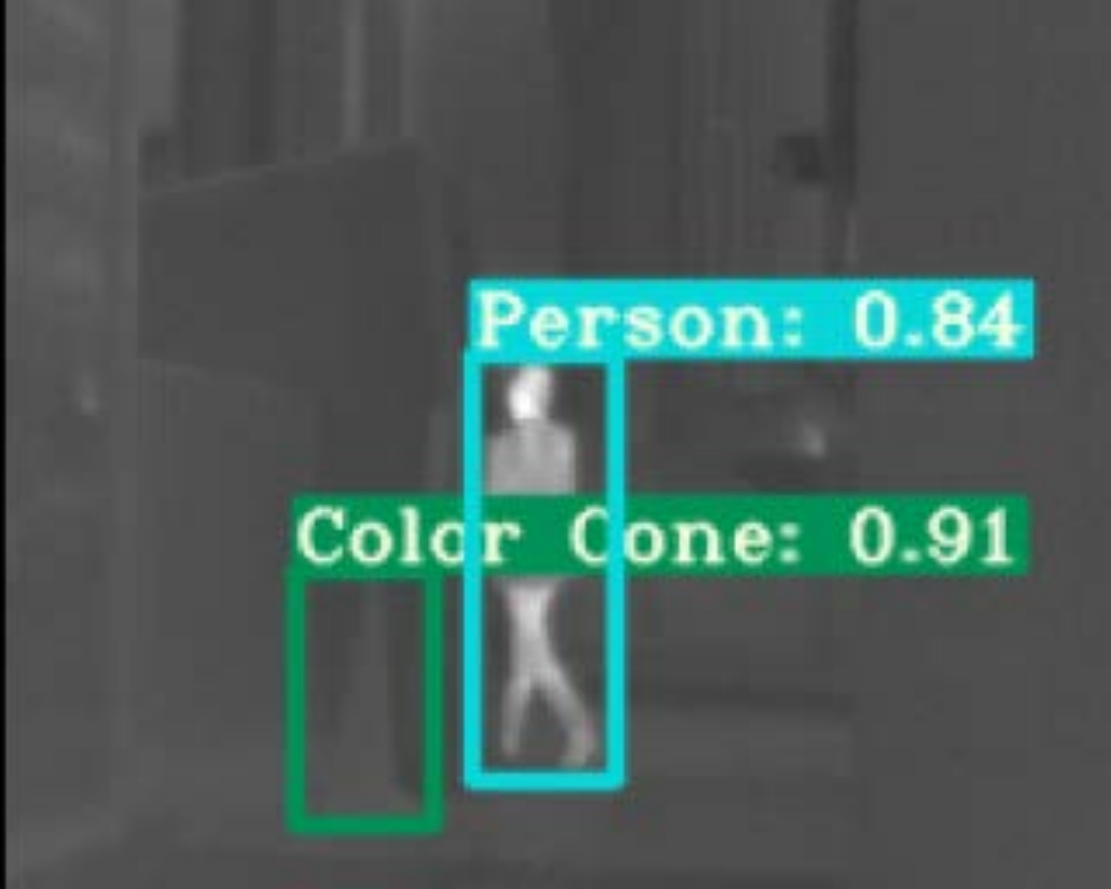}
		&\includegraphics[width=0.0945\textwidth,height=0.06\textheight]{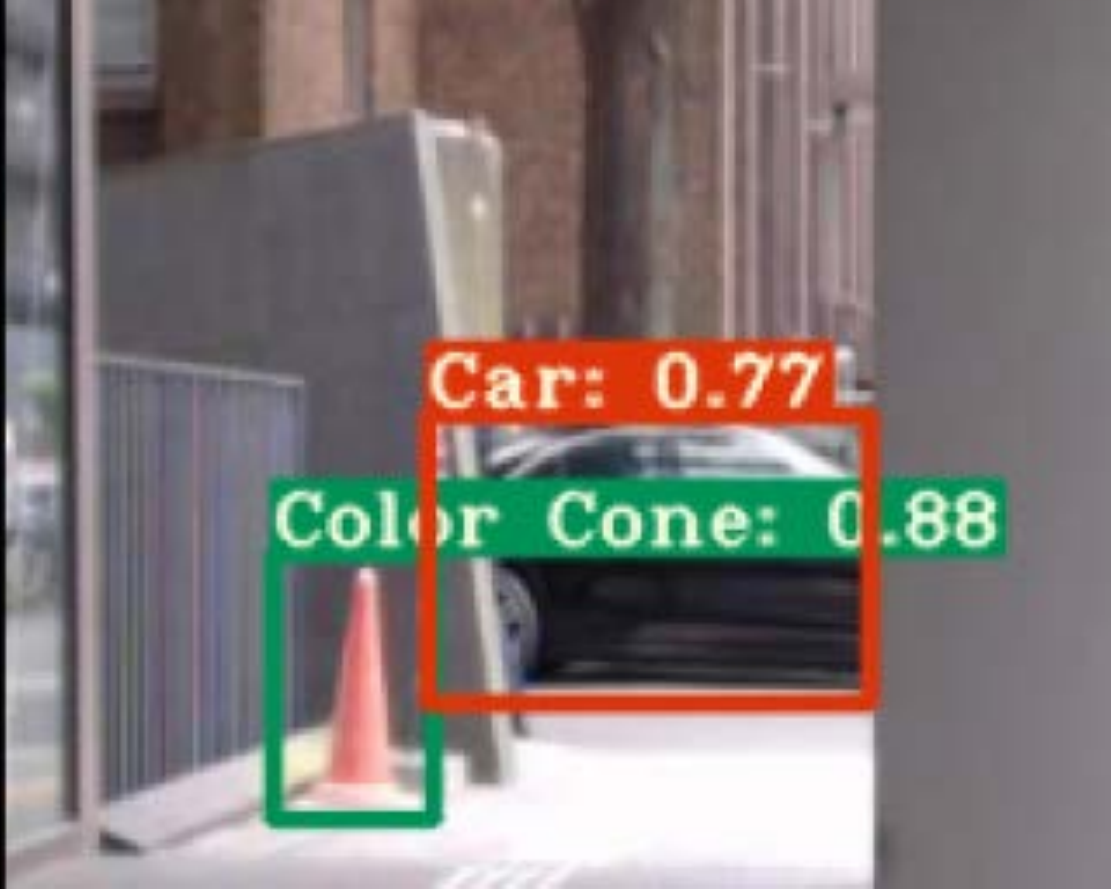}
		&\includegraphics[width=0.0945\textwidth,height=0.06\textheight]{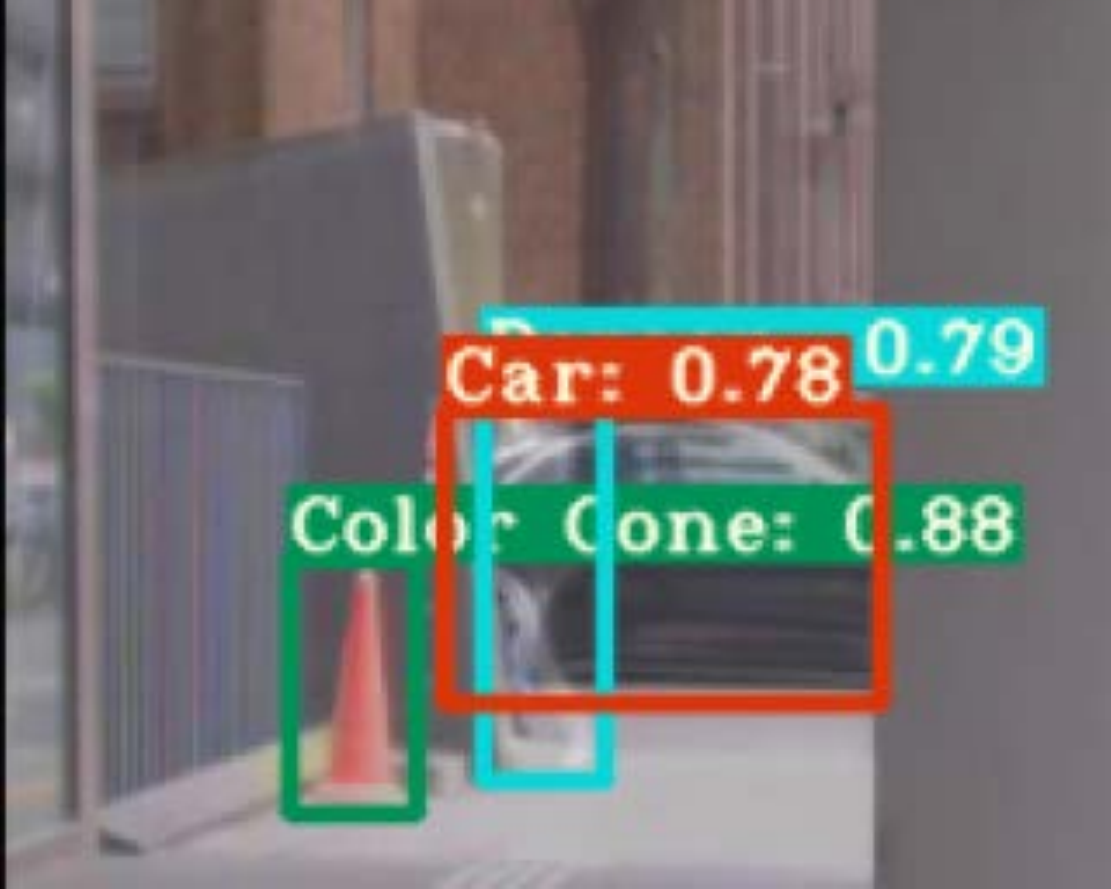}
		&\includegraphics[width=0.0945\textwidth,height=0.06\textheight]{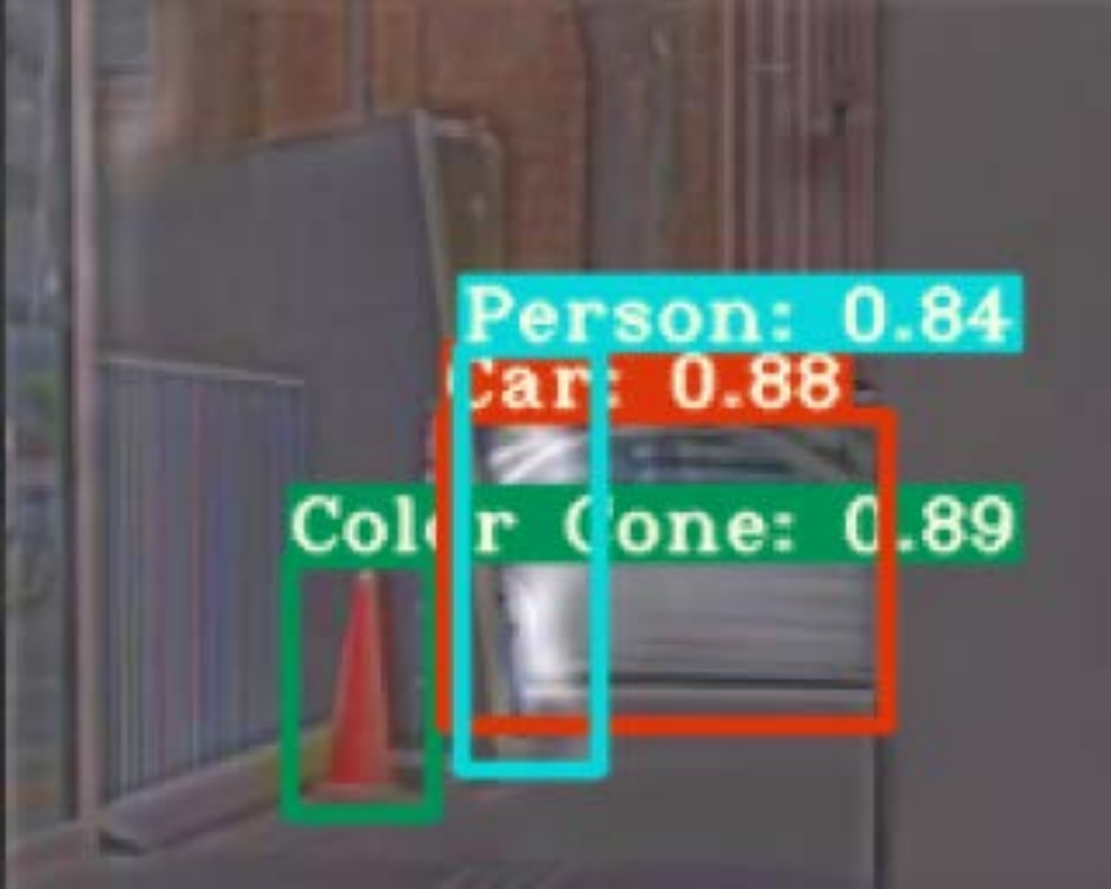}
		&\includegraphics[width=0.0945\textwidth,height=0.06\textheight]{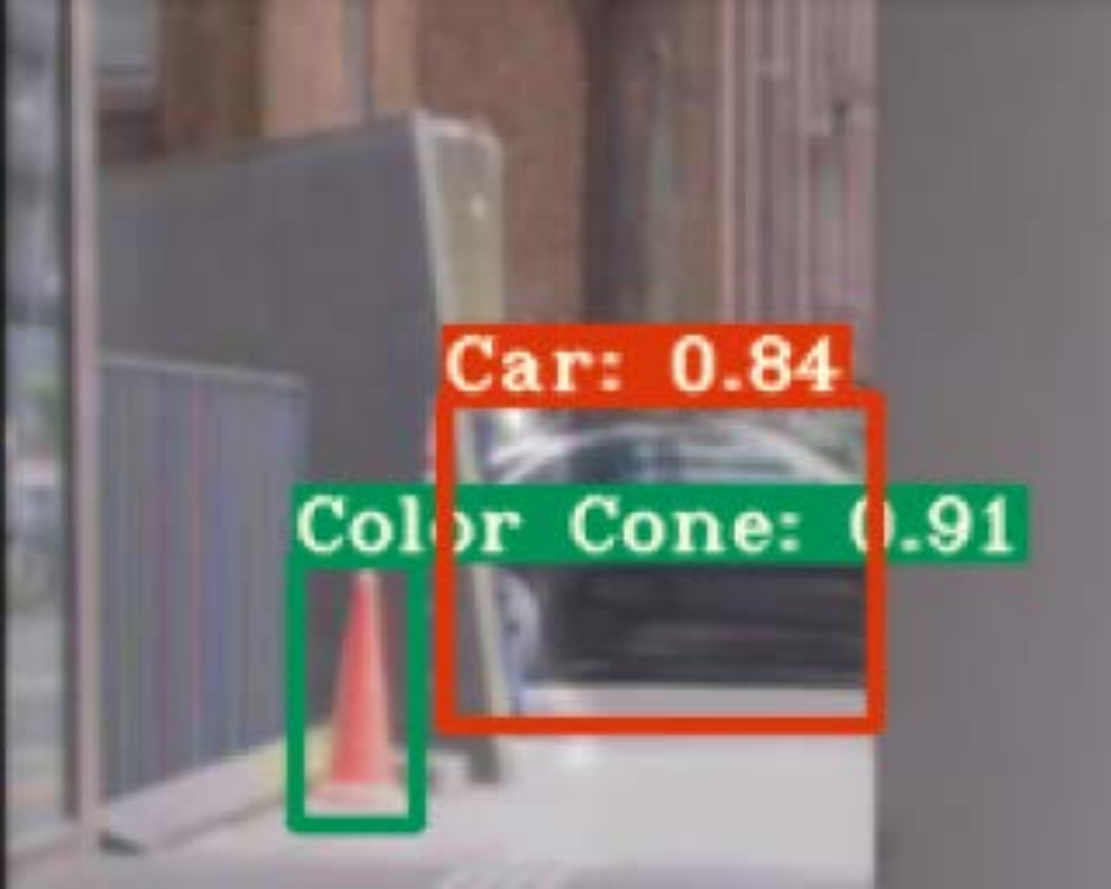}
		&\includegraphics[width=0.0945\textwidth,height=0.06\textheight]{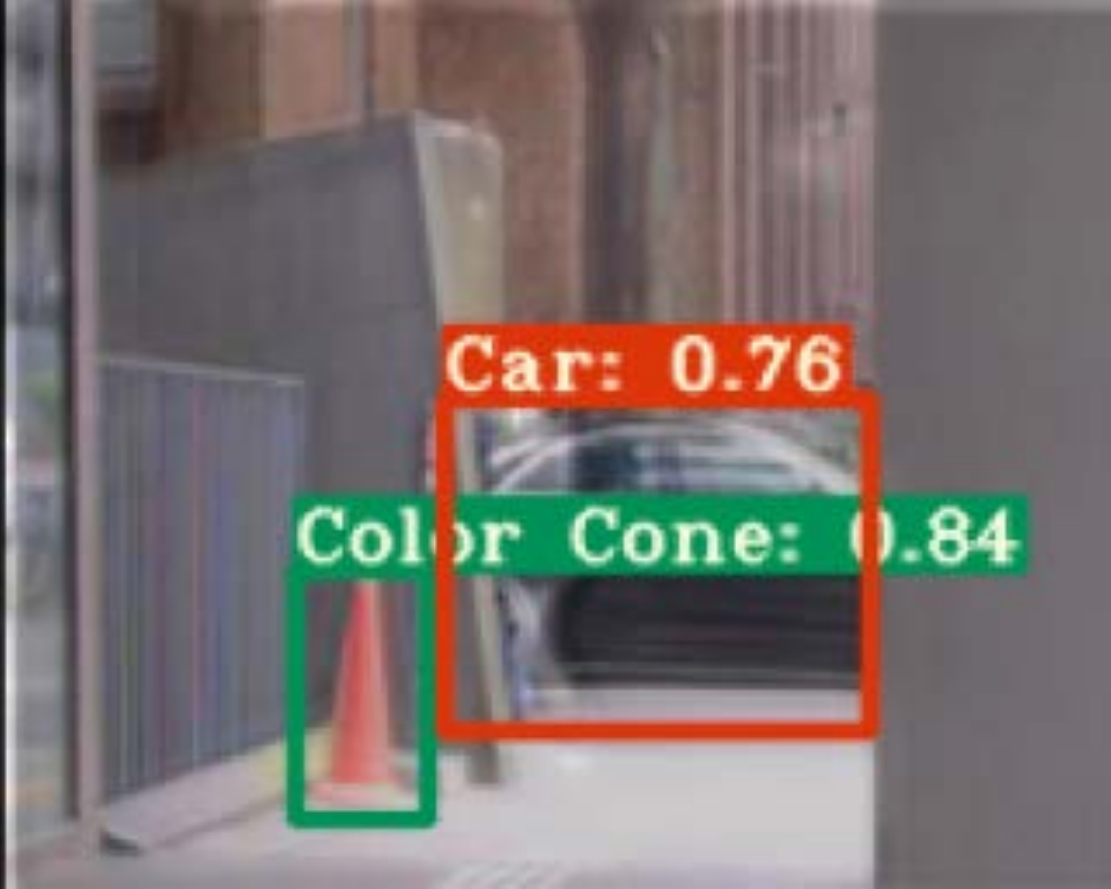}
		&\includegraphics[width=0.0945\textwidth,height=0.06\textheight]{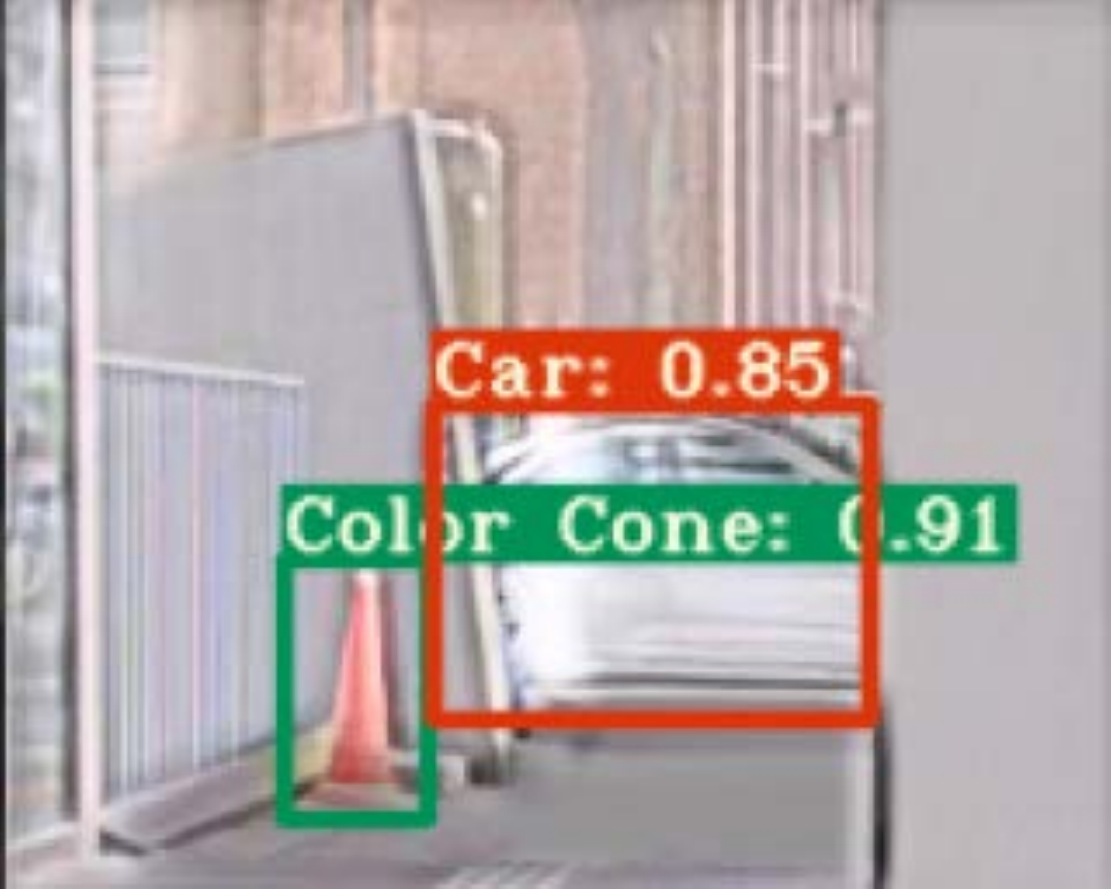}
		&\includegraphics[width=0.0945\textwidth,height=0.06\textheight]{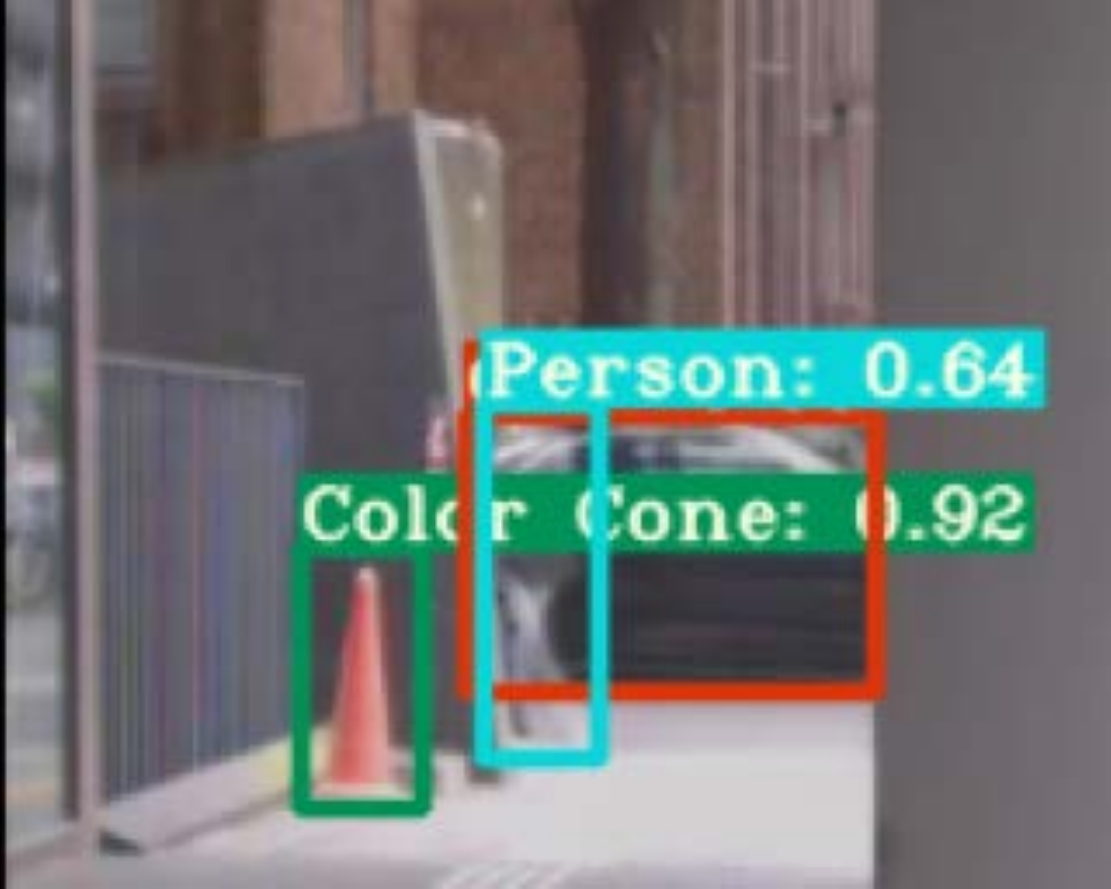}
		&\includegraphics[width=0.0945\textwidth,height=0.06\textheight]{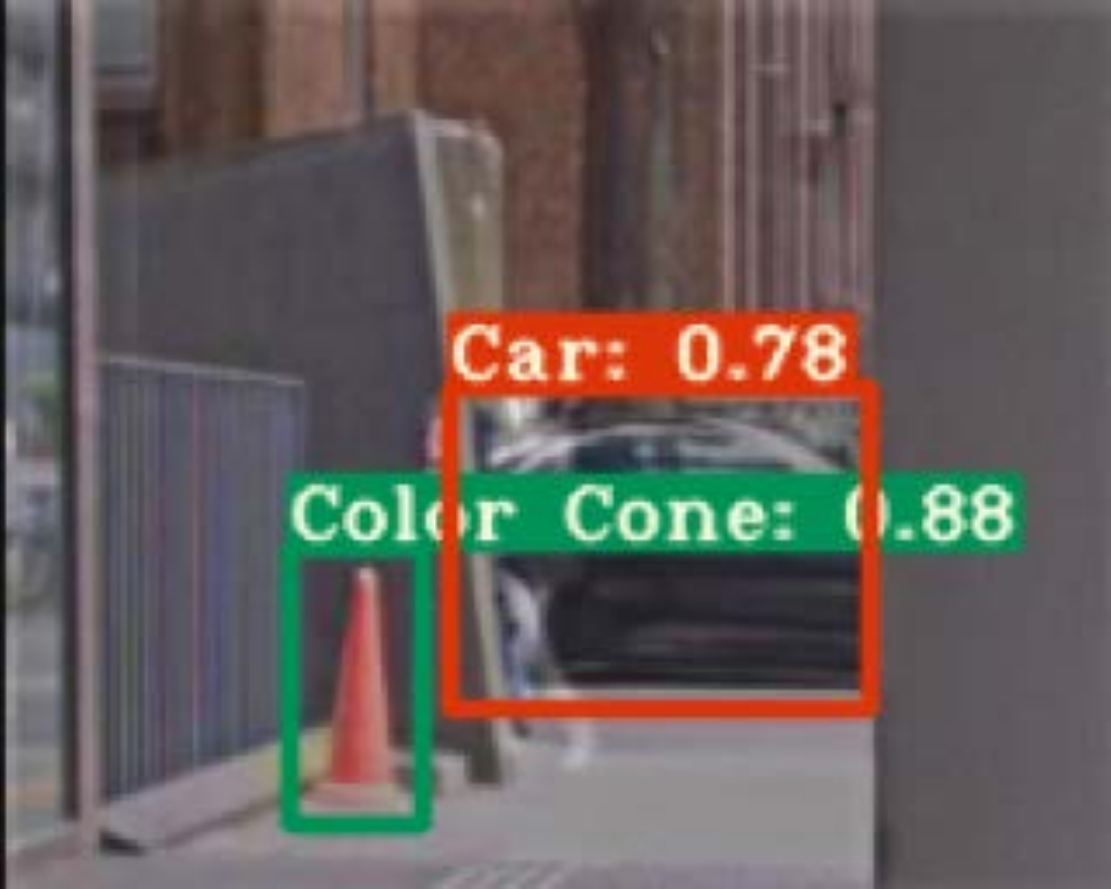}
		&\includegraphics[width=0.0945\textwidth,height=0.06\textheight]{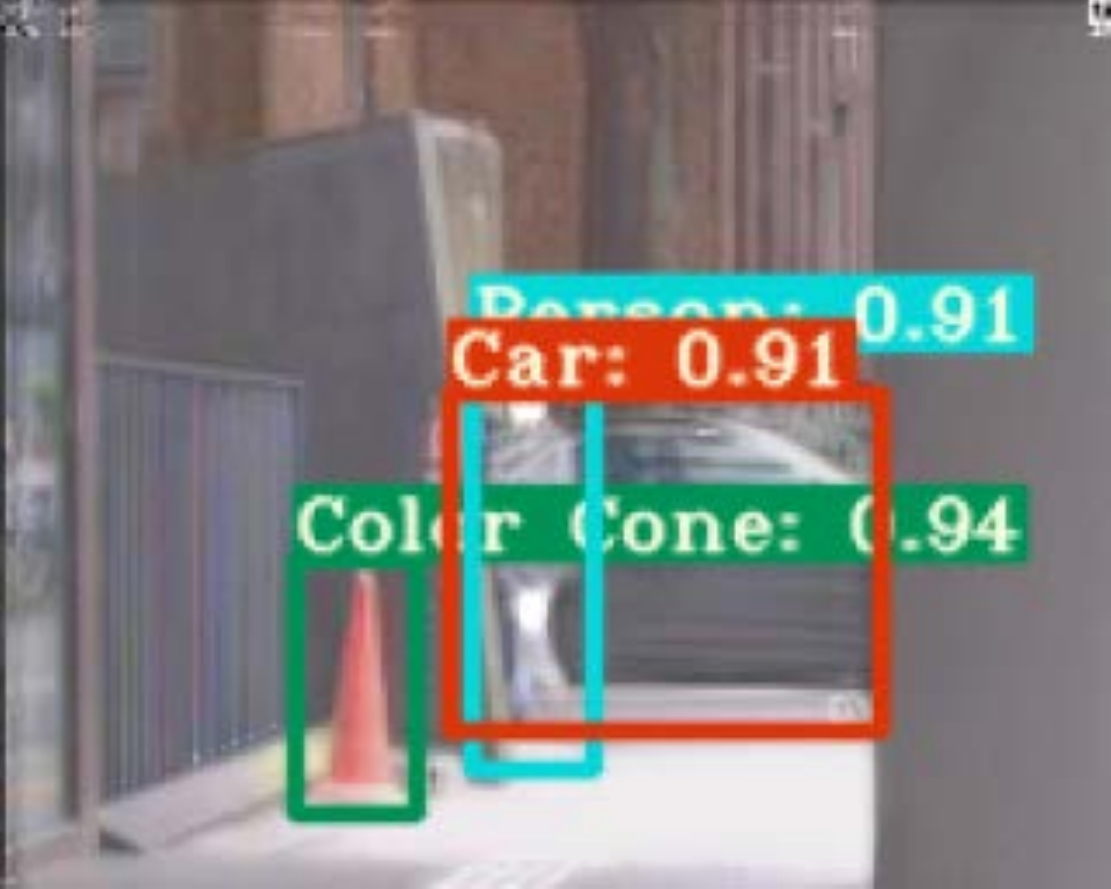}
		\\
		\includegraphics[width=0.0945\textwidth,height=0.06\textheight]{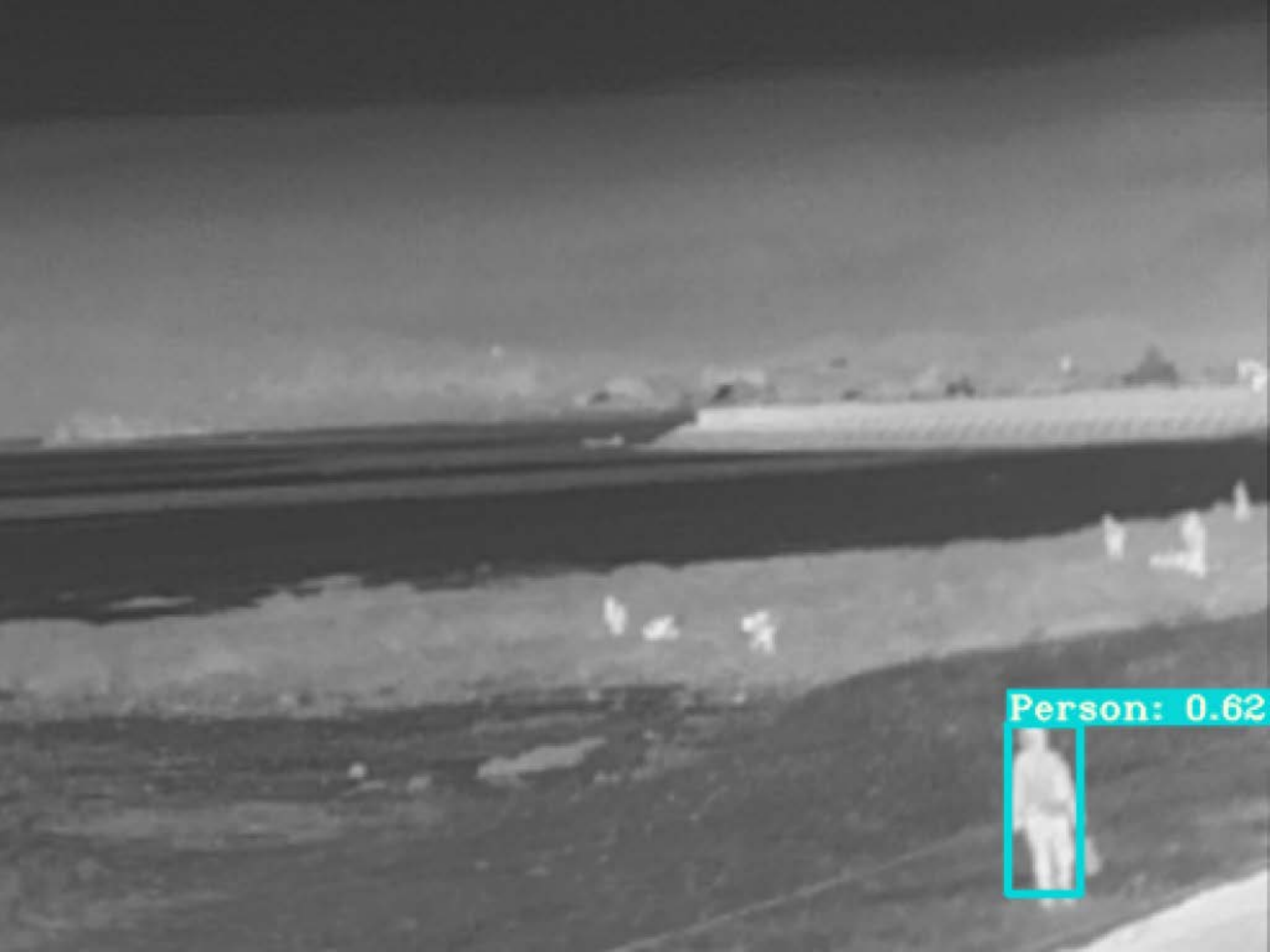}
		&\includegraphics[width=0.0945\textwidth,height=0.06\textheight]{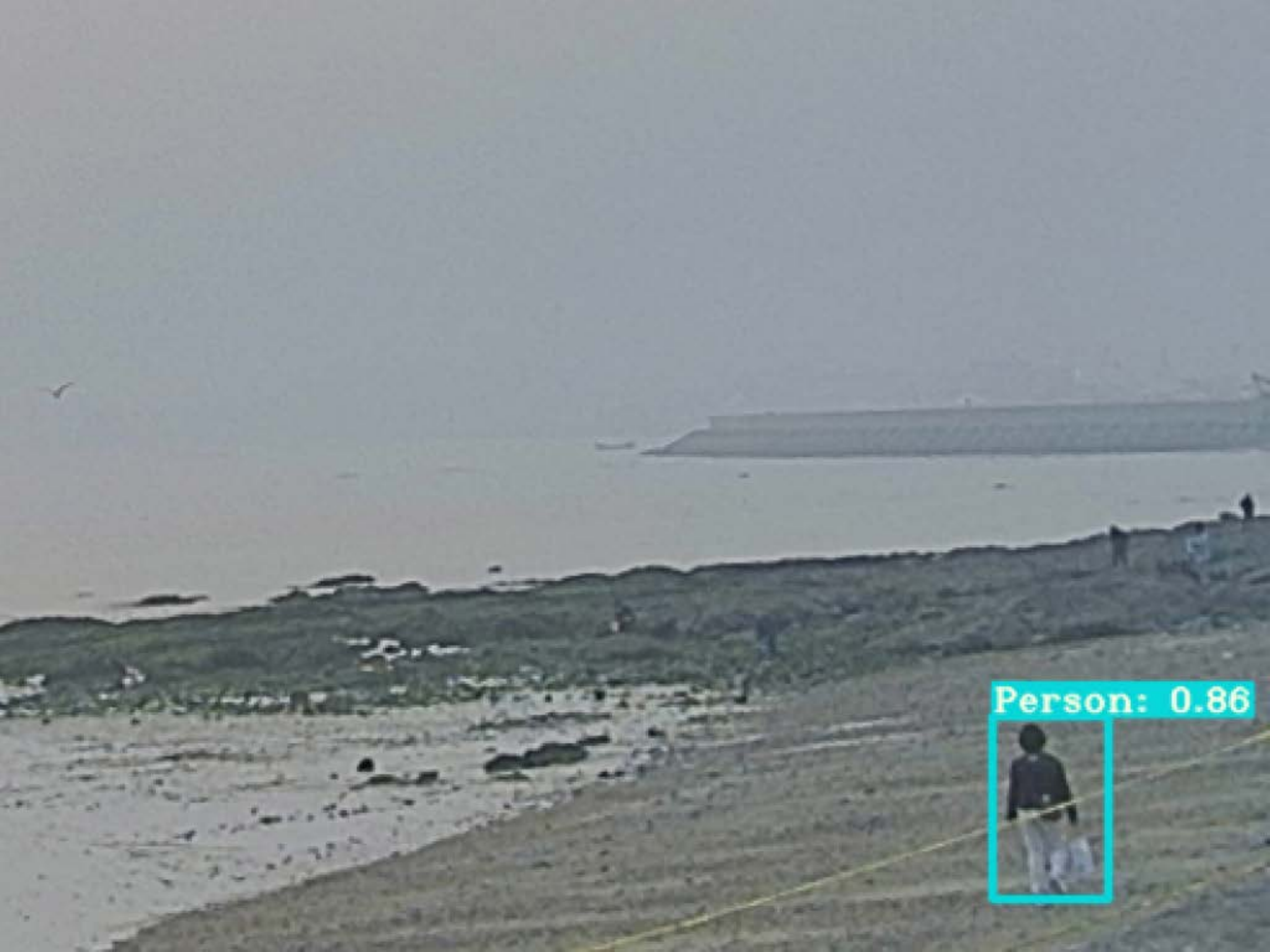}
		&\includegraphics[width=0.0945\textwidth,height=0.06\textheight]{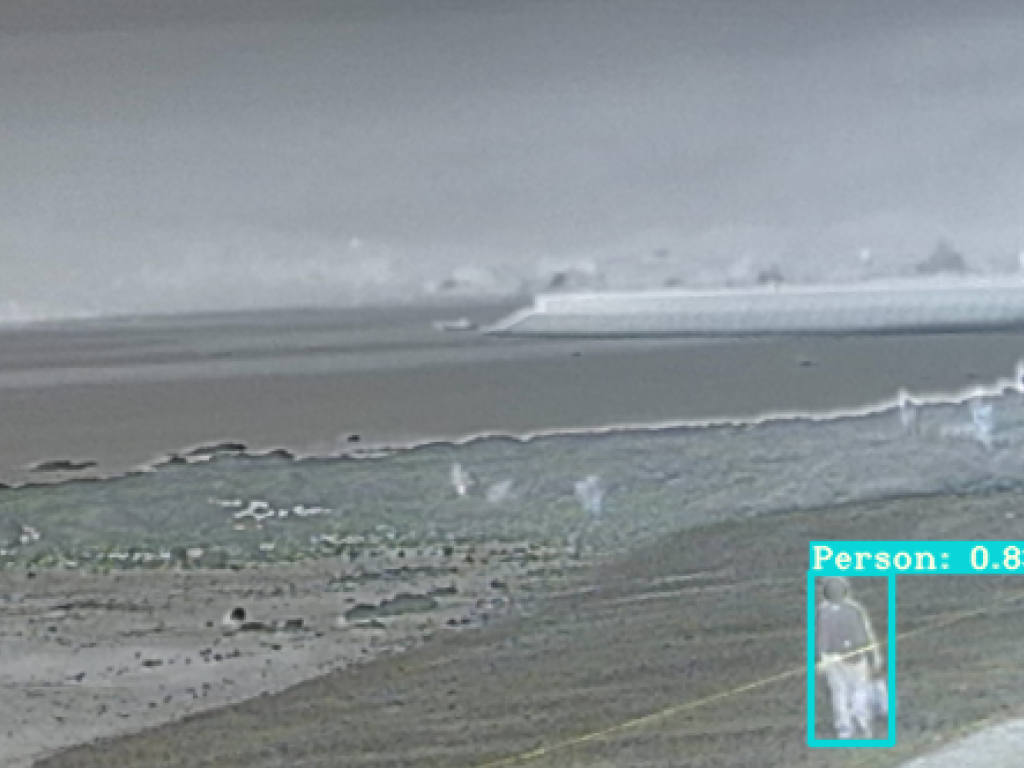}
		&\includegraphics[width=0.0945\textwidth,height=0.06\textheight]{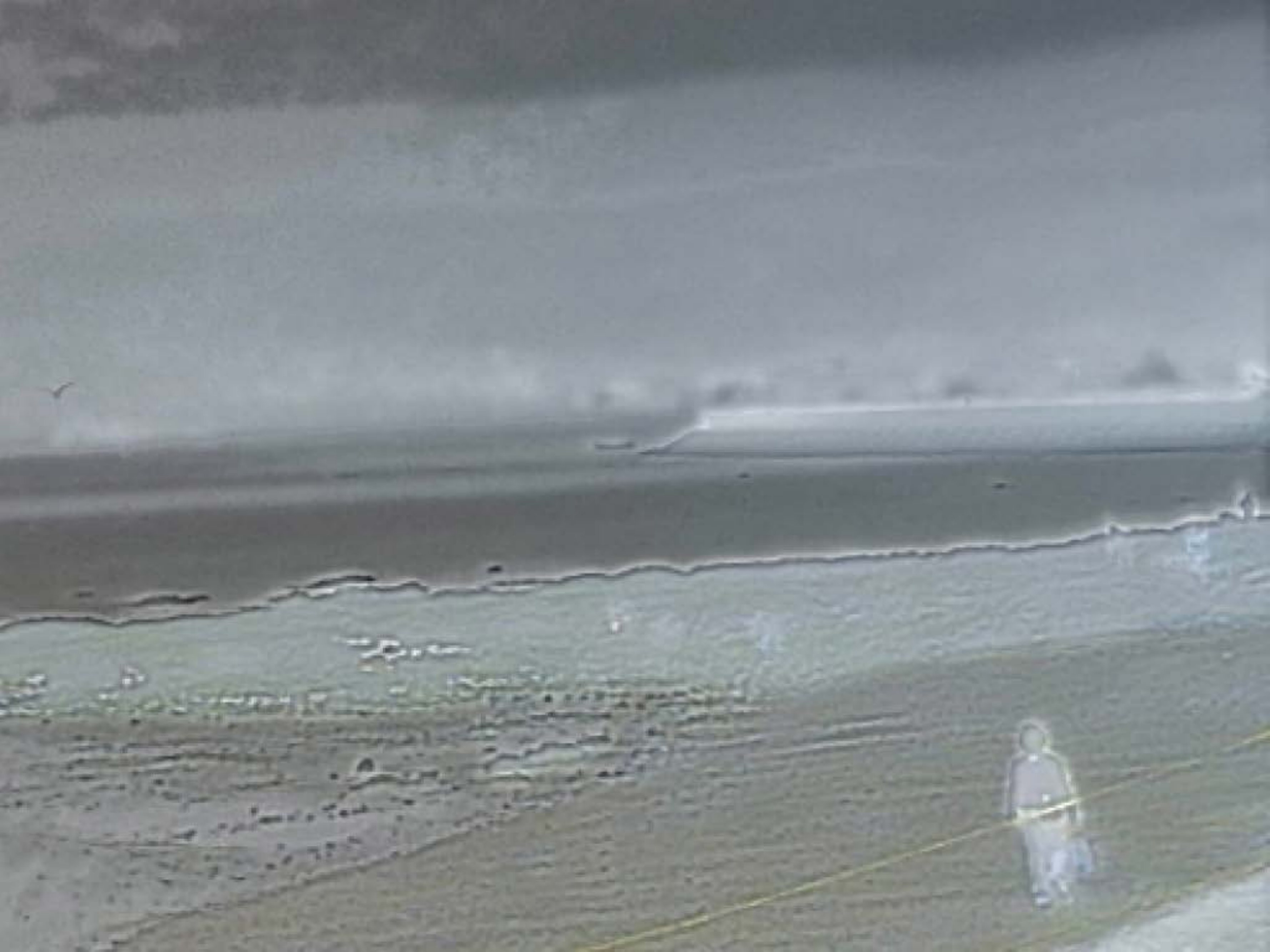}
		&\includegraphics[width=0.0945\textwidth,height=0.06\textheight]{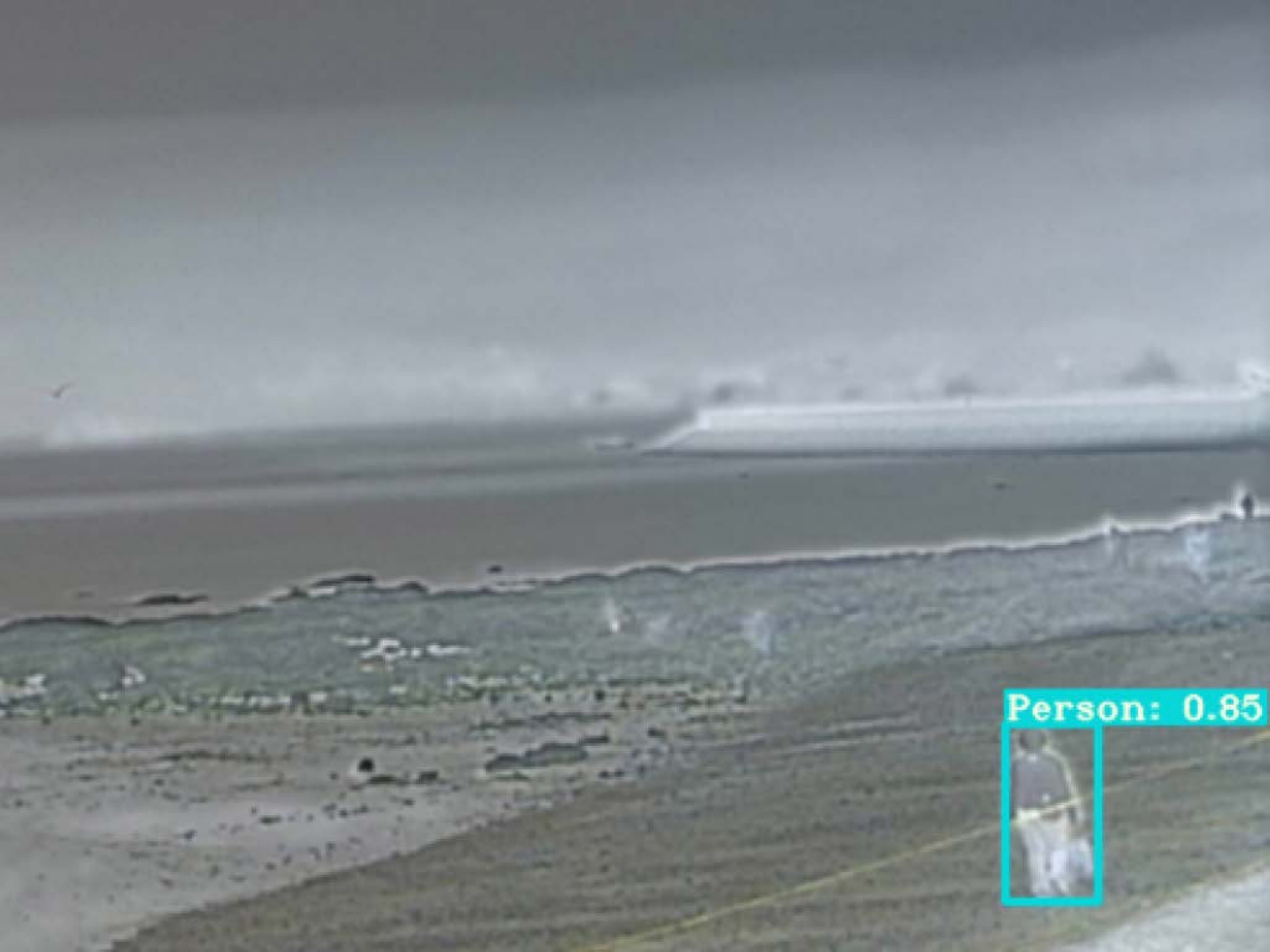}
		&\includegraphics[width=0.0945\textwidth,height=0.06\textheight]{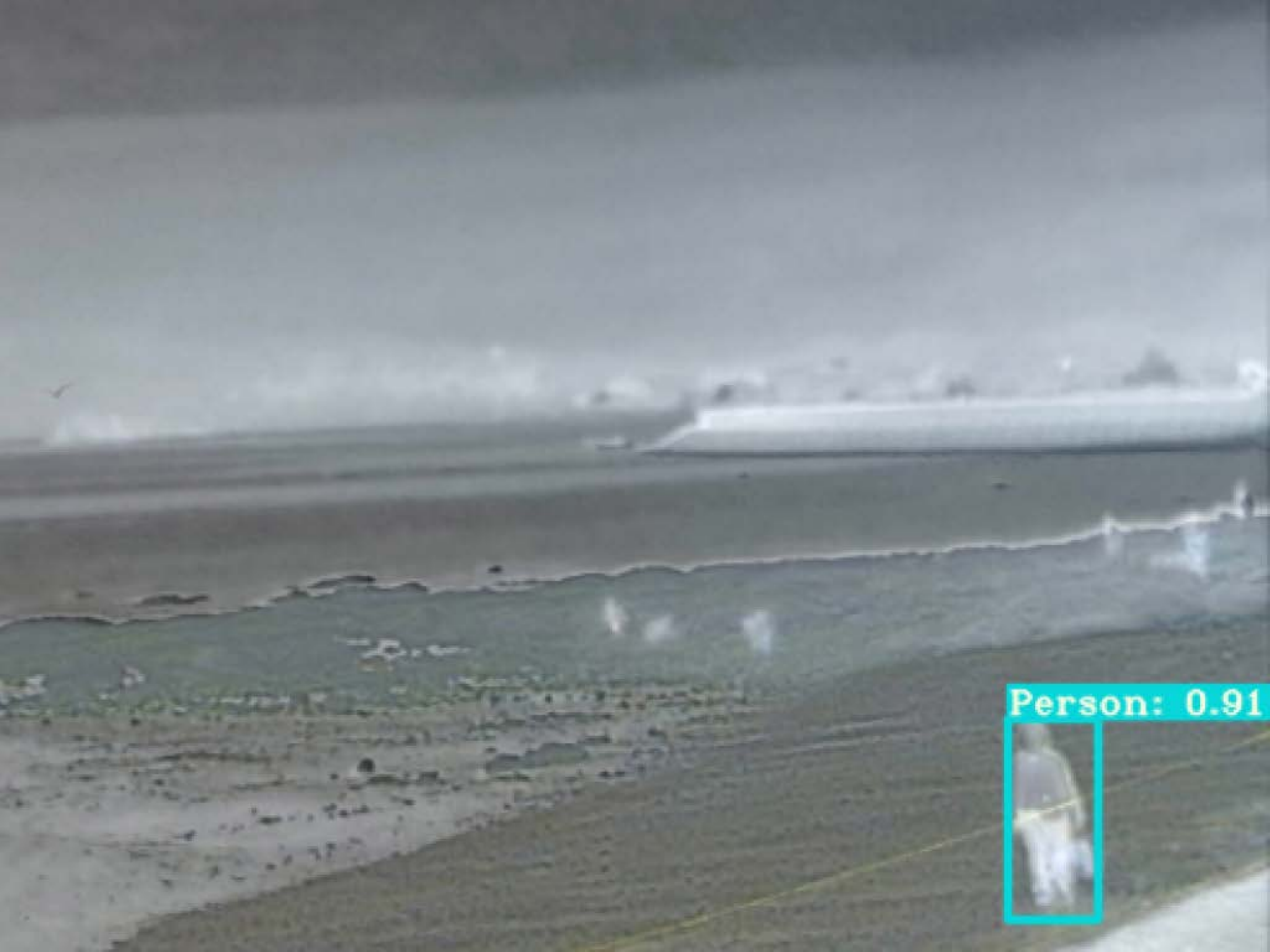}
		&\includegraphics[width=0.0945\textwidth,height=0.06\textheight]{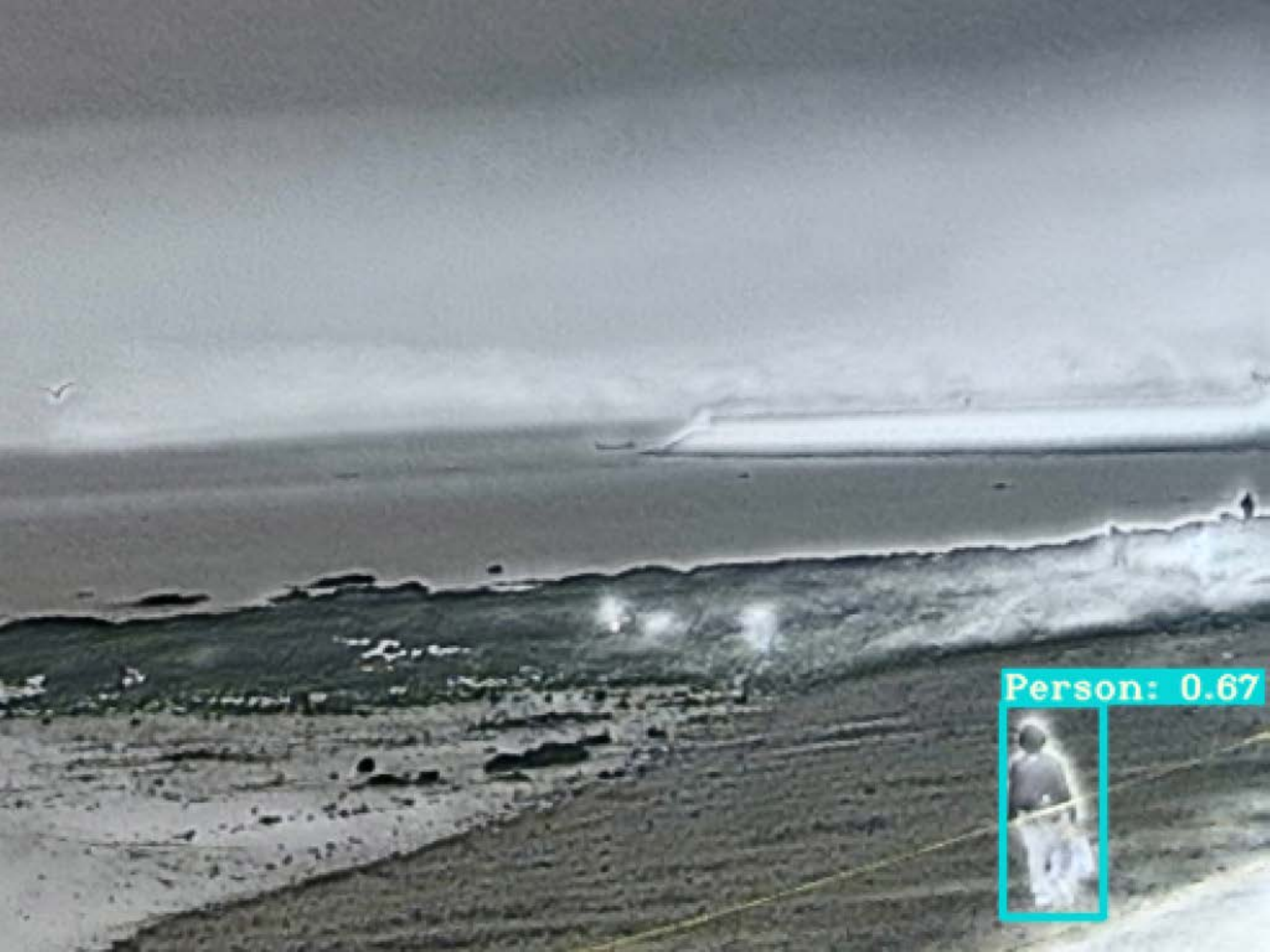}
		&\includegraphics[width=0.0945\textwidth,height=0.06\textheight]{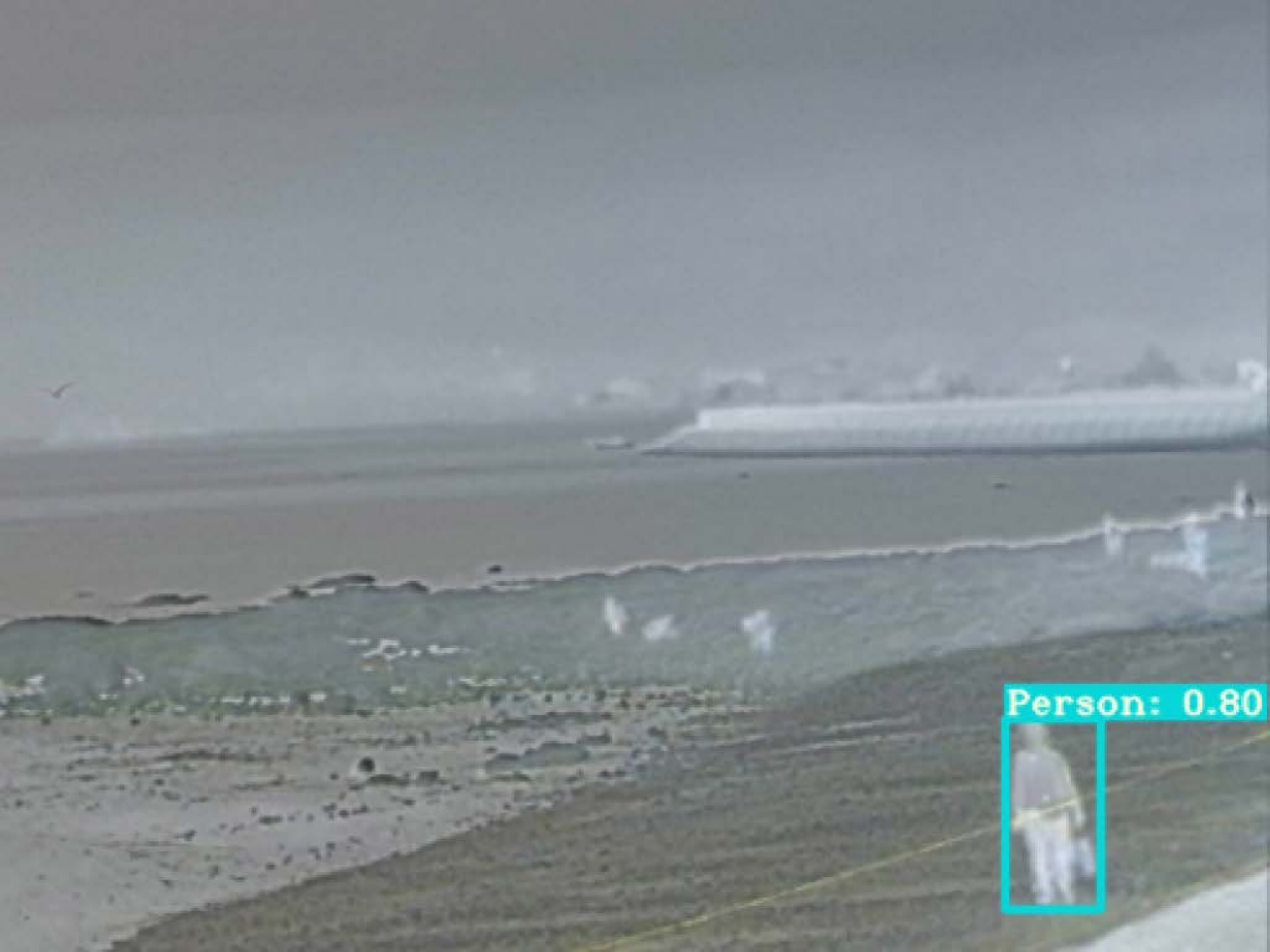}
		&\includegraphics[width=0.0945\textwidth,height=0.06\textheight]{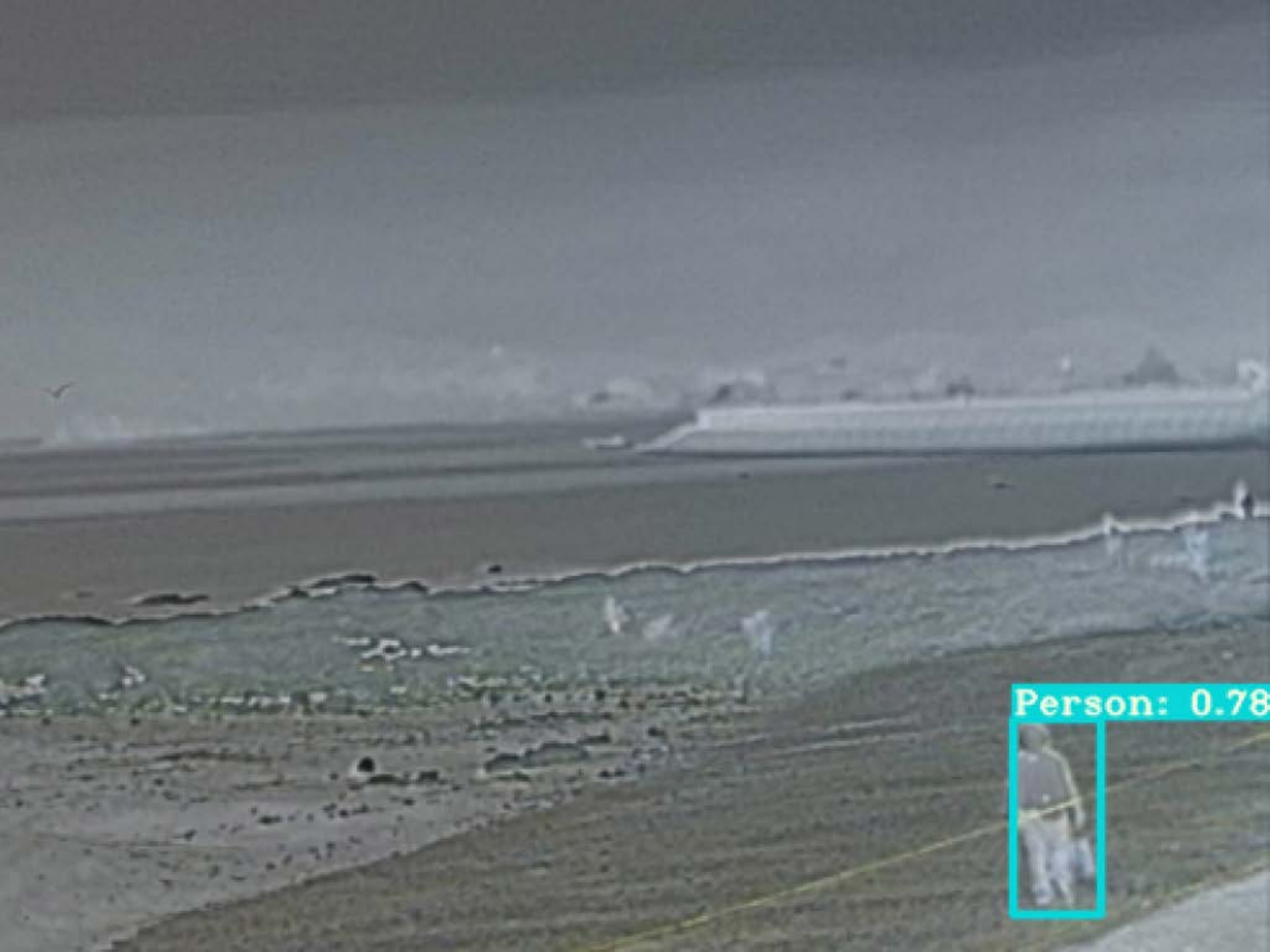}
		&\includegraphics[width=0.0945\textwidth,height=0.06\textheight]{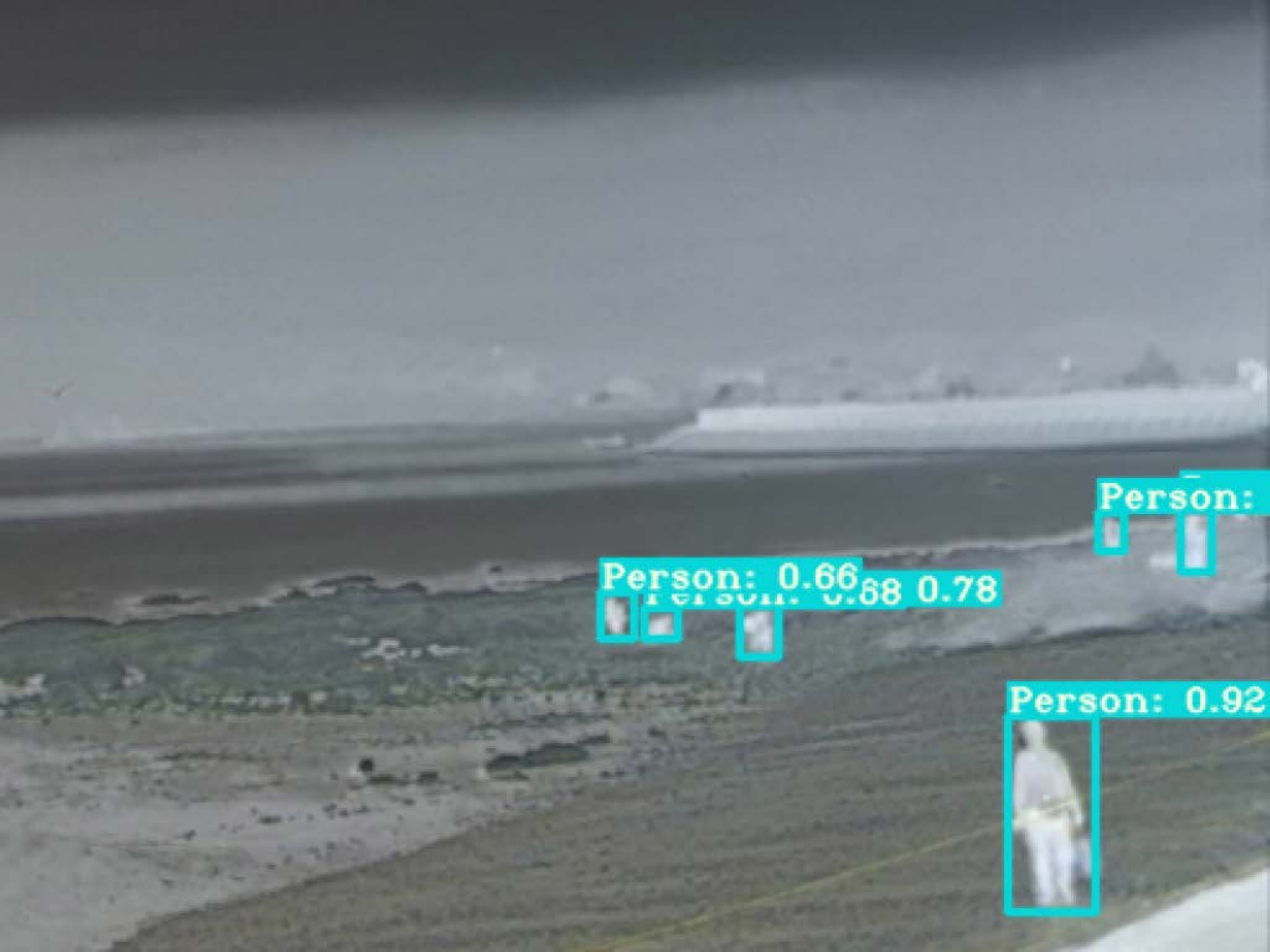}
		\\
		\footnotesize Ir image&\footnotesize Vis image&\footnotesize DenseFuse&\footnotesize FusionGAN&\footnotesize RFN&\footnotesize GANMcC&\footnotesize DDcGAN&\footnotesize MFEIF&\footnotesize U2Fusion&\footnotesize TarDAL	
		\\			
	\end{tabular}
	\vspace{-0.3cm}  
	\caption{Visual comparison of our TarDAL with state-of-the-art methods on the Multispectral and M$^3$FD datasets. }
	\label{fig:Visualdec}
\end{figure*}
\begin{table*}[!htb]
	\centering
	\renewcommand\arraystretch{1.1} 
	\setlength{\tabcolsep}{0.5mm}
	\begin{tabular}{|c|cccccc|c|ccccc|c|ccc|}
		\hline
		\multirow{2}{*}{\footnotesize Method}&\multicolumn{7}{c|}{\footnotesize Multispectral dataset}&\multicolumn{6}{c|}{\footnotesize M$^3$FD dataset}&\multicolumn{3}{c|}{\footnotesize Efficient Analysis}\\
		\cline{2-17} 
		~&\footnotesize Person&\footnotesize Car&\footnotesize Bike&\footnotesize Car Stop 
		&\footnotesize Cone &\footnotesize All&\footnotesize mAP@.5&\footnotesize Day&\footnotesize Overcast&\footnotesize Night&\footnotesize Challenge&\footnotesize All&\footnotesize mAP@.5&\footnotesize SIZE(M)&\footnotesize FLOPS(G)&\footnotesize TIME(s)\\
		\hline
		\footnotesize Infrared&\footnotesize 0.753 &\footnotesize 0.753 &\footnotesize 0.733 &\footnotesize 0.739 &\footnotesize 0.492 &\footnotesize 0.709 &\footnotesize 0.589 &\footnotesize 0.803 &\footnotesize 0.795 &\footnotesize 0.709 &\footnotesize 0.734 &\footnotesize 0.748 &\footnotesize 0.781&-&-&-\\
		\hline 
		\footnotesize Visible&\footnotesize 0.717 &\footnotesize 0.822 &\footnotesize 0.740 &\footnotesize 0.691 &\footnotesize 0.531 &\footnotesize 0.739 &\footnotesize 0.591 &\footnotesize \textcolor{red}{\textbf{0.824}} &\footnotesize 0.787 &\footnotesize 0.759 &\footnotesize 0.756 &\footnotesize 0.779 &\footnotesize 0.756 &-&-&-\\
		\hline 
		
		\footnotesize DenseFuse&\footnotesize 0.754&\footnotesize 0.833&\footnotesize \textcolor{blue}{\textbf{0.829}}&\footnotesize 0.749&\footnotesize 0.607 &\footnotesize 0.755&\footnotesize \textcolor{blue}{\textbf{0.608}}&\footnotesize 0.759&\footnotesize 0.806&\footnotesize \textcolor{blue}{\textbf{0.837}}&\footnotesize 0.776&\footnotesize 0.791&\footnotesize 0.783&\footnotesize \textcolor{red}{\textbf{0.074}}&\footnotesize 48.92&\footnotesize 0.251  \\
		\hline 
		
		\footnotesize FusionGAN&\footnotesize \textcolor{red}{\textbf{0.763}}&\footnotesize \textcolor{blue}{\textbf{0.846}}&\footnotesize 0.828&\footnotesize  0.751&\footnotesize  0.575&\footnotesize 0.756&\footnotesize 0.601&\footnotesize 0.816&\footnotesize 0.798&\footnotesize 0.667&\footnotesize 0.773&\footnotesize 0.765&\footnotesize 0.788 &\footnotesize 0.925&\footnotesize 497.76 &\footnotesize 0.124 \\
		\hline 
		
		\footnotesize RFN&\footnotesize 0.505 &\footnotesize 0.619 &\footnotesize 0.520 &\footnotesize 0.512 &\footnotesize 0.427 &\footnotesize 0.605 &\footnotesize 0.592 &\footnotesize 0.796 &\footnotesize 0.803 &\footnotesize 0.827 &\footnotesize \textcolor{blue}{\textbf{0.793}} &\footnotesize 0.794 &\footnotesize 0.796 &\footnotesize 10.93&- &\footnotesize 0.238 \\
		\hline 
		
		\footnotesize GANMcC&\footnotesize 0.472&\footnotesize 0.811&\footnotesize 0.765&\footnotesize  0.680&\footnotesize  0.620&\footnotesize 0.724&\footnotesize 0.603&\footnotesize 0.796&\footnotesize 0.811&\footnotesize 0.827&\footnotesize 0.790&\footnotesize \textcolor{blue}{\textbf{0.805}}&\footnotesize \textcolor{blue}{\textbf{0.797}} &\footnotesize 1.864&\footnotesize 1002.56 &\footnotesize 0.246\\
		\hline 
		
		\footnotesize DDcGAN&\footnotesize 0.735&\footnotesize 0.841&\footnotesize 0.810&\footnotesize  \textcolor{red}{\textbf{0.761}}&\footnotesize  \textcolor{blue}{\textbf{0.645}}&\footnotesize \textcolor{blue}{\textbf{0.766}}&\footnotesize 0.594&\footnotesize 0.780&\footnotesize 0.771&\footnotesize 0.689&\footnotesize 0.776&\footnotesize 0.748&\footnotesize 0.744 &\footnotesize 1.097 &\footnotesize 896.84 &\footnotesize 0.211\\
		
		\hline  
		\footnotesize MFEIF&\footnotesize 0.760&\footnotesize 0.837&\footnotesize 0.790&\footnotesize  0.741&\footnotesize  0.640&\footnotesize 0.755&\footnotesize 0.607&\footnotesize 0.770&\footnotesize \textcolor{blue}{\textbf{0.812}}&\footnotesize 0.683&\footnotesize 0.778&\footnotesize 0.744&\footnotesize 0.718 &\footnotesize \textcolor{blue}{\textbf{0.158}} &\footnotesize \textcolor{blue}{\textbf{25.32}} &\footnotesize \textcolor{blue}{\textbf{0.045}} \\
		\hline 
		\footnotesize U2Fusion&\footnotesize 0.574&\footnotesize 0.599&\footnotesize 0.579&\footnotesize 0.530&\footnotesize  0.432&\footnotesize 0.562&\footnotesize 0.604&\footnotesize 0.793&\footnotesize 0.783&\footnotesize 0.836&\footnotesize 0.773&\footnotesize 0.801&\footnotesize 0.782 &\footnotesize 0.659 &\footnotesize 366.34 &\footnotesize 0.123\\
		\hline 
		
		\footnotesize TarDAL&\footnotesize\textcolor{blue}{\textbf{0.762}}&\footnotesize \textcolor{red}{\textbf{0.868}}&\footnotesize \textcolor{red}{\textbf{0.833}}&\footnotesize  \textcolor{blue}{\textbf{0.757}} &\footnotesize \textcolor{red}{\textbf{0.678}}&\footnotesize \textcolor{red}{\textbf{0.780}}&\footnotesize \textcolor{red}{\textbf{0.613}}&\footnotesize \textcolor{blue}{\textbf{0.823}}&\footnotesize \textcolor{red}{\textbf{0.816}}&\footnotesize \textcolor{red}{\textbf{0.846}}&\footnotesize \textcolor{red}{\textbf{0.869}} &\footnotesize \textcolor{red}{\textbf{0.846}}&\footnotesize \textcolor{red}{\textbf{0.807}} &\footnotesize 0.296 &\footnotesize \textcolor{red}{\textbf{14.88}} &\footnotesize \textcolor{red}{\textbf{0.041}}\\
		\hline 
	\end{tabular}
	\caption{Quantitative results of object detection on the Multispectral and M$^3$FD datasets among all the image fusion method + detector (YOLOv5). The best result is in {\textcolor{red}{\textbf{red}}} whereas the second best one is in {\textcolor{blue}{\textbf{blue}}}.}
	\label{tab: detectionnum}
\end{table*}
\noindent\textbf{Qualitative Comparisons} 
As shown in Figure~\ref{fig:Visualdec}, note that merely using an infrared or visible sensor cannot detect well,~\emph{e.g.,} a stopped car for infrared image and person for the visible one. On the contrary, almost all the fusion methods improve the detection performance by utilizing complementary information from both sides. With designing target-aware bilevel adversarial learning and a cooperative training scheme integration in our method, we can continuously generate a detection-friendly fused result, which has advantage in detecting person and vehicle,~\emph{e.g.,} the sheltered car and pedestrians on distant rocks.

\noindent\textbf{Quantitative Comparisons} 
Table~\ref{tab: detectionnum} reported the quantitative results on two datasets. Almost all the fusion methods achieve promising detection results, in which the detection AP greatly exceed the case of using only the visible or infrared images. Note that our TarDAL is superior to other methods in terms of detection mAP on two datasets, which obtain 1.4$\%$ and 1.1$\%$ improvement compared to the second one, \emph{i.e.,}~DenseFuse and GANMcC. It is worth pointing out that our TarDAL has advantage in dealing with the challenge scenes because TarDAL fully discovers the unique information from different modalities.

\noindent\textbf{Computational Complexity Analysis} 
To comprehensively analyze the computational complexity of our method, we provide the time consumption and the computational efficiency of all the methods. As shown in the last column of Table~\ref{tab: detectionnum}, the strong computing ability of CNNs allows these learning-based methods to achieve high speed.~Note that our method simultaneously achieves the highest running speed and lower computing complexity in terms of FLOPs and training parameters, ensembling the follow-up high-level vision application with high efficiency.

\subsection{Ablation studies}
\begin{figure}[!htb]
	\centering
	\setlength{\tabcolsep}{1pt} 
	
	\includegraphics[width=0.48\textwidth,height=0.2\textheight]{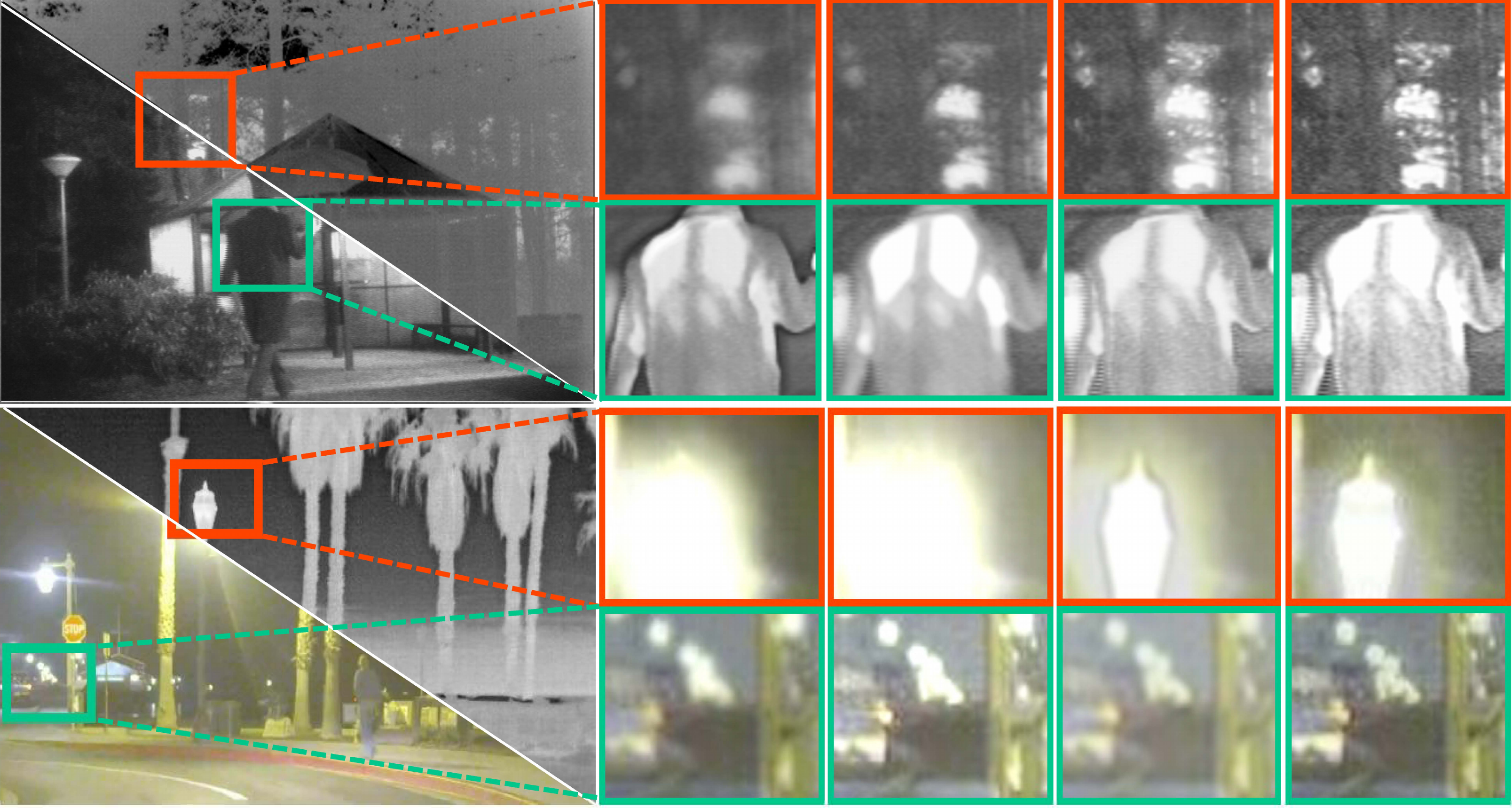}
	
	\caption{Progressive fusion results. From left to right: source images, base network, w/o-~$D_T$, w/o-~$D_D$ and the full model.}
	\label{fig:abstructure}
	\vspace{-0.5cm}  
\end{figure}

\noindent\textbf{Study on model architectures}
We investigate the model architecture of our method and further validate the effectiveness of different individual components. First, we remove the target discriminator $D_T$ from our whole network. In Figure~\ref{fig:abstructure}, due to the lack of distinguishing significant infrared targets in this variant, the fused results tend to blur the target to a certain degree. Besides, in Table~\ref{tab: strurecture}, note that $D_T$ also plays a vital role in boosting the detection performance after fusion.  Second, the detail discriminator $D_D$ has a contribution in preserving textural details from the visible images. In the absence of $D_D$, the background details of the fused image cannot be fully recovered, and the intuitive visual results can be found in~Figure~\ref{fig:abstructure}.  However, $D_D$ has a tiny negative impact on object detection due to redundant background details.  Furthermore, Without $D_T$ and $D_D$ integrating into our whole network, EN and SD can achieve the highest value on the TNO dataset. This is because that the heavy noise on the fusion results may cause a significant rise in terms of EN and SD. In summary, our method depends on the intermediate results of each step, and each step plays a positive effect on the final fused result. 

\begin{table*}[!htb]
	\centering
	\renewcommand\arraystretch{1.1} 
	\setlength{\tabcolsep}{2.75mm}
	\begin{tabular}{|c|cc|ccc|ccc|cccc|}
		\hline
		\multirow{2}{*}{\footnotesize Model}&\multicolumn{2}{c|}{\footnotesize Discriminator}&\multicolumn{3}{c|}{\footnotesize TNO Dateset}&\multicolumn{3}{c|}{\footnotesize Roadscene Dataset}&\multicolumn{4}{c|}{\footnotesize M$^3$FD Dataset}\\
		\cline{2-13} 
		~&{\footnotesize $\mathcal{D}_T$}&{\footnotesize $\mathcal{D}_D$}&\footnotesize MI&\footnotesize EN&\footnotesize SD&\footnotesize MI 
		&\footnotesize EN &\footnotesize SD&\footnotesize MI&\footnotesize EN&\footnotesize SD&\footnotesize mAP@.5\\
		\hline
		\footnotesize M1& \ding{55}  & \ding{55}&\footnotesize 2.506 &\footnotesize \textcolor{red}{\textbf{7.223}} &\footnotesize \textcolor{red}{\textbf{53.107}} &\footnotesize \textcolor{blue}{\textbf{3.307}} &\footnotesize \textcolor{blue}{\textbf{7.295}}&\footnotesize \textcolor{blue}{\textbf{48.561}} &\footnotesize \textcolor{blue}{\textbf{2.942}} &\footnotesize \textcolor{blue}{\textbf{7.282}} &\footnotesize \textcolor{blue}{\textbf{44.857}} &\footnotesize 0.722 \\
		\hline 
		\footnotesize M2& \ding{55}  & \ding{52}&\footnotesize 2.591 &\footnotesize 7.045 &\footnotesize 50.245 &\footnotesize 3.274 &\footnotesize 7.128 &\footnotesize 46.751 &\footnotesize 2.842 &\footnotesize 6.981 &\footnotesize 39.364 &\footnotesize 0.719 \\
		\hline 
		\footnotesize M3& \ding{52}  & \ding{55}&\footnotesize 2.596&\footnotesize 7.024&\footnotesize 46.727&\footnotesize 3.127&\footnotesize 7.037 &\footnotesize 42.656&\footnotesize 2.814&\footnotesize 7.086&\footnotesize 41.255&\footnotesize\textcolor{blue}{\textbf{0.781}}\\
		\hline 
		
		\footnotesize M4& \ding{52}  & \ding{52}&\footnotesize \textcolor{red}{\textbf{2.766}}&\footnotesize \textcolor{blue}{\textbf{7.177}}&\footnotesize \textcolor{blue}{\textbf{51.352}}&\footnotesize \textcolor{red}{\textbf{3.378}}&\footnotesize \textcolor{red}{\textbf{7.355}} &\footnotesize \textcolor{red}{\textbf{49.637}}&\footnotesize \textcolor{red}{\textbf{3.211}}&\footnotesize \textcolor{red}{\textbf{7.313}}&\footnotesize \textcolor{red}{\textbf{45.827}}&\footnotesize \textcolor{red}{\textbf{0.807}}\\
		\hline 		
	\end{tabular}
	\vspace{-0.2cm}
	\caption{Quantitative comparisons of different model architectures. The best result is in {\textcolor{red}{\textbf{red}}} whereas the second best one is in {\textcolor{blue}{\textbf{blue}}}.}
	\label{tab: strurecture}
\end{table*}
\begin{table*}[!htb]
	\centering
	\renewcommand\arraystretch{1.1} 
	\setlength{\tabcolsep}{1.6mm}
	\begin{tabular}{|c|cccccc|c|ccccc|c|}
		\hline
		\multirow{2}{*}{\footnotesize Training Strategy}&\multicolumn{7}{c|}{\footnotesize Multispectral dataset}&\multicolumn{6}{c|}{\footnotesize M$^3$FD dataset}\\
		\cline{2-14} 
		~&\footnotesize Person&\footnotesize Car&\footnotesize Bike&\footnotesize Car Stop 
		&\footnotesize Cone &\footnotesize All&\footnotesize mAP@.5&\footnotesize Day&\footnotesize Overcast&\footnotesize Night&\footnotesize Challenge&\footnotesize All&\footnotesize mAP@.5\\
		
		\hline
		\footnotesize TarDAL$_{DT}$&\footnotesize 0.762&\footnotesize\textcolor{red}{\textbf{0.868}}&\footnotesize 0.833&\footnotesize  \textcolor{blue}{\textbf{0.757}} &\footnotesize \textcolor{blue}{\textbf{0.678}}&\footnotesize \textcolor{blue}{\textbf{0.780}}&\footnotesize 0.613&\footnotesize \textcolor{blue}{\textbf{0.823}}&\footnotesize 0.816&\footnotesize 0.846&\footnotesize 0.869 &\footnotesize 0.846&\footnotesize 0.807\\
		\hline 
		
		\footnotesize TarDAL$_{TT}$&\footnotesize \textcolor{blue}{\textbf{0.827}}&\footnotesize 0.862&\footnotesize 0.881&\footnotesize 0.667&\footnotesize  0.539&\footnotesize 0.755&\footnotesize \textcolor{blue}{\textbf{0.615}}&\footnotesize \textcolor{red}{\textbf{0.827}}&\footnotesize \textcolor{blue}{\textbf{0.828}}&\footnotesize \textcolor{blue}{\textbf{0.862}}&\footnotesize \textcolor{blue}{\textbf{0.881}}&\footnotesize \textcolor{blue}{\textbf{0.850}}&\footnotesize \textcolor{blue}{\textbf{0.809}}\\
		\hline 
		
		\footnotesize TarDAL$_{CT}$&\footnotesize \textcolor{red}{\textbf{0.843}}&\footnotesize \textcolor{blue}{\textbf{0.863}}&\footnotesize \textcolor{red}{\textbf{0.892}}&\footnotesize \textcolor{red}{\textbf{0.762}} &\footnotesize\textcolor{red}{\textbf{0.679}}&\footnotesize \textcolor{red}{\textbf{0.807}}&\footnotesize \textcolor{red}{\textbf{0.624}}&\footnotesize 0.816&\footnotesize \textcolor{red}{\textbf{0.844}}&\footnotesize \textcolor{red}{\textbf{0.904}}&\footnotesize\textcolor{red}{\textbf{0.935}} &\footnotesize \textcolor{red}{\textbf{0.875}}&\footnotesize \textcolor{red}{\textbf{0.811}}\\
		\hline 
		
		\hline 
	\end{tabular}
	\vspace{-0.2cm}
	\caption{Quantitative comparisons of different training strategies. The best result is in {\textcolor{red}{\textbf{red}}} whereas the second best one is in {\textcolor{blue}{\textbf{blue}}}.}
	\label{tab: training}	
\end{table*}

\begin{figure}[!htb]
	\centering
	\setlength{\tabcolsep}{1pt}
	\begin{tabular}{ccccc}
		
		\includegraphics[width=0.093\textwidth,height=0.06\textheight]{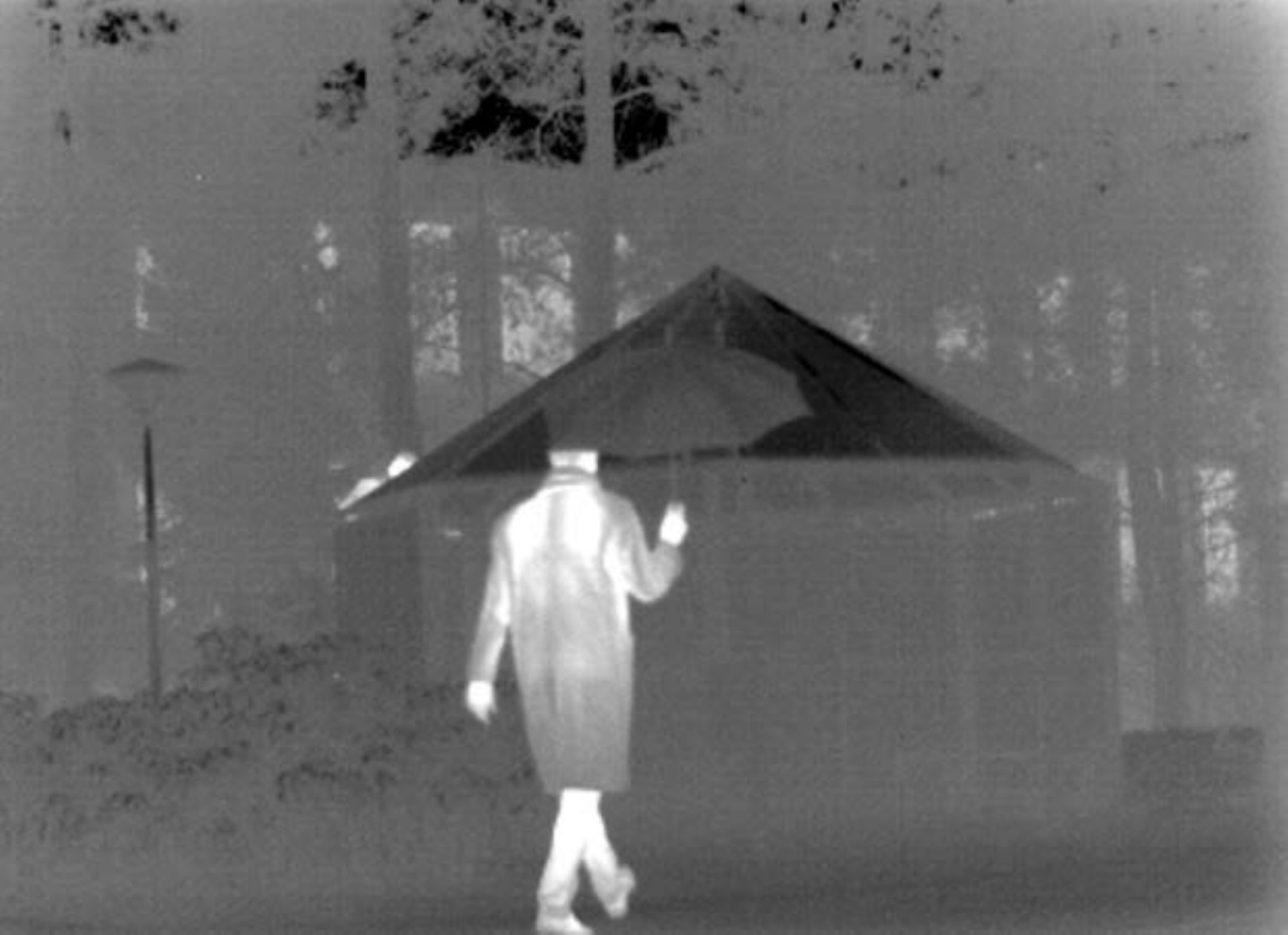}
		&\includegraphics[width=0.093\textwidth,height=0.06\textheight]{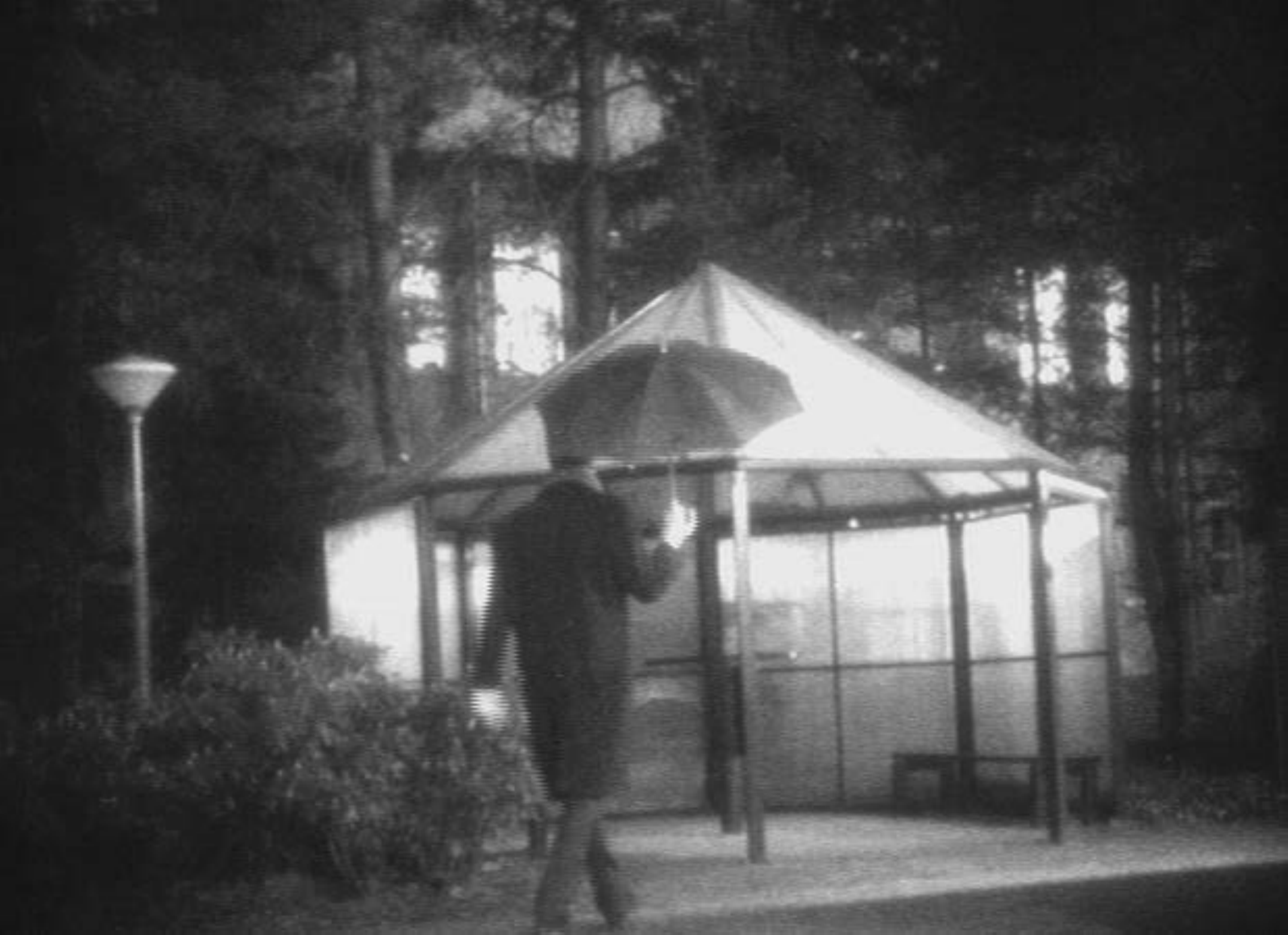}
		&\includegraphics[width=0.093\textwidth,height=0.06\textheight]{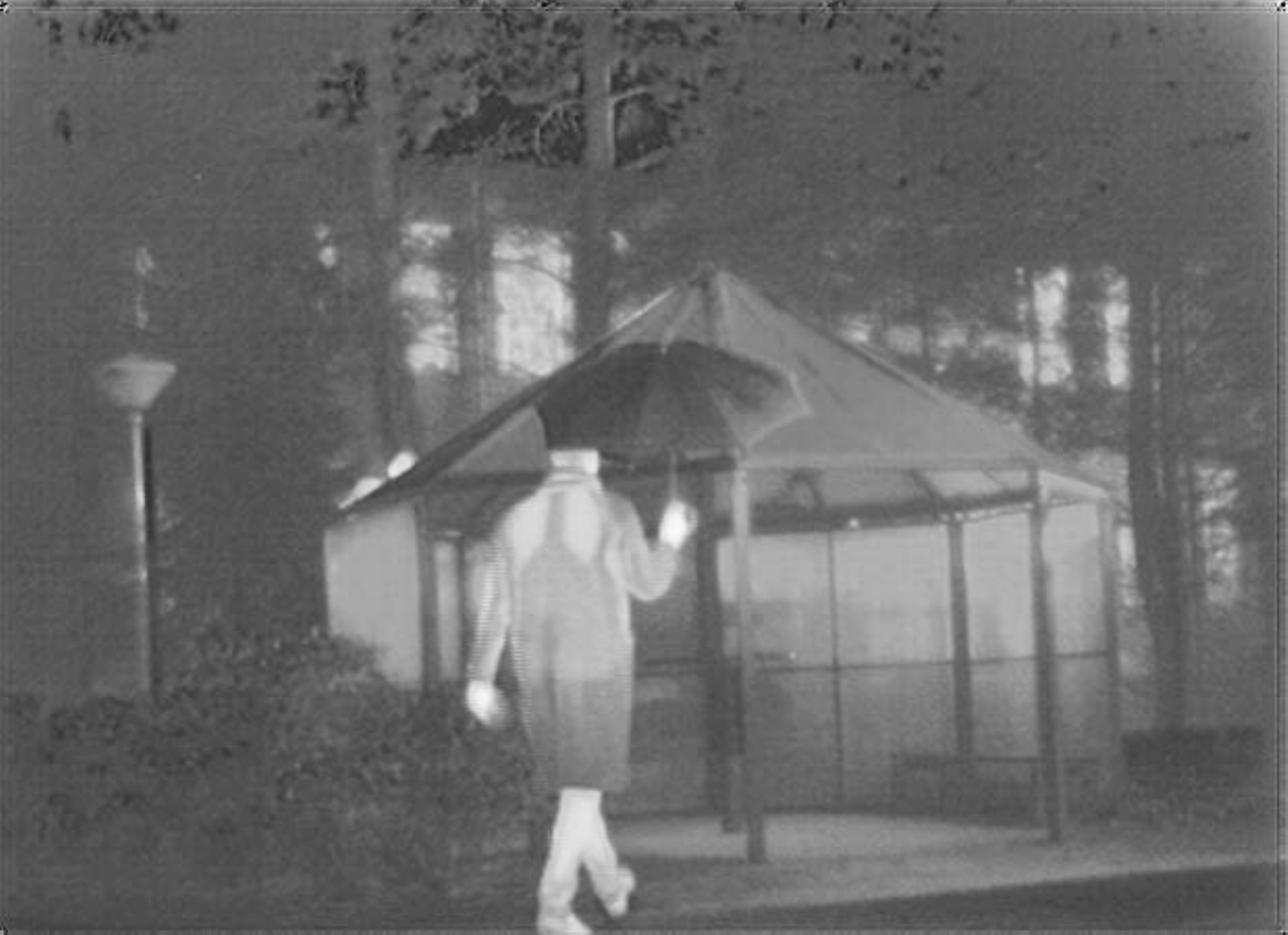}
		&\includegraphics[width=0.093\textwidth,height=0.06\textheight]{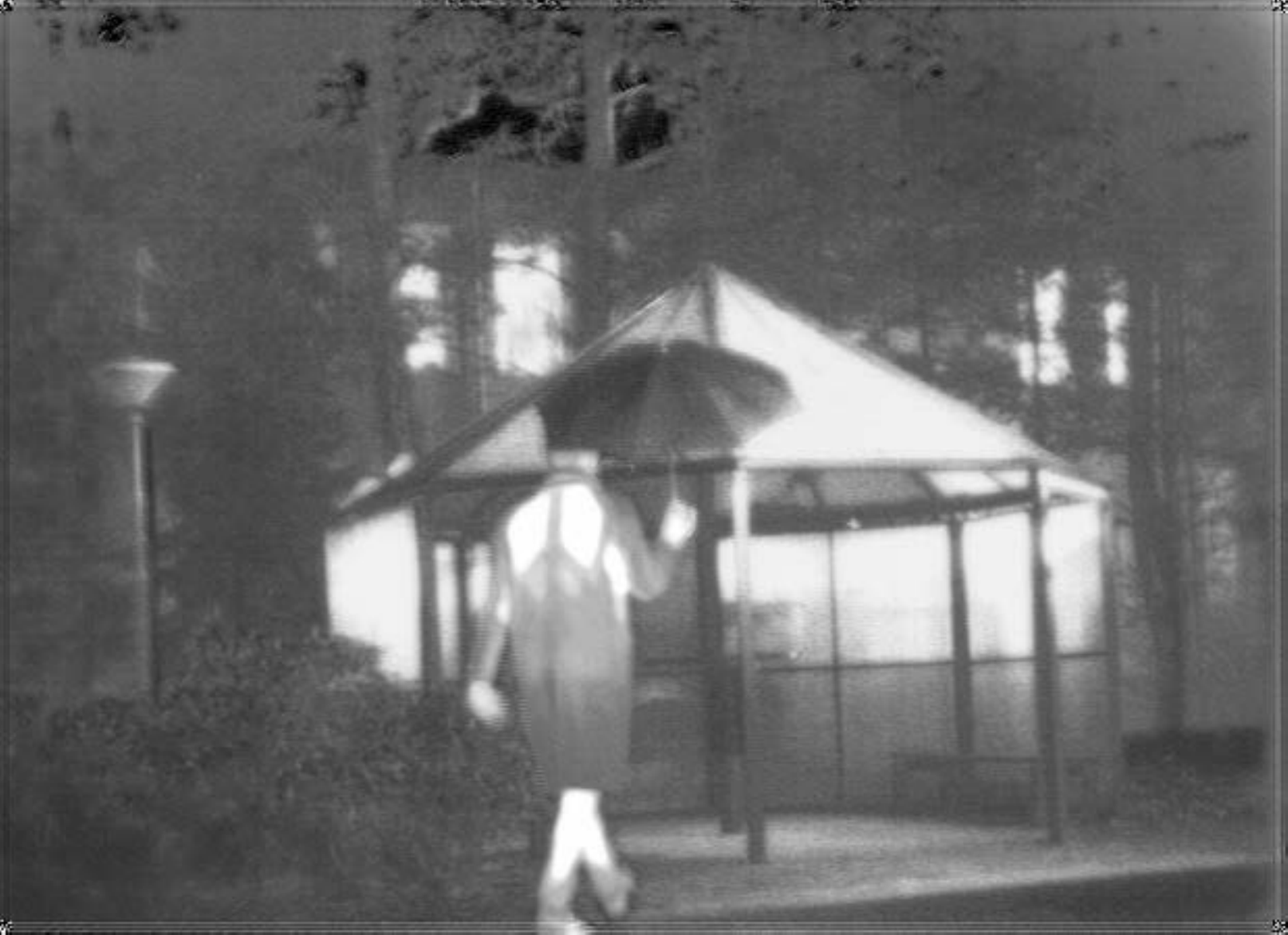}
		&\includegraphics[width=0.093\textwidth,height=0.06\textheight]{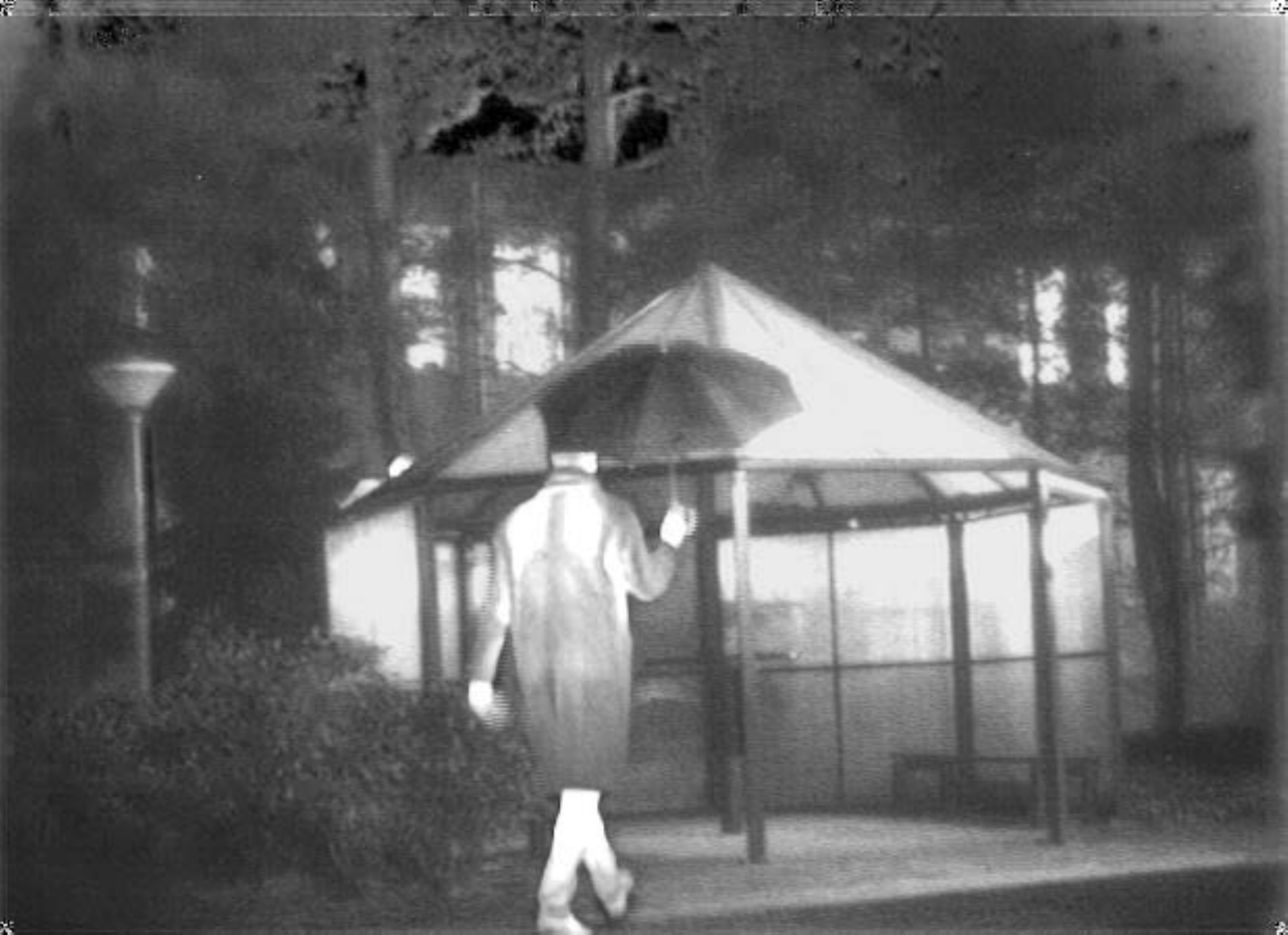}
		\\
		\includegraphics[width=0.093\textwidth,height=0.06\textheight]{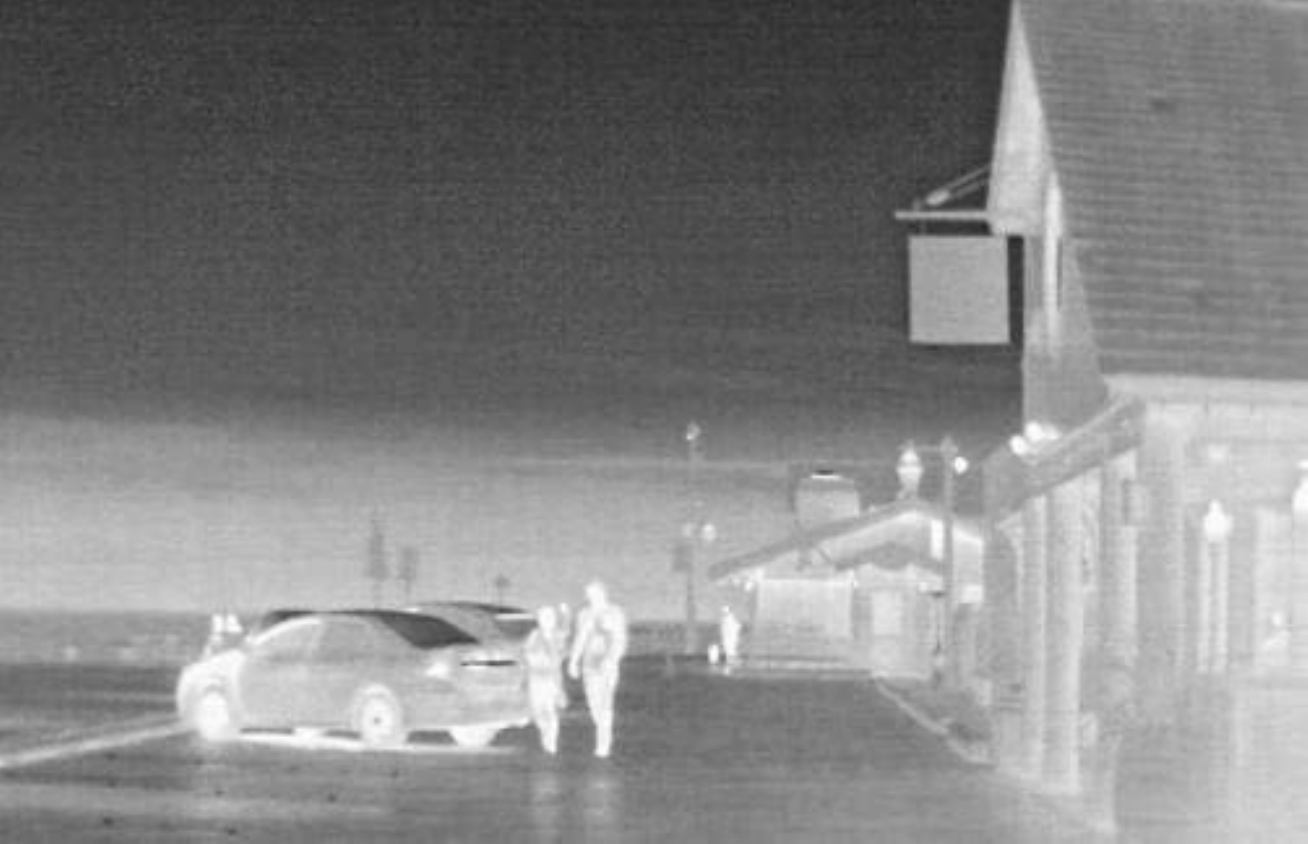}
		&\includegraphics[width=0.093\textwidth,height=0.06\textheight]{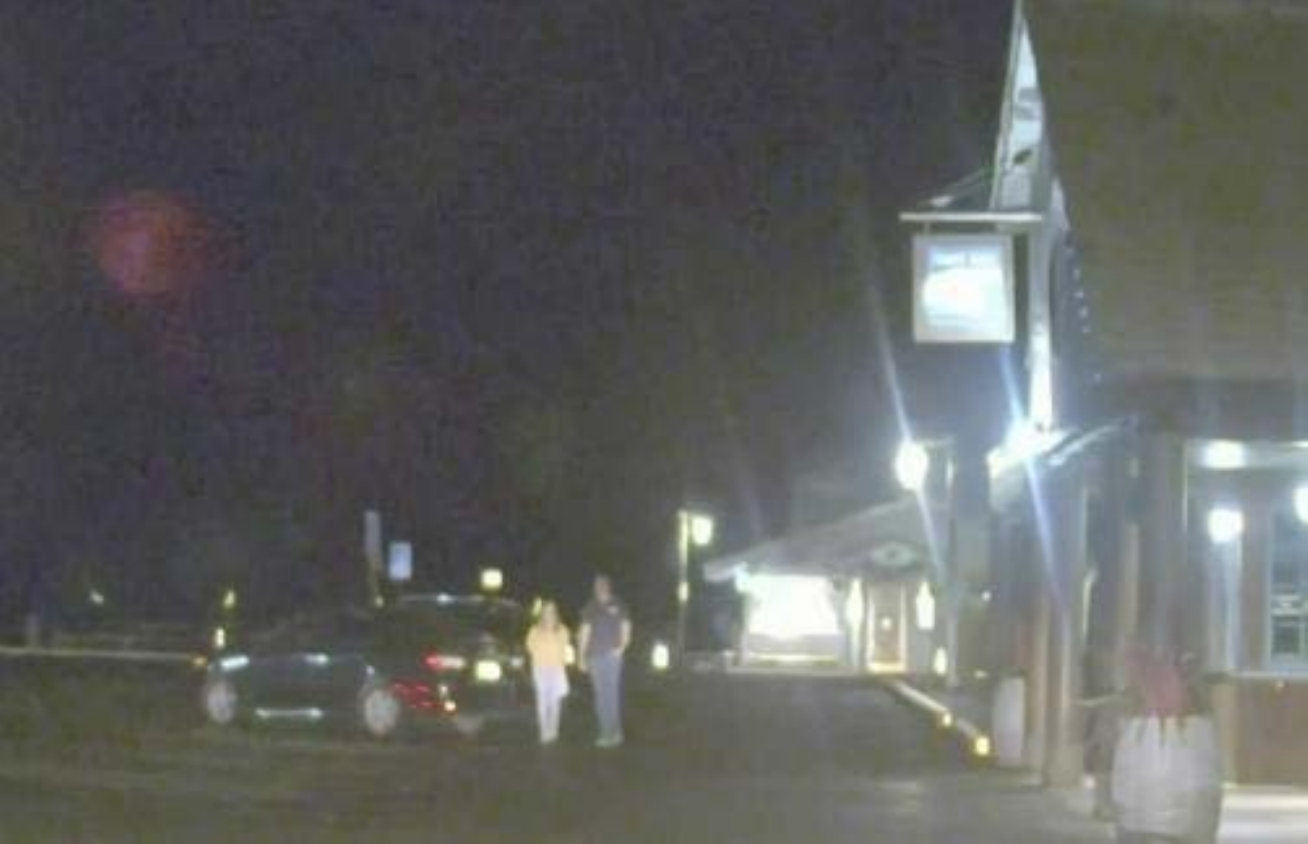}
		&\includegraphics[width=0.093\textwidth,height=0.06\textheight]{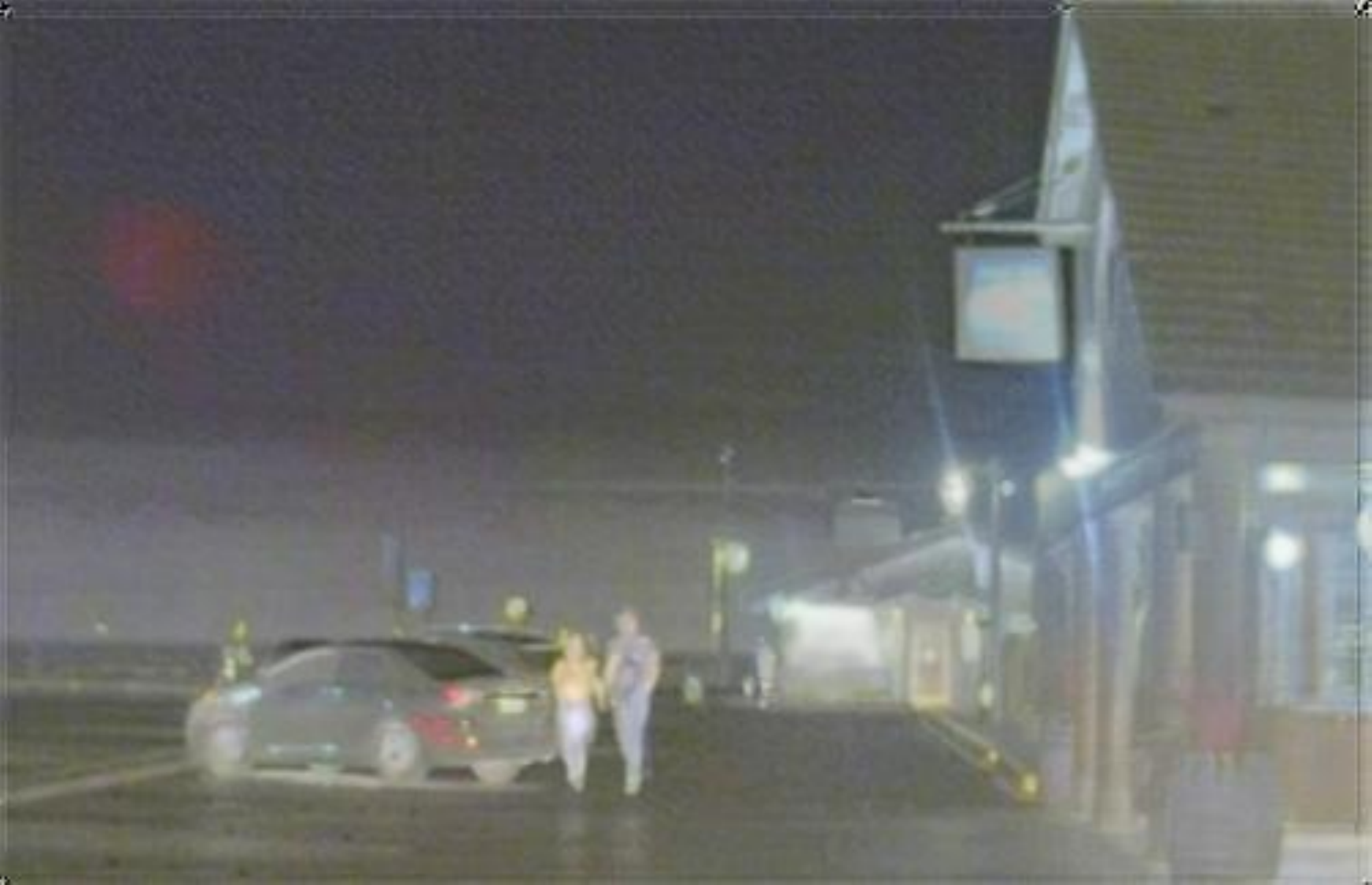}
		&\includegraphics[width=0.093\textwidth,height=0.06\textheight]{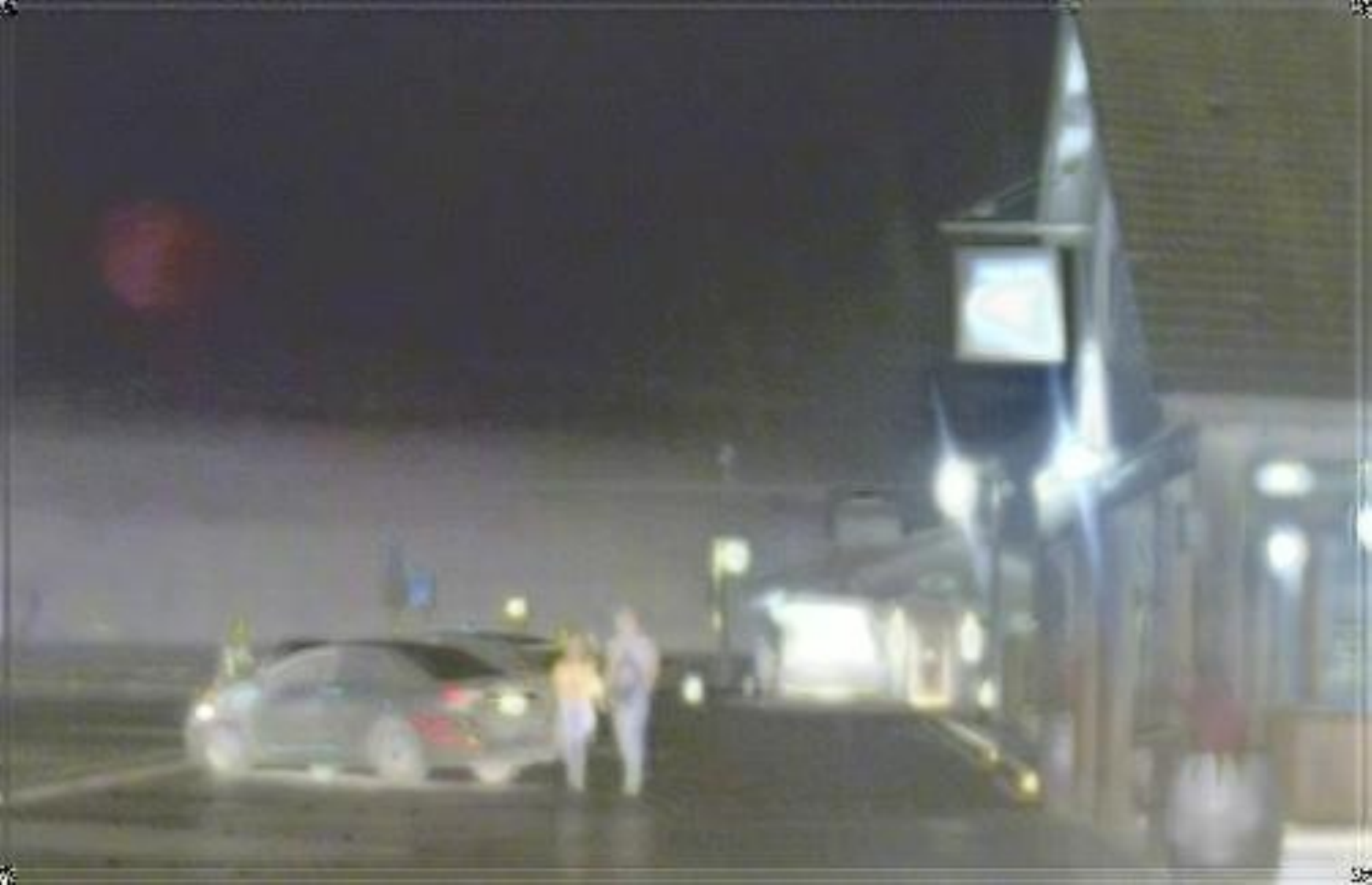}
		&\includegraphics[width=0.093\textwidth,height=0.06\textheight]{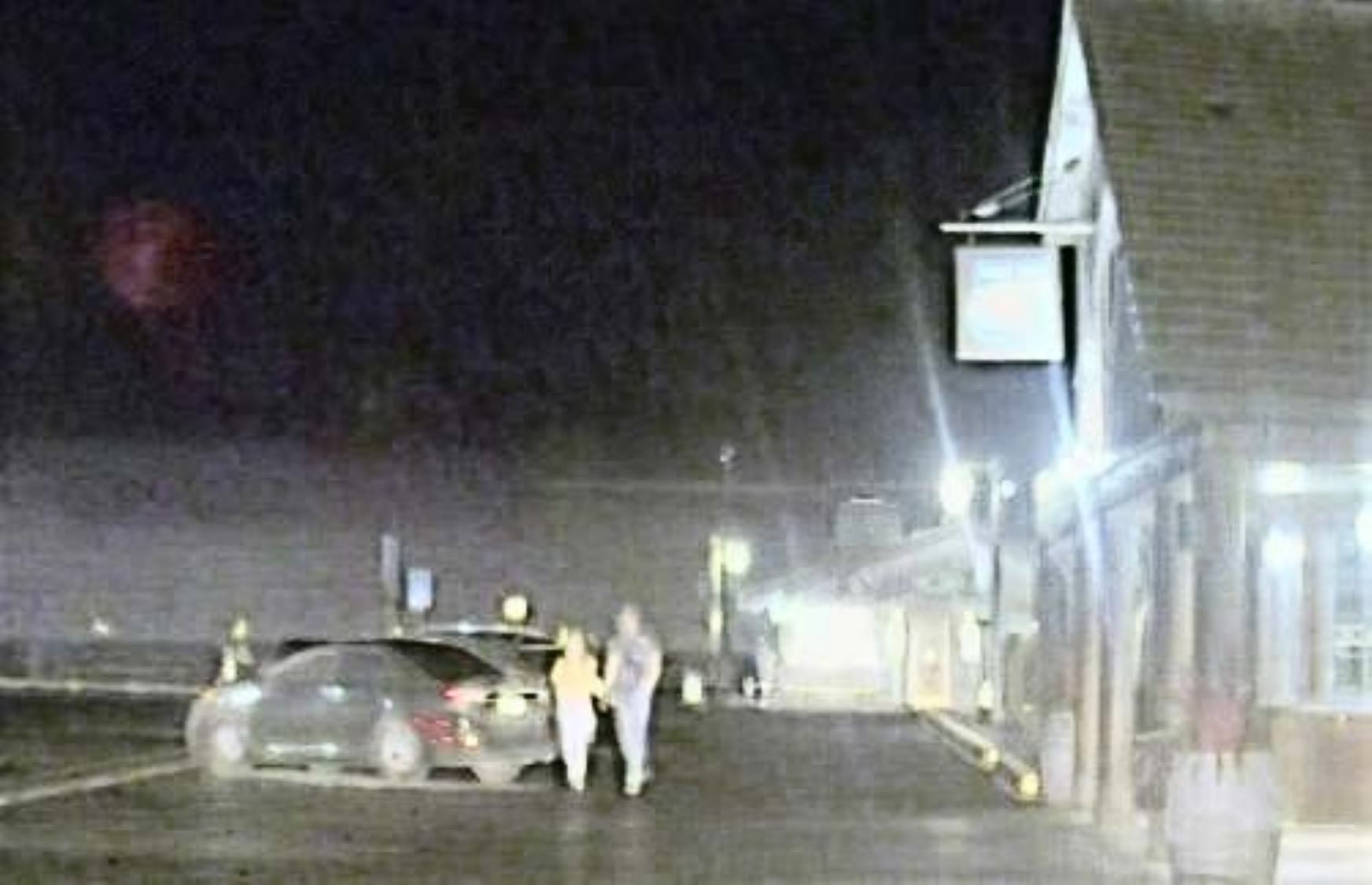}
		\\
		\includegraphics[width=0.093\textwidth,height=0.06\textheight]{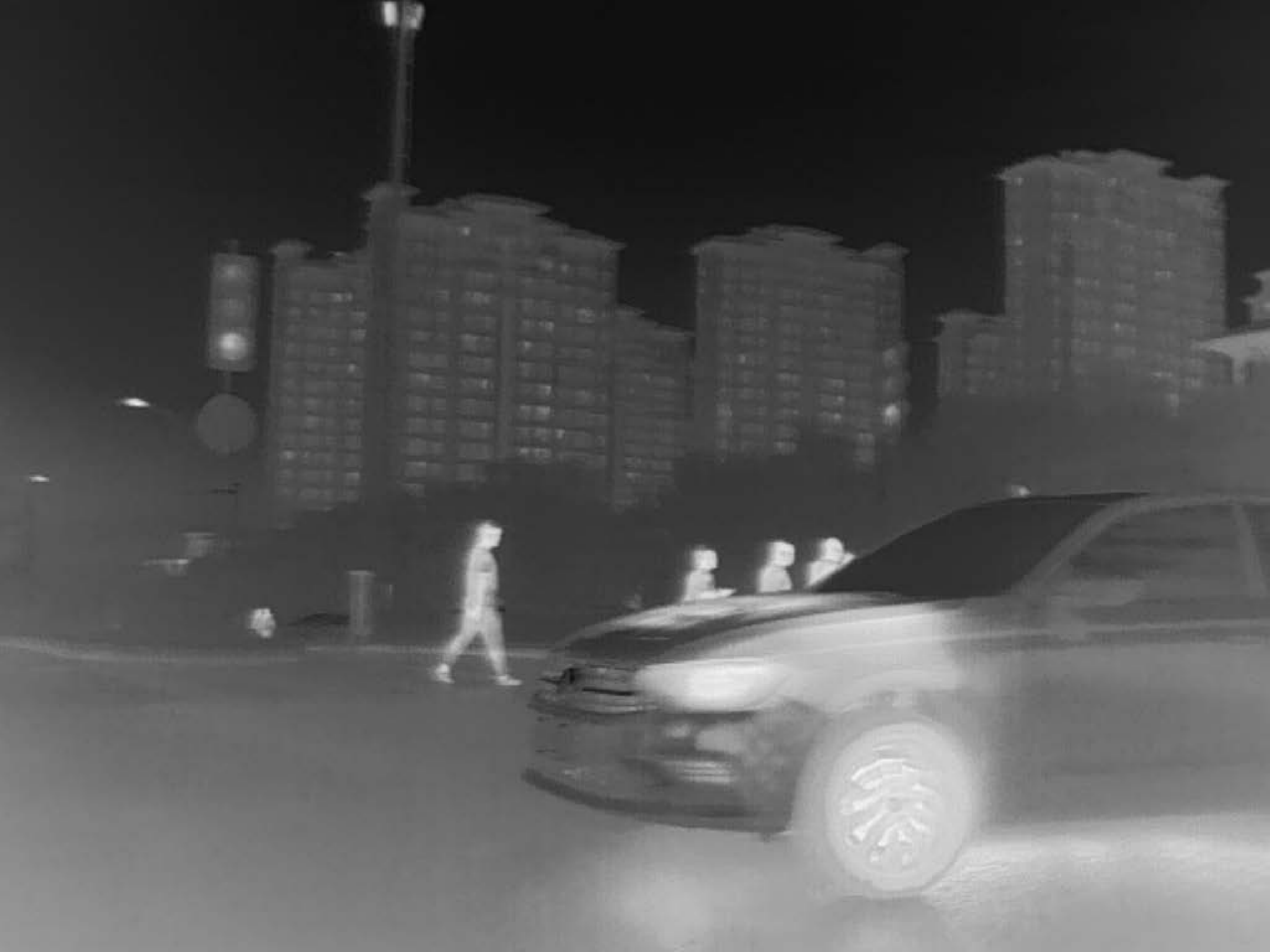}
		&\includegraphics[width=0.093\textwidth,height=0.06\textheight]{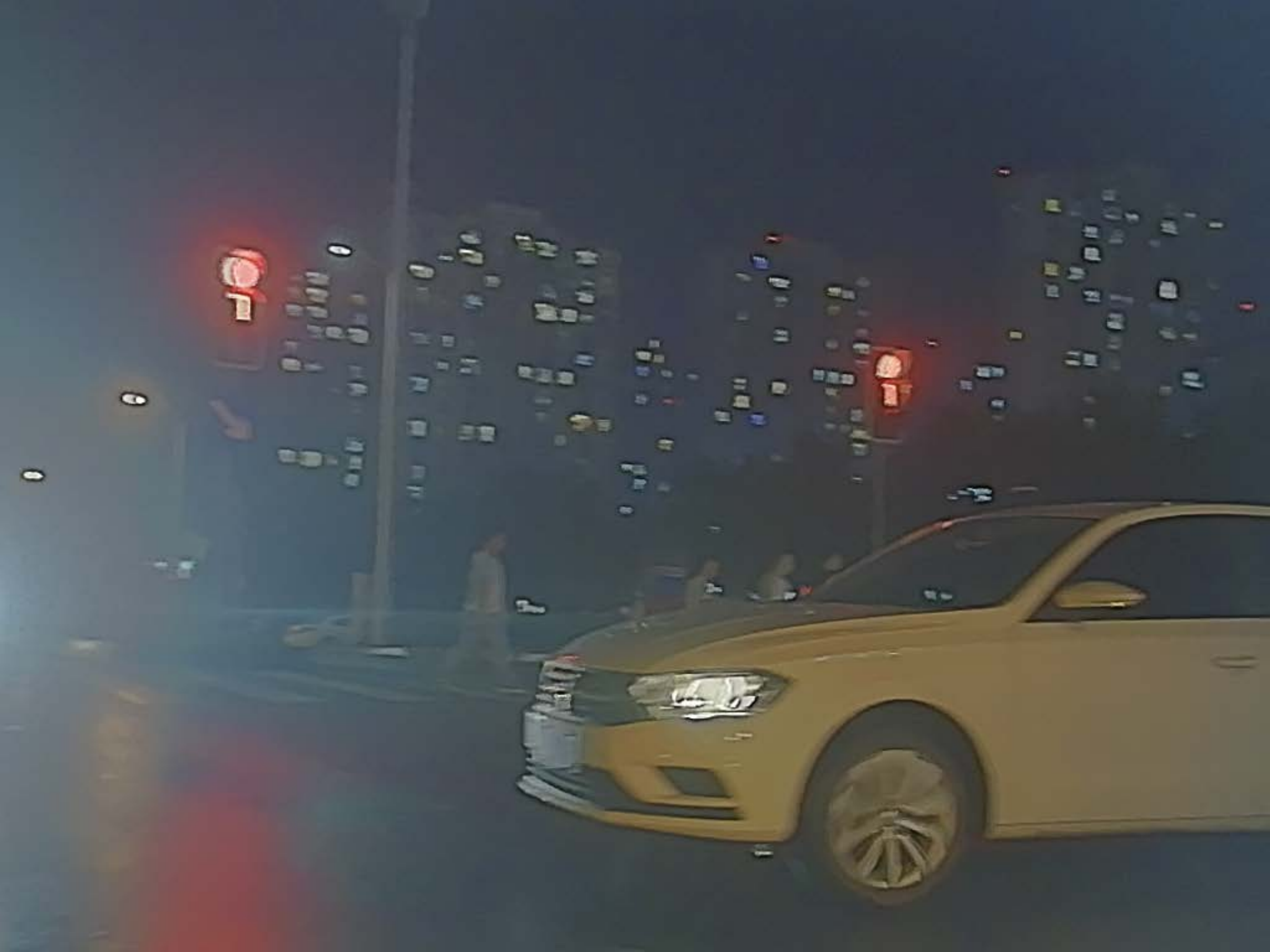}
		&\includegraphics[width=0.093\textwidth,height=0.06\textheight]{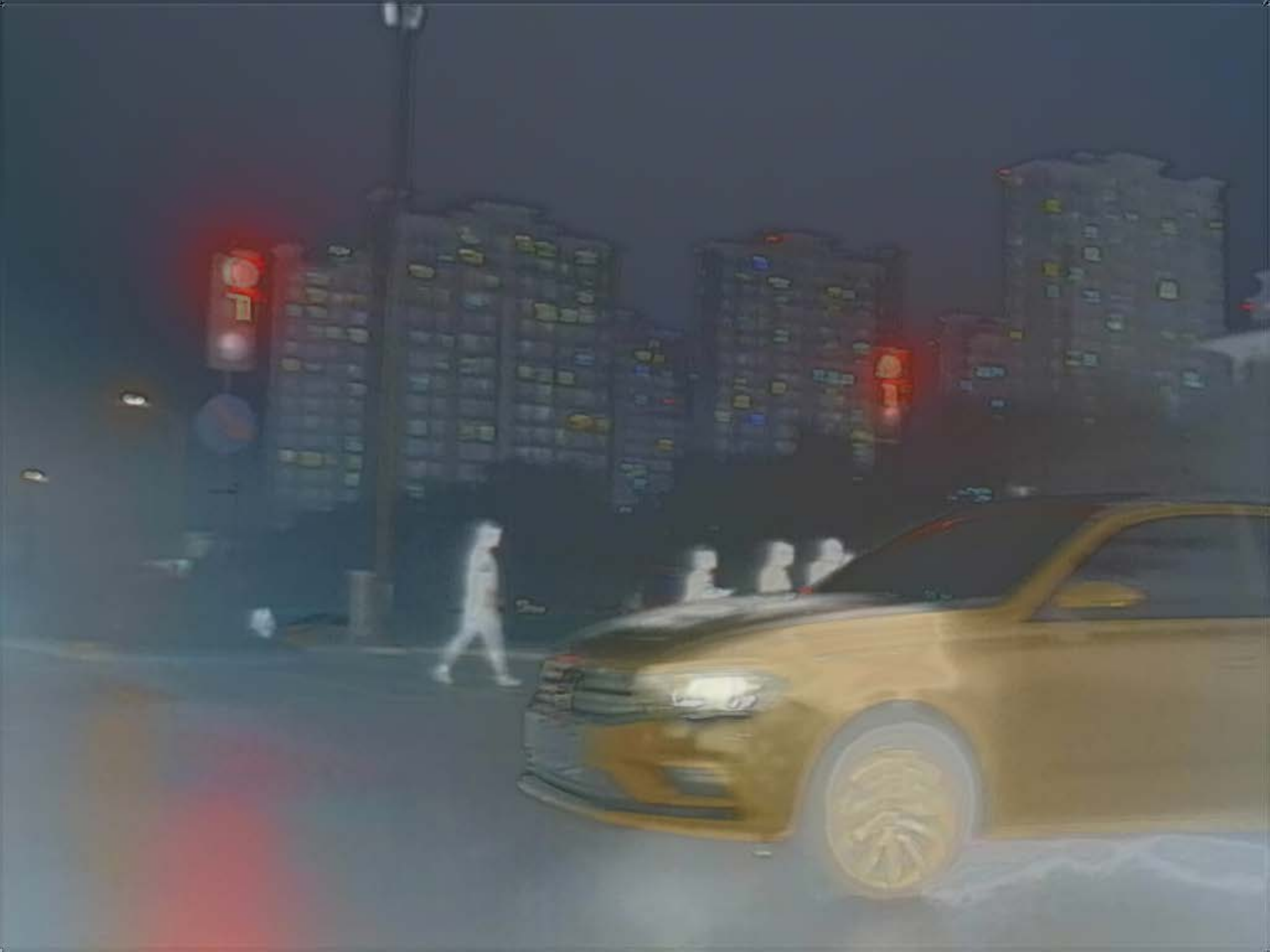}
		&\includegraphics[width=0.093\textwidth,height=0.06\textheight]{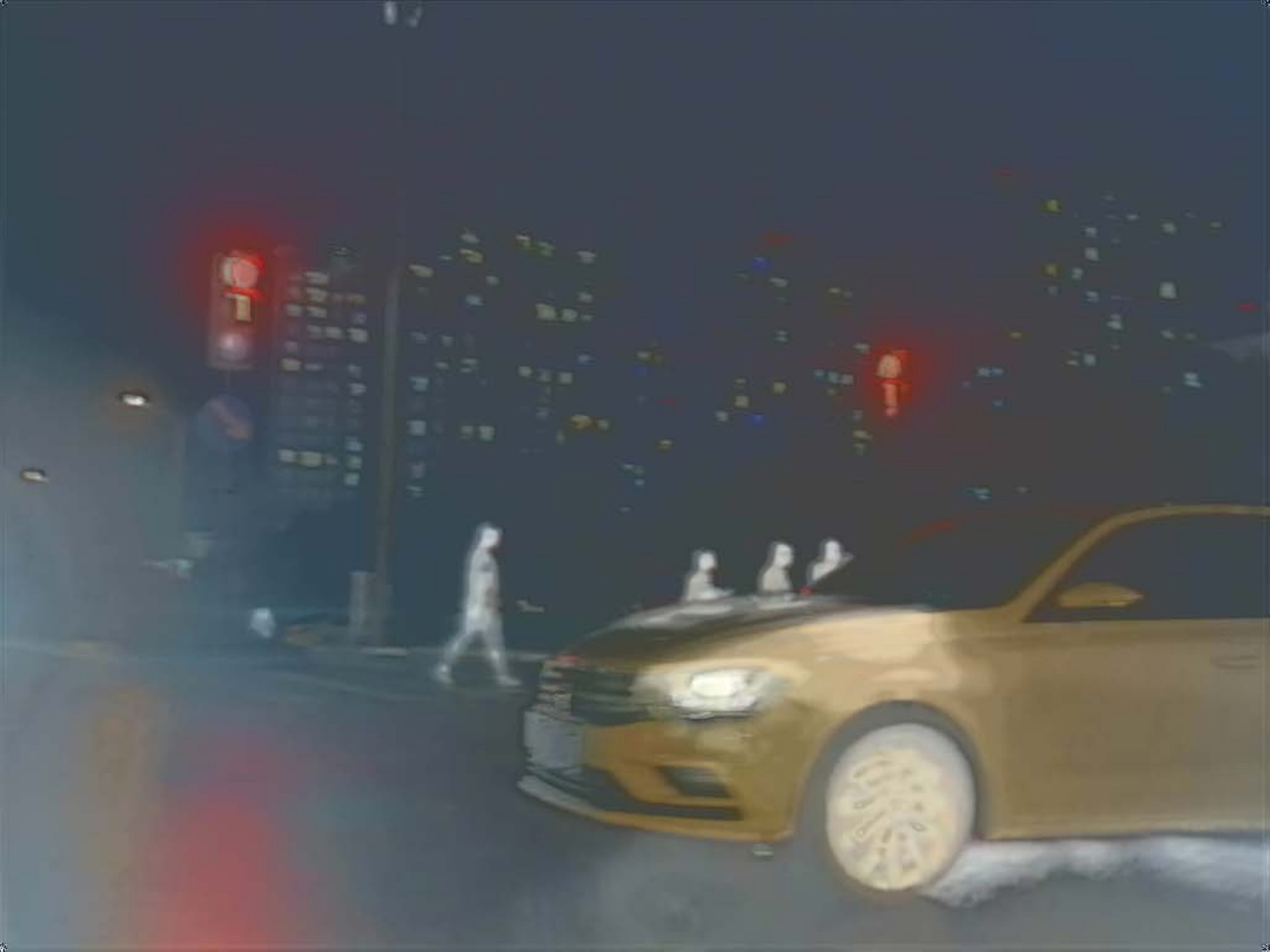}
		&\includegraphics[width=0.093\textwidth,height=0.06\textheight]{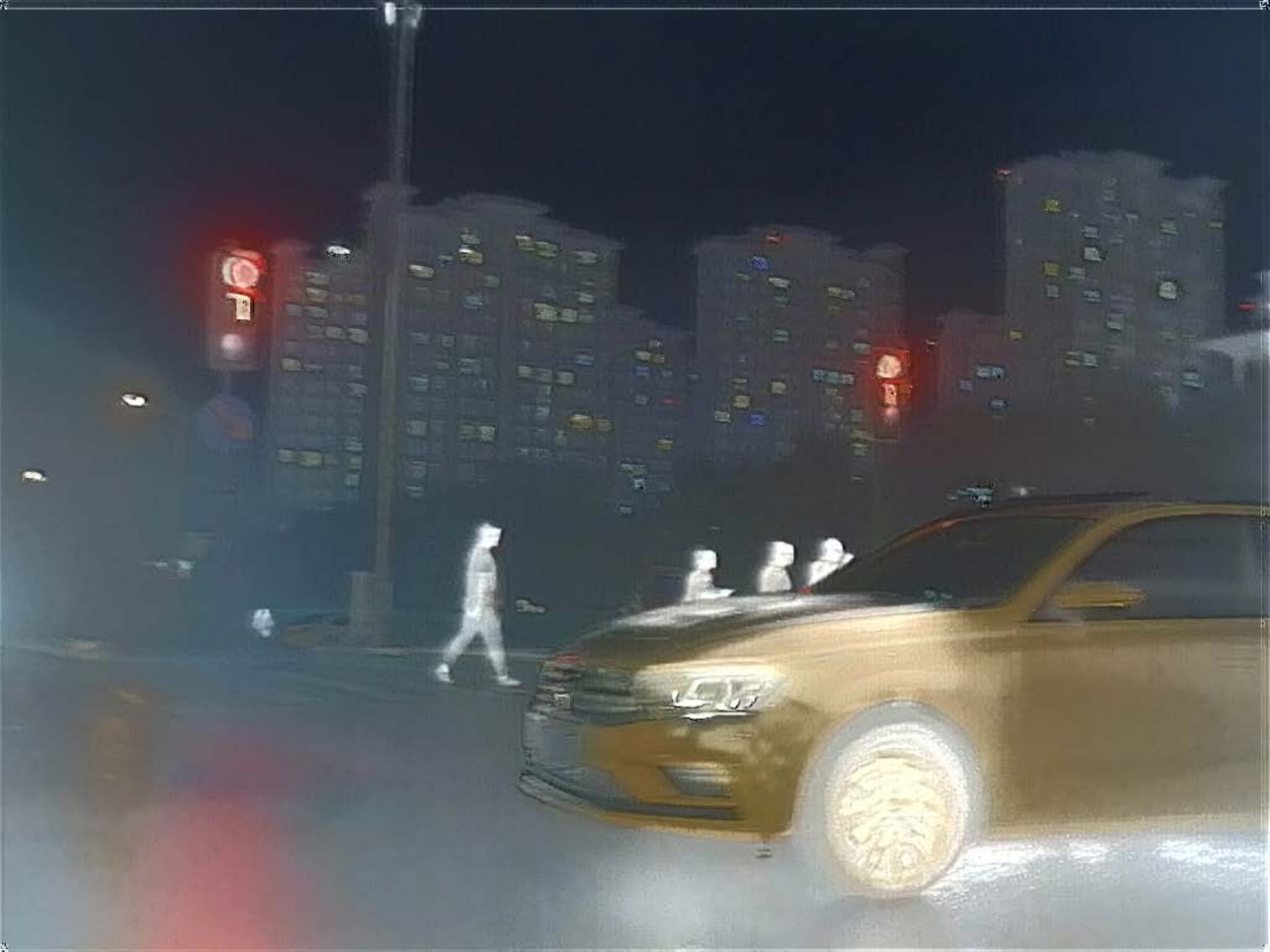}
		\\
		\footnotesize Ir image&\footnotesize Vis image&\footnotesize w/o~SDW &\footnotesize w/o~$\mathbf{m}$ &\footnotesize Ours				
	\end{tabular}
	\caption{Qualitative results on discussing loss functions. }
	\label{fig:lossfunction}
\end{figure}

\noindent\textbf{Analyzing the training loss functions}
We discuss the impact of different loss functions on our method. In Figure~\ref{fig:lossfunction}, it is easy to notice that our method can maintain much salient pixel distribution with high contrast than the method without SDW, which can illustrate the effectiveness of the newly designed SDW function. Meanwhile, the method without~$\textbf{m}$ may lose some vital details, \emph{e.g.}, leaves and chimney silhouette. This is because that~$\mathbf{m}$ allows two discriminators to perform adversarial learning under their respective region, hence paying more attention to their unique features. 

\begin{figure}[!htb]
	\centering
	\setlength{\tabcolsep}{1pt}
	\begin{tabular}{ccccc}
		
		\includegraphics[width=0.09\textwidth,height=0.06\textheight]{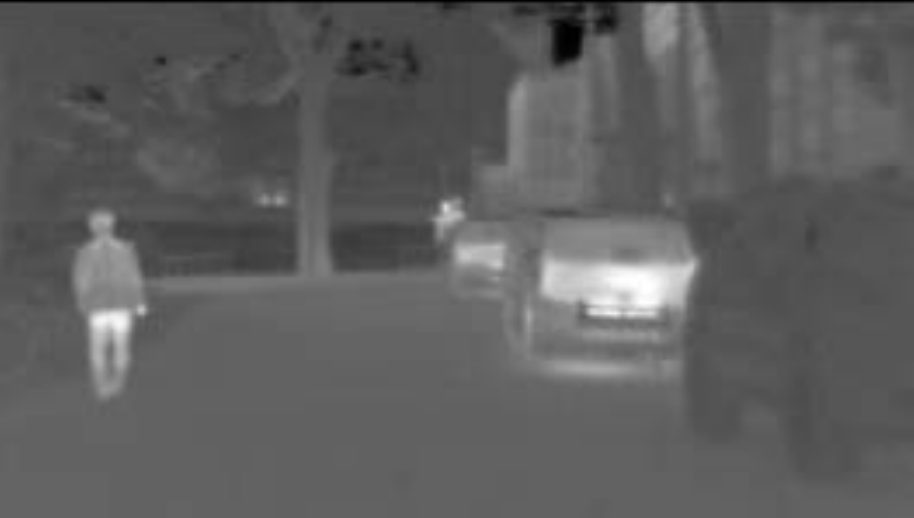}
		&\includegraphics[width=0.09\textwidth,height=0.06\textheight]{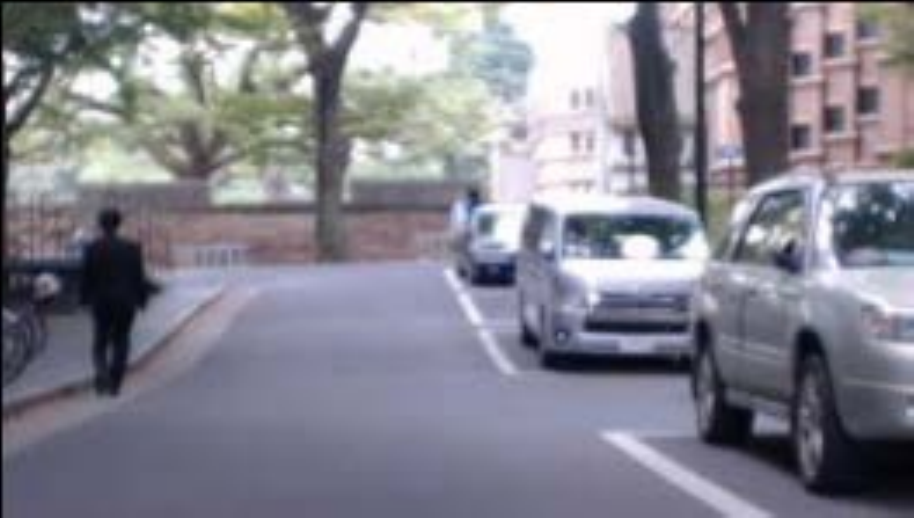}
		&\includegraphics[width=0.09\textwidth,height=0.06\textheight]{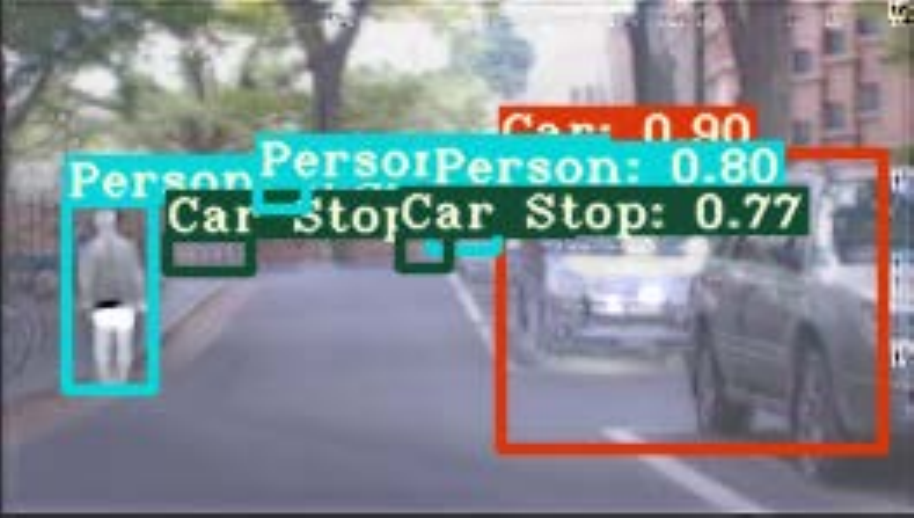}
		&\includegraphics[width=0.09\textwidth,height=0.06\textheight]{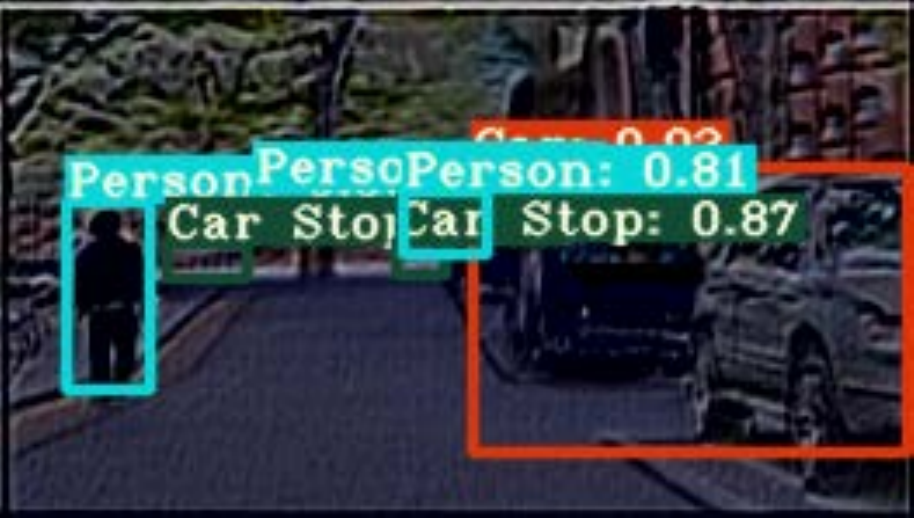}
		&\includegraphics[width=0.09\textwidth,height=0.06\textheight]{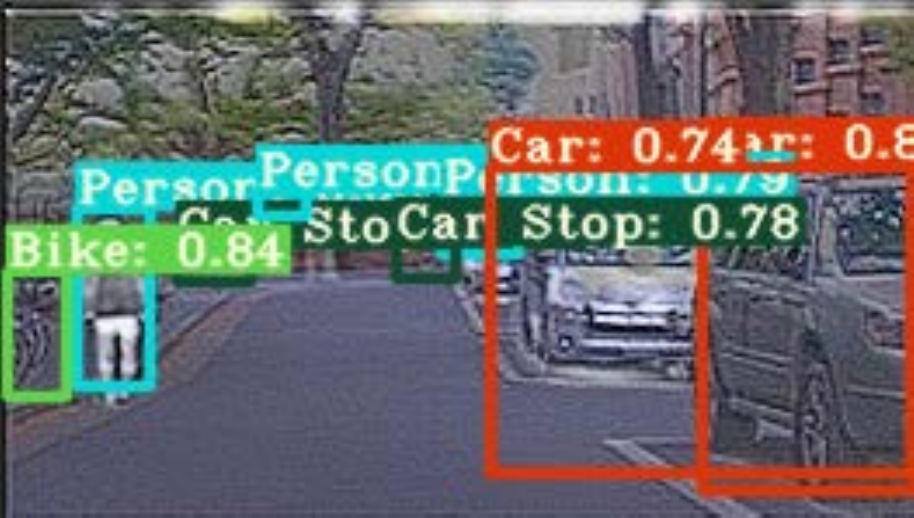}
		\\
		\includegraphics[width=0.09\textwidth,height=0.06\textheight]{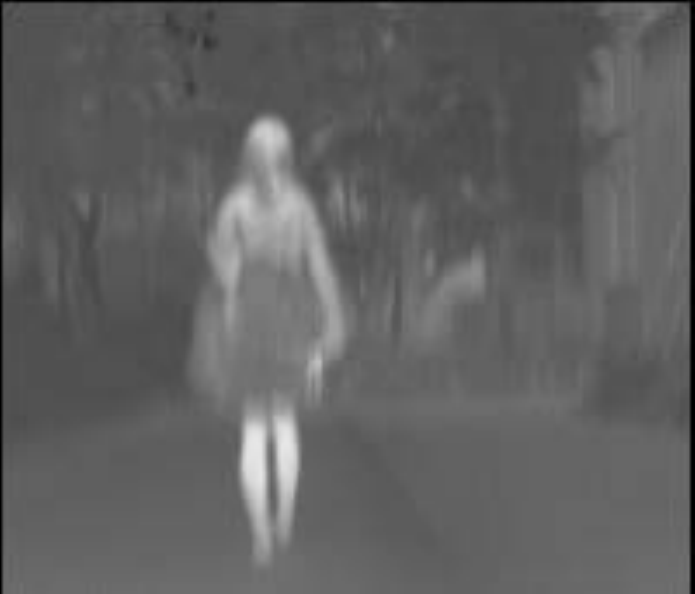}
		&\includegraphics[width=0.09\textwidth,height=0.06\textheight]{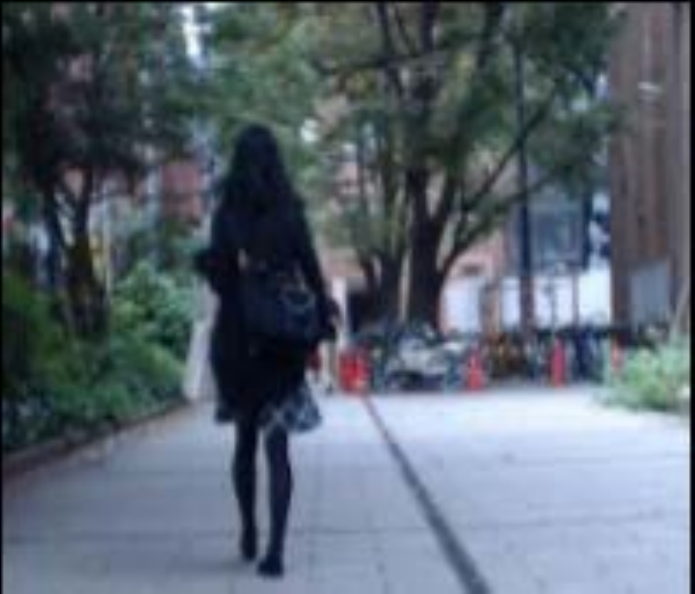}
		&\includegraphics[width=0.09\textwidth,height=0.06\textheight]{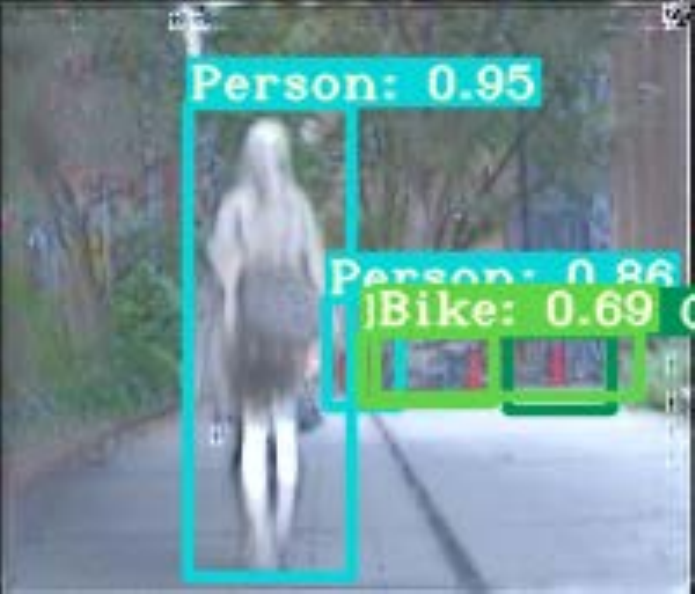}
		&\includegraphics[width=0.09\textwidth,height=0.06\textheight]{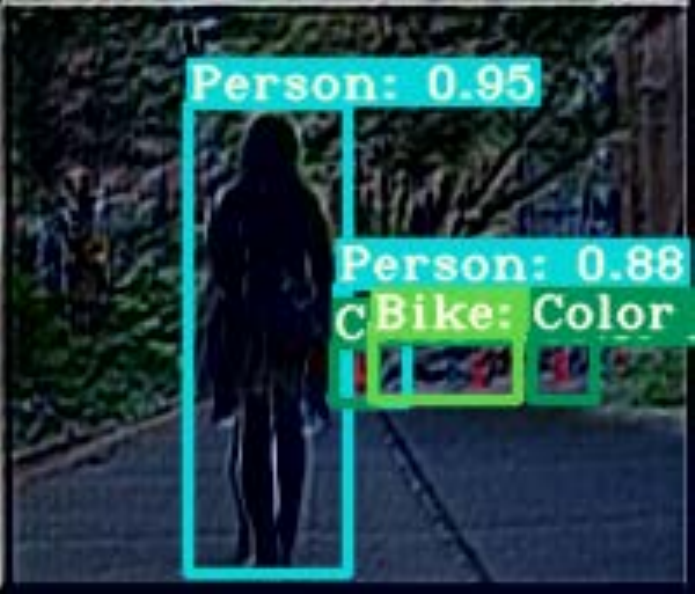}
		&\includegraphics[width=0.09\textwidth,height=0.06\textheight]{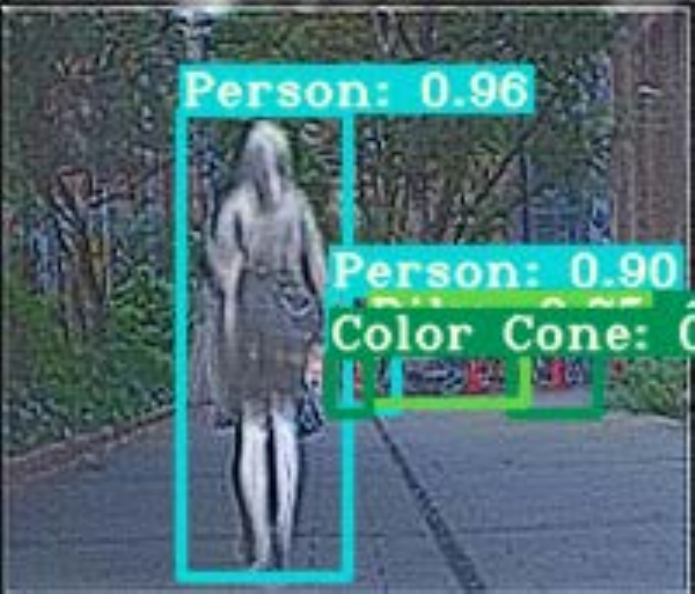}
		\\
		\footnotesize Ir image&\footnotesize Vis image&\footnotesize TarDAL$_{DT}$&\footnotesize TarDAL$_{TT}$&\footnotesize TarDAL$_{CT}$				
	\end{tabular}
	\caption{Visual comparisons of different training strategies. }
	\label{fig:trainings}
\end{figure}
\noindent\textbf{Evaluating different versions of training strategy}
We further verify the advantages of our cooperative training~($CT$) in comparing with direct training~($DT$) and task-oriented training~($TT$). As shown in Figure~\ref{fig:trainings}, $TT$ only uses detection loss to train the network, resulting in a worse visual effect for observation. In contrast, $CT$ has a significant advantage in boosting the detection performance and better visual effects.  The same trend can be found in Table~\ref{tab: training}, $CT$ reaches the largest or the second-largest scores among the two different datasets.

\section{Conclusion}
\vspace{-0.1cm}
Within this paper,~a bilevel optimization formulation for jointly realizing fusion and detection is proposed. By unrolling the model to a well-designed fusion network and a commonly used detection network, we can generate a visual-friendly result for fusion and object detection. To promote future researches in this field, we raise a synchronized imaging system with visible-infrared sensors and collect a multi-scenario multi-modality benchmark.

{\small
\bibliographystyle{ieee_fullname}
\bibliography{egbib}
}

\end{document}